\documentclass{elsarticle}

\usepackage{url}
\usepackage{tikz}
\usepackage[utf8]{inputenc}
\usepackage{graphicx}
\usepackage{amsmath}
\usepackage{amsthm}
\usepackage{booktabs}
\usepackage{algorithm}
\usepackage{algorithmic}
\usepackage{microtype}
\usepackage{amssymb}
\usepackage{inconsolata}
\usepackage{problogconfig}
\usepackage{amsthm}
\usepackage{mdframed}
\usepackage{rotating}
\usepackage{xcolor}
\usepackage{makecell}
\usepackage{diagbox}
\usepackage{subcaption}
\usepackage{bbm}
\usepackage[most]{tcolorbox}
\usepackage{afterpage}

\newcommand{\smallimg}[1]{\vcenter{\hbox{\frame{\includegraphics[height=10pt]{images/#1}}}}}
\newcommand{\digit}[1]{\vcenter{\hbox{\includegraphics[height=12pt]{mnist/#1}}}}

\newtcolorbox[auto counter]{boxedexample}[2][]{
		breakable,
		left=0pt,
		right=0pt,
		top=0pt,
		bottom=0pt,
		colback=gray!10,
		colframe=gray!10,
		width=\dimexpr\textwidth\relax, 
		enlarge left by=0mm,
		boxsep=5pt,
		arc=0pt,outer arc=0pt,
		fonttitle=\scshape,
		coltitle=black,
		title={Example \thetcbcounter: 
            #2},
            #1
}
\tcbuselibrary{listings,breakable}
\newtcolorbox[auto counter]{dimension}[1][]{
	enhanced,
	breakable,
	float,
	fonttitle=\scshape,
	title={Dimension \thetcbcounter: #1}	
}
\tcbuselibrary{listings,breakable}
\newtcolorbox{evidencebox}[1][]{
		breakable,
		left=0pt,
		right=0pt,
		top=0pt,
		bottom=0pt,
		colback=blue!10,
		colframe=blue!10,
		width=\dimexpr\textwidth\relax, 
		enlarge left by=0mm,
		boxsep=5pt,
		arc=0pt,outer arc=0pt,
}
\tcbuselibrary{listings,breakable}
\newcommand{\nesy}{NeSy}

\definecolor{verydarkgreen}{rgb}{0.0, 0.5, 0.0}

\newcommand{\old}[1]{} 

\newcommand{\rev}[1]{\textcolor{black}{#1}}

\theoremstyle{definition}

\begin{document}
	
	\begin{frontmatter}
		
		\title{From Statistical Relational to Neurosymbolic \\ Artificial Intelligence: a Survey.}
		
		\author[1]{Giuseppe Marra}
		\ead{giuseppe.marra@kuleuven.be}
		
		\author[3]{Sebastijan Dumančić}
		\ead{s.dumancic@tudelft.nl}
		
		\author[1]{Robin Manhaeve}
		\ead{robin.manhaeve@kuleuven.be}
		
		\author[1,2]{Luc De Raedt}
		\ead{luc.deraedt@kuleuven.be}
		
		\address[1]{KU Leuven, Department of Computer Science and Leuven.AI}
		\address[2]{Örebro University, Center for Applied Autonomous Sensor Systems}
		\address[3]{Delft University of Technology, Department of Software Technology}

		\begin{abstract}

   \rev{This survey explores the integration of learning and reasoning in two different fields of artificial intelligence: neurosymbolic  and statistical relational artificial intelligence. Neurosymbolic artificial intelligence  (NeSy) studies the integration of symbolic reasoning and neural networks, while statistical relational artificial intelligence (StarAI) focuses on integrating logic with probabilistic graphical models. This survey identifies  seven shared dimensions between these two subfields of AI. These dimensions can be used to characterize different NeSy and StarAI systems. They are concerned with (1) the approach to logical inference, whether model or proof-based; (2) the syntax of the used logical theories; (3) the logical semantics of the systems and their extensions to facilitate learning; (4) the scope of learning, encompassing either parameter or structure learning; (5) the presence of symbolic and subsymbolic representations;  (6) the degree to which  systems capture the original logic, probabilistic, and neural paradigms; and (7) the classes of learning tasks the systems are applied to. By positioning various NeSy and StarAI systems along these dimensions and pointing out similarities and differences between them, this survey contributes  fundamental concepts for understanding the integration of learning and  reasoning.
}
		\end{abstract}

	\end{frontmatter}

	\section{Introduction}

	The integration of learning and reasoning is a key challenge in artificial intelligence and machine learning today. Various communities are addressing it, especially
	 the field of neurosymbolic artificial intelligence (NeSy) \cite{besold2017neural,garcez2019neural}.
	NeSy's goal is to integrate symbolic reasoning with neural networks.
	\nesy{} already has a long tradition, and it has recently attracted a lot of attention.
Indeed, the topic has been addressed by prominent researchers such as Y. Bengio and H. Kautz  in their keynotes at AAAI 2020, by Y. Bengia and G. Marcus in the AI Debate \cite{aidebate} and Hochreiter has recently stated \cite{hochreiter2022toward} that \nesy{} is ``the most promising approach to a broad AI''.

	
	Another domain with a rich tradition in integrating learning and reasoning is that of statistical relational learning and artificial intelligence (StarAI) \cite{Getoor07:book,DeRaedtKerstingEtAl16}. StarAI focuses on integrating logical and probabilistic reasoning.

 
Historically, these two endeavours have adopted  different learning paradigms, probabilistic versus neural, for integrating logic into machine learning. This in turn has resulted in two different subcommunities.  StarAI has focused on probabilistic logics,  their semantics and making inference more tractable, while learning is usually based on parameter learning techniques from probabilistic graphical models.
On the other hand,  \nesy{} extends neural networks with symbolic knowledge, focusing on scalable approximate models, paying less attention to semantical issues. In particular, \nesy{} techniques can often be characterized by a clear parameterization in terms of neural networks, i.e. layered structures of latent representations, and by resorting to the gradient-based backpropagation paradigm for learning.
\rev{Despite a different focus and approach, the  two domains want to achieve the same goal, that is, to integrate learning and reasoning. 
It is therefore surprising that there has been relatively little interaction between the two domains, but see \cite{Dagstuhl14381, Dagstuhl17192}}.

	This discrepancy is the key motivation behind this survey: it aims at pointing out the similarities between these two endeavours and, in this way, it wants to stimulate 
	cross-fertilization. 
	We start from the literature on StarAI, following the key concepts, \rev{definitions} and techniques outlined in several textbooks and tutorials such as \cite{russell2015unifying,DeRaedtKerstingEtAl16}, because it turns out that  the same issues and techniques that arise in StarAI apply to \nesy{}  as well.

	\rev{The key contributions of this paper are:}
	\begin{enumerate}
		\item 
		{\em We identify seven dimensions that these fields have in common and that can be used to categorize both StarAI and \nesy{} approaches}.  
		These seven dimensions are concerned with (1) model vs proof-based inference,  \rev{(2) logic syntax,} (3) semantics, (4) learning parameters or structure, (5) representing entities as symbols or subsymbols, (6)  integrating logic with probabilistic and/or neural concepts, and \rev{(7) learning tasks}.
		\item We provide evidence for our claim by positioning a wide range of StarAI and \nesy{} systems along these dimensions and pointing out analogies between them. 
		This provides not only new insights into the relationships between StarAI and NeSy, but it also allows one to carry over and adapt techniques from one field to another. These  insights provide  opportunities for cross-fertilization between StarAI and NeSy, by focusing on those dimensions that have not  been fully exploited yet.
		
	\item  \rev{We gently introduce key logical concepts and techniques inherited from StarAI. In this way, the paper also provides a gentle introduction to symbolic AI and  StarAI techniques for the interested ``connectionist'' practitioner.} 
		\item We illustrate each dimension using  existing methods, 
  and in this  way, also present an intuitive and concrete overview of the research field. 
	\end{enumerate}

	Unlike some other perspectives on neurosymbolic computation \cite{besold2017neural,garcez2019neural,chaudhuri2021neurosymbolic}, the present survey limits itself to a logical perspective and to developments in neurosymbolic computation that are consistent with this perspective. \rev{Therefore, we usually refer to symbols and symbolic algorithms as synonyms for logical representations and logical reasoning.}
	Furthermore, the survey focuses on representative and prototypical systems rather than aiming at completeness (which would not be possible given the fast developments in the field). 
	Several other surveys about neurosymbolic AI have been proposed. An early overview of neurosymbolic computation is that of \cite{bader2005dimensions}. Unlike the present survey it focuses very much on a logical and a reasoning perspective. Today, the focus has shifted very much to learning. More recently, \cite{lamb2020graph} analysed the intersection between NeSy and graph neural networks (GNN). \cite{VanBekkum2021} described neurosymbolic systems in terms of the composition of blocks described by few patterns, concerning processes and exchanged data. In contrast, this survey is more focused on the underlying principles that govern such a composition. Finally, \cite{dash2021tell} exploits  a neural network viewpoint by investigating in which components (i.e. input, loss or structure) symbolic knowledge is injected.

	\paragraph{Structure of the paper}
        \rev{The next seven sections each describe one dimension by first introducing the main underlying concepts, either based on logic, probability or machine learning, and then showing how they are incorporated in StarAI and NeSy systems. Section \ref{sec:proof_vs_model} presents how to use logic for inference by distinguishing between proof-based and model-based systems, while Section \ref{sec:syntax} introduces logic at the syntac level, in particular, propositional, relational and first-order logic. Section \ref{sec:semantics}  then introduces the semantics of logic and shows how to extend it to a continuous semantics, using fuzzy and probabilistic logics. Section \ref{sec:struct} discusses the dimension of learning, distinguishing parameter learning from structure learning. Section \ref{sec:symb_vs_subsymb} focuses on the representational level and to what extent neurosymbolic models use symbolic and/or subsymbolic features. Section \ref{sec:paradigms} positions neurosymbolic approaches along the spectrum of three main paradigms, i.e. logic, probability and neural networks. \rev{Section \ref{sec:tasks} describes general classes of learning tasks to which neurosymbolic systems are usually applied.} Finally, in Section 10, we conclude by introducing open challenges in the neurosymbolic landscape.}
	
	We summarize various neurosymbolic approaches along these dimensions in Table \ref{tab:big_table}.


	\begin{sidewaystable}[ht!]
    \setlength{\tabcolsep}{2.8pt}
    \renewcommand{\arraystretch}{1.2}
    \centering
    \begin{tabular}{l|ccccccc}

    \hline
        \textbf{Frameworks} & \textbf{Inference} & \textbf{Syntax} & \textbf{Semantics} & \textbf{Learning} & \textbf{Representations} & \textbf{Paradigms} & \textbf{Tasks}\\ \hline
        ~ & \makecell[l]{\small (P)roof \\ (M)odel} & \makecell[l]{(P)ropositional \\ (R)elational \\  (FOL)} & \makecell[l]{(M)inimal \\ (S)table \\ (C)lassical \\ (F)uzzy \\ (P)robability} & \makecell[l]{(P)arameters \\ (S)tructure} & \makecell[l]{(S)ymbolic \\ (Sub)symbolic} & \makecell[l]{Logic (L/l) \\ Probability (P/p) \\ Neural(N/n)}  & \makecell[l]{(D)istant (S)upervision \\ (S)emi (S)upervised  \\ (KGC)ompletion \\ (G)enerative \\ (K)nowledge (I)nduction}\\  \hline
        
        $\alpha$ILP \cite{Shindo2023} & P+M & FOL & S + P & P + S  & S & Ln & KI \\ \hline
        
        $\partial$ILP \cite{evans:dilp}  & P & R & M + F & P + S  & S & Ln & DS + KI \\ \hline
        
        DeepProbLog \cite{manhaeve2018deepproblog} & P+M & FOL & M + P & P+S & S+Sub & LpN & DS + KI \\ \hline

        DeepStochLog\cite{winters2021deepstochlog} & P & FOL & M + P & P & S & LpN & DS + SS\\ \hline
        
        DiffLog\cite{si2019difflog} & P & R & M + F & P+S & S & Ln & KI \\ \hline
        
        DL2\cite{fisher2019training} & M & P & C + F & P & S+Sub & lN & DS + SS \\ \hline
        
        DLM\cite{marra2019integrating} & M & FOL & C + F + P & P & S & lPN & SS + KGC \\ \hline
        
        LRNN\cite{sourek2018lrnn} & P & R & M + F & P + S  & S + Sub & LN & KGC + KI \\ \hline
        
        LTN\cite{ltn2021aij} & M & FOL & C + F & P & S + Sub & lN & DS + SS\\ \hline
        
        NeuralLP\cite{Cohen_NeuralLP} & P & R & M + F & P & S & Ln & KGC + KI\\ \hline
        
        NeurASP\cite{yang2020neurasp} & P+M & FOL & S + P & P & S & LpN & DS\\ \hline
        
        NLM\cite{NLM} & P & R & M + F & P + S  & S & Ln &  KGC + KI \\ \hline
        
        NLog\cite{tsamoura2021neural} & P & R & M + P & P & S & LpN & DS \\ \hline
        
        NLProlog\cite{weber2019nlprolog} & P & R & M + P & P + S  & S + Sub & LpN & KGC + KI\\ \hline
        
        NMLN\cite{marra2019nmln} & M & FOL & C + P & P + S & S + Sub & lPN & KGC + G \\ \hline
        
        NTP\cite{rocktaschel2017ntp} & P & R & M + F & P + S  & S + Sub & Ln & KGC + KI\\ \hline
        
        RNM\cite{marra2020relational} & M & FOL & C + P & P & S & lPN & SS \\ \hline
        
        SBR\cite{diligenti2017sbr} & M & FOL & C + F & P & S+Sub & lN & DS + SS\\ \hline
        
        Scallop\cite{huang2021scallop} & P & FOL & M + P & P & S & LpN & DS\\ \hline
        
        SL\cite{xu2018semantic} & M & P & C + P & P & S & LpN & SS\\ \hline

        Slash\cite{skryagin2022slash} & P+M & FOL & S + P & P & S & LpN & DS +SS \\ \hline
        
        TensorLog\cite{cohen2017tensorlog} & P & R & M + P & P & S & LpN & DS + KGC\\ \hline
    \end{tabular}
        \caption{Logic-based NeSy frameworks according to the 6 dimensions outlined in the paper.}
    \label{tab:big_table}
\end{sidewaystable}

	\section{Proof- vs Model-theoretic View of Logic}
	\label{sec:proof_vs_model}
	
	In this paper, we focus on clausal logic as it is a standard form to which any first order logical formula  can be converted.  In clausal logic, theories are represented in terms of {clauses}. 
	More formally, a \textit{clause} is an expression of the form  $ h_1 \vee ... \vee h_k \leftarrow b_1 \wedge ... \wedge b_n$.
	The
	$h_k$ are  \textit{head literals} or conclusions, while  the $b_i$ are  \textit{body literals} or conditions. Clauses with no conditions ($n=0$) and one conclusion ($k=1$) are \textit{facts}.
	Clauses with only one conclusion ($k=1$) are \textit{definite clauses}.
	
	The question we want to answer in this section is how to use such clausal theories to reason? And, how to infer new facts from the known clauses?  Along this first dimension, we will investigate the two fundamental ways to view logical inference and determine the implications for StarAI and NeSy systems.  In one view, we want to find \textit{proofs} for a certain query, which leads to the proof-theoretic approach to logic. In the other view, we want to find  \textit{models} (that is, truth assignments to the logical atoms) that  satisfy a given theory. This leads to the model-theoretic approach to logic.

	\paragraph{\rev{Proof-theoretic logic}}
	The proof-theoretic approach finds proofs for a query in a logic theory. While this approach to inference is applicable to any logic theory, we focus on logic programs in this paper. Syntactically, a logic program is a \textit{definite} clause theory, which is a theory where all the clauses are definite (i.e. only one conclusion).
	In logic programs, definite clauses are interpreted as if-then \textbf{rules} ($h$ is true if $b_1, ..., b_n$ are true). 

    \rev{A proof for a query $q$ is a sequence of logical inference steps that demonstrates the truth  of a query based on the given program. A compact way of representing the set of all proofs in a logic program uses an AND/OR tree, which consists of AND and OR nodes and edges amongst them. Each node represents a goal. An AND node branches into one or more outgoing edges, each representing one of the sub-goals that need to be simultaneously satisfied for the goal in the AND node to be true. An OR node represents  choices or alternatives between multiple clauses that can be used to prove a particular sub-goal. The OR node branches into multiple outgoing edges, each representing one of these possible choices. Leaf nodes in the AND/OR tree represent true facts.}
	Typically, forward or backward chaining inference is used  to search for proofs for queries. 
	We illustrate this in Example \ref{ex:logic_program}.

	
	\begin{boxedexample}[label=ex:logic_program]{Logic programs and proofs}
		
		Consider the following logic program (adapted from \cite{pearl1988probabilistic}):

		\begin{lstlisting}
		burglary.
		hears_alarm_mary. 
		
		earthquake.
		hears_alarm_john. 
		
		alarm <- earthquake. 
		alarm <- burglary.
		calls_mary <- alarm,hears_alarm_mary.
		calls_john <- alarm,hears_alarm_john.  
		\end{lstlisting}
		
        The rules for \textit{alarm} state that there will be an \textit{alarm} if there is a \textit{burglary} or an \textit{earthquake}.

		

\noindent The set of proofs for the query \textit{calls\_mary} can be represented compactly as an AND/OR tree.
    \tikzset{every picture/.style={line width=0.75pt}} 

\begin{tikzpicture}[x=0.75pt,y=0.75pt,yscale=-1,xscale=1]

\draw   (325,5.8) .. controls (325,3.7) and (326.7,2) .. (328.8,2) -- (360.2,2) .. controls (362.3,2) and (364,3.7) .. (364,5.8) -- (364,17.2) .. controls (364,19.3) and (362.3,21) .. (360.2,21) -- (328.8,21) .. controls (326.7,21) and (325,19.3) .. (325,17.2) -- cycle ;
\draw   (243,50) .. controls (243,47.79) and (244.79,46) .. (247,46) -- (278,46) .. controls (280.21,46) and (282,47.79) .. (282,50) -- (282,62) .. controls (282,64.21) and (280.21,66) .. (278,66) -- (247,66) .. controls (244.79,66) and (243,64.21) .. (243,62) -- cycle ;
\draw    (343,27) -- (262,41) ;
\draw    (423,40) -- (343,27) ;
\draw    (263,72) -- (182,86) ;
\draw    (343,85) -- (263,72) ;

\draw (239,2) node [anchor=north west][inner sep=0.75pt]   [align=left] {\texttt{(calls\_mary)}};
\draw (344.5,11.5) node   [align=left] {AND};
\draw (190,47) node [anchor=north west][inner sep=0.75pt]   [align=left] {\texttt{(alarm)}};
\draw (372,44) node [anchor=north west][inner sep=0.75pt]   [align=left] {\texttt{hears\_alarm\_mary}};
\draw (262.5,56) node   [align=left] {OR};
\draw (154,88) node [anchor=north west][inner sep=0.75pt]   [align=left] {\texttt{burglary}};
\draw (307,89) node [anchor=north west][inner sep=0.75pt]   [align=left] {\texttt{earthquake}};

\end{tikzpicture}
\end{boxedexample}

	\paragraph{\rev{Model-theoretic logic}} On the other hand, the model theoretic perspective on logic is to find a model or truth assignment to the logical atoms that satisfy a given logic theory. 
	An \textit{interpretation}, or possible world, is a truth-assignment to the propositions (or ground atoms) of the language, and
	can be uniquely identified with the set of propositions it assigns $True$ (thus considering all the other \textit{False}).  
	An interpretation is a \textit{model} of a clause $ h_1 \vee ... \vee h_k \leftarrow b_1 \wedge ... \wedge b_n$  if at least one of the $h_i$ is in the interpretation when all the $b_1 \wedge ... \wedge b_n$ are in the interpretation as well. An interpretation $I$  is a model of a theory $T$, and we write $I \models T$, if it is a model of all clauses in the theory.  We say that the theory is \textit{satisfiable} if it has a model. The satisfiability problem, that is, deciding whether a theory has a model, is one of the most fundamental ones in computer science (cf. the SAT problem for propositional logic). 
	

	\begin{boxedexample}[label = ex:model_based]{Model-theoretic}

		Consider the following theory (adopted from \cite{richardson2006mln}): 
		\begin{lstlisting}
		smokes_mary $\leftarrow$ smokes_john, influences_john_mary.
		smokes_john $\leftarrow$ smokes_mary, influences_mary_john.
		
		smokes_mary $\leftarrow$ stress_mary.
		smokes_john $\leftarrow$ stress_john.
		\end{lstlisting}
		
		A model of the previous theory is the set:
		$$M = \{\mathtt{stress\_john, smokes\_john}\}$$ 
		By considering  all the elements of this set \textit{True} and all  others \textit{False}, the four clauses are satisfied.
	\end{boxedexample}

	In the model-theoretic perspective, one uses the logic theory as a set of  
	{\em constraints} on the propositions, that is,  the propositions are related to one another, without imposing a directed inference relationship between them as in forward or backward chaining. More details on these connections can be found in \cite{DeRaedtKerstingEtAl16,flach:simplylogical}.

	\subsection{Implications for StarAI}
	

	



	Statistical Relational AI's  focus is on unifying logical and probabilistic graphical models (PGMs). A PGM \cite{koller2009probabilistic} is a graphical model that compactly
	represents a (joint) probability distribution $P(X_1, ... , X_n)$ over $n$ discrete or continuous random variables $X_1, ... ,X_n$.
	The key idea is that the joint factorizes over some factors $f_i$ specified over subsets $X^i$ of the variables $\{X_1, ... ,X_n\}$.
	
	$$P(X_1, ... , X_n) = \frac{1}{Z} f_1(X^1) \times ... \times f_k(X^k) $$
	The random variables correspond to the nodes in the graphical structure, and the factorization is determined
	by the edges in the graph.

	There are two classes of graphical models: \textit{directed} ones, or Bayesian networks, and \textit{undirected} ones, or Markov Networks.
	In Bayesian networks, the underlying graph structure is a directed acyclic graph,
	and the factors $f^i(X_i | parents(X_i))$ correspond to the conditional probabilities $P(X_i | parents(X_i))$, where $parents(X_i)$
	denotes the set of random variables that are a parent of $X_i$ in the graph.
	In Markov networks, the graph is undirected and the factors  $f^i(X^i)$  correspond to the set of nodes $X^i$ that form (maximal) cliques in the graph. Furthermore, the factors are non-negative and $Z$ is a normalisation constant.
	
	\begin{evidencebox}
		\rev{The distinction between directed and undirected graphical models is parallel to the proof- vs model-theoretic view of logic. This parallel is at the very core of  StarAI. In fact, by viewing each variable $X_i$  (or proposition) \textit{at the same time} as a random  and as a logical variable \cite{sato1995distributionsemantics}, clausal theories can be extended to define probabilistic models. Clauses can then be translated into binary valued factors by labeling them with  weights (or probabilities), thus parameterizing the corresponding factors.}
	\end{evidencebox}
	In the remainder of this section, we will show how  StarAI  has used  this parallel to define two  types of systems \cite{DeRaedtKerstingEtAl16}.

	The first type of StarAI system generalizes directed models and resembles Bayesian networks. It includes well-known representations such as plate notation \cite{koller2009probabilistic}, probabilistic relational models (PRMs) \cite{friedman1999learning}, probabilistic logic programs (PLPs) \cite{de2015probabilistic},  and Bayesian logic programs (BLPs) \cite{kersting20071}. 
	Today the most typical and popular representatives of this category are the probabilistic (logic) programs. 
	
	Probabilistic logic programs were introduced by 
	Poole~\cite{poole1993} and the first learning algorithm is due to Sato~\cite{sato1995distributionsemantics}. 
	Probabilistic logic programs are essentially definite clause programs where every fact is annotated with the probability that it is \textit{True}. 
	This then results in a possible world semantics.
	The reason why probabilistic logic programs are viewed as directed models is clear when looking at the derivations
	for a query, cf. Example \ref{ex:logic_program}. At the top of the AND-OR tree, there is the query that one wants to prove and the structure of the tree
	is that of a directed graph (even though it need not be acyclic).  One can straightforwardly map
	directed graphical models, that is, Bayesian networks,
	onto such probabilistic logic programs by associating one definite clause to every entry in the conditional probability tables,
	yielding factors of the form $P(X | Y_1, ... , Y_n)$. Assuming boolean random variables, each entry ${x,y_1, ...,y_n}$ with parameter value $v$ can be represented using the definite clause $X(x) \leftarrow Y_1 (y_1) \wedge ... \wedge Y_n(y_n) \wedge p_{x,y_1, ...,y_n}$ and 
	probabilistic fact $v::p_{x,y_1, ...,y_n}$. 
	A probabilistic version of Example~\ref{ex:logic_program} is shown in Example~\ref{ex:problog} using the syntax of ProbLog \cite{raedt:problog}.

	\begin{figure}[t]
		\centering
		\begin{tikzpicture}[x=0.75pt,y=0.75pt,yscale=-1,xscale=1]

\draw   (160,105) .. controls (160,96.72) and (177.91,90) .. (200,90) .. controls (222.09,90) and (240,96.72) .. (240,105) .. controls (240,113.28) and (222.09,120) .. (200,120) .. controls (177.91,120) and (160,113.28) .. (160,105) -- cycle ;

\draw   (110,40) .. controls (110,28.95) and (125.67,20) .. (145,20) .. controls (164.33,20) and (180,28.95) .. (180,40) .. controls (180,51.05) and (164.33,60) .. (145,60) .. controls (125.67,60) and (110,51.05) .. (110,40) -- cycle ;

\draw   (220,40) .. controls (220,28.95) and (239.7,20) .. (264,20) .. controls (288.3,20) and (308,28.95) .. (308,40) .. controls (308,51.05) and (288.3,60) .. (264,60) .. controls (239.7,60) and (220,51.05) .. (220,40) -- cycle ;

\draw   (271.04,105) .. controls (271.04,96.72) and (299.91,90) .. (335.52,90) .. controls (371.13,90) and (400,96.72) .. (400,105) .. controls (400,113.28) and (371.13,120) .. (335.52,120) .. controls (299.91,120) and (271.04,113.28) .. (271.04,105) -- cycle ;

\draw   (100,175) .. controls (100,166.72) and (120.15,160) .. (145,160) .. controls (169.85,160) and (190,166.72) .. (190,175) .. controls (190,183.28) and (169.85,190) .. (145,190) .. controls (120.15,190) and (100,183.28) .. (100,175) -- cycle ;

\draw    (200,120) -- (252.57,158.24) ;
\draw [shift={(255,160)}, rotate = 216.03] [fill={rgb, 255:red, 0; green, 0; blue, 0 }  ][line width=0.08]  [draw opacity=0] (8.93,-4.29) -- (0,0) -- (8.93,4.29) -- cycle    ;
\draw    (200,120) -- (147.43,158.24) ;
\draw [shift={(145,160)}, rotate = 323.97] [fill={rgb, 255:red, 0; green, 0; blue, 0 }  ][line width=0.08]  [draw opacity=0] (8.93,-4.29) -- (0,0) -- (8.93,4.29) -- cycle    ;
\draw    (335.52,120) -- (257.69,158.67) ;
\draw [shift={(255,160)}, rotate = 333.58000000000004] [fill={rgb, 255:red, 0; green, 0; blue, 0 }  ][line width=0.08]  [draw opacity=0] (8.93,-4.29) -- (0,0) -- (8.93,4.29) -- cycle    ;
\draw    (60,120) -- (142.29,158.72) ;
\draw [shift={(145,160)}, rotate = 205.2] [fill={rgb, 255:red, 0; green, 0; blue, 0 }  ][line width=0.08]  [draw opacity=0] (8.93,-4.29) -- (0,0) -- (8.93,4.29) -- cycle    ;
\draw    (260,60) -- (202.74,85.77) ;
\draw [shift={(200,87)}, rotate = 335.77] [fill={rgb, 255:red, 0; green, 0; blue, 0 }  ][line width=0.08]  [draw opacity=0] (8.93,-4.29) -- (0,0) -- (8.93,4.29) -- cycle    ;
\draw    (150,60) -- (197.36,85.57) ;
\draw [shift={(200,87)}, rotate = 208.37] [fill={rgb, 255:red, 0; green, 0; blue, 0 }  ][line width=0.08]  [draw opacity=0] (8.93,-4.29) -- (0,0) -- (8.93,4.29) -- cycle    ;
\draw   (1.04,105) .. controls (1.04,96.72) and (29.91,90) .. (65.52,90) .. controls (101.13,90) and (130,96.72) .. (130,105) .. controls (130,113.28) and (101.13,120) .. (65.52,120) .. controls (29.91,120) and (1.04,113.28) .. (1.04,105) -- cycle ;

\draw   (210,175) .. controls (210,166.72) and (230.15,160) .. (255,160) .. controls (279.85,160) and (300,166.72) .. (300,175) .. controls (300,183.28) and (279.85,190) .. (255,190) .. controls (230.15,190) and (210,183.28) .. (210,175) -- cycle ;

\draw (119,31) node [anchor=north west][inner sep=0.75pt]   [align=left] {\texttt{burglary}};
\draw (228,31) node [anchor=north west][inner sep=0.75pt]   [align=left] {\texttt{earthquake}};
\draw (181.14,95.38) node [anchor=north west][inner sep=0.75pt]   [align=left] {\texttt{alarm}};
\draw (276,96.13) node [anchor=north west][inner sep=0.75pt]   [align=left] {\texttt{ hears\_alarm\_mary}};
\draw (106.92,166.13) node [anchor=north west][inner sep=0.75pt]   [align=left] {\texttt{calls\_john}};
\draw (13,96.13) node [anchor=north west][inner sep=0.75pt]   [align=left] {\texttt{hears\_alarm\_john}};
\draw (216.92,166.13) node [anchor=north west][inner sep=0.75pt]   [align=left] {\texttt{calls\_mary}};

\end{tikzpicture}
		\caption{The Bayesian network corresponding to the ProbLog program in Example \ref{ex:problog}}
		\label{fig:bayes_net_problog}
	\end{figure}
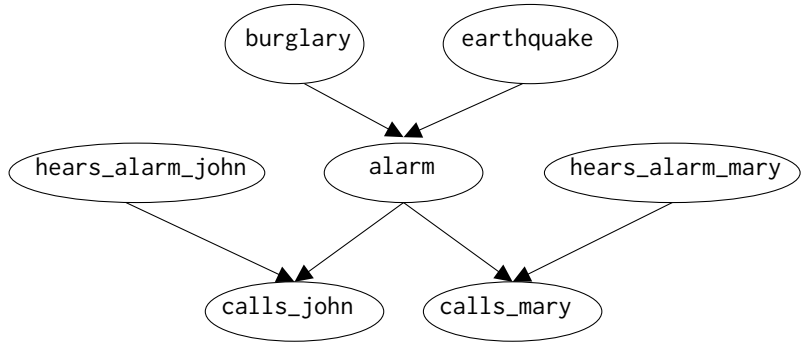
	
	\begin{boxedexample}
 [label = ex:problog]{ProbLog}

		We show a probabilistic extension of the alarm program using ProbLog.

		\begin{lstlisting}
		0.1::burglary.
		0.3::hears_alarm_mary. 
		0.05::earthquake.
		0.6::hears_alarm_john.
		
		alarm <- earthquake. 
		alarm <- burglary.
		calls_mary <- alarm,hears_alarm_mary.
		calls_john <- alarm,hears_alarm_john.  
		\end{lstlisting}
		
		This program can be mapped to the Bayesian network in Figure~\ref{fig:bayes_net_problog}
		
		This probabilistic logic program defines a distribution $p$ over possible worlds $\omega$. Let $P$ be a Problog program and $F = \{p_1::c_1, \cdots, p_n::c_n\}$ be the set of ground probabilistic facts $c_i$ of the program and $p_i$ their corresponding probabilities.
		ProbLog defines a probability distribution over $\omega$ in the following way:
		
		\begin{equation*}
		p(\omega) = \begin{cases}  0, &\mbox{if } \omega \not\models P \\
		\displaystyle \prod_{c_i \in \omega: c_i = T} p_i  \cdot \prod_{c_j \in \omega: c_j = F} (1 - p_j), & \mbox{if } \omega \models P  \end{cases}
		\end{equation*}
	\end{boxedexample}

	The second type of StarAI system generalizes undirected graphical models such as Markov networks or random fields.
	The prototypical example is  Markov Logic Networks (MLNs) \cite{richardson2006mln}, and also Probabilistic Soft Logic (PSL) \cite{bach2017psl} follows this idea.
	
	Undirected StarAI models consist of  a set of weighted clauses $w:h_1 \vee ... \vee h_k \leftarrow  b_1 \wedge  ... \wedge b_m$
  that become soft constraints. The higher the weight of a ground clause, the less likely  possible worlds that violate these constraints are. In the limit, when the weight is $+\infty$ the constraint must be satisfied and becomes a purely logical constraint, a hard constraint.
	The weighted clauses specify a more general relationship between the conclusion and the condition than the definite clauses of directed models. 
	While clauses of undirected models can still be used in (resolution) theorem provers, they are commonly viewed as constraints that relate these two sets of atoms.
	
	
	Such undirected  StarAI models can be mapped to an undirected probabilistic graphical model in which there is a one-to-one correspondence between grounded weighted clauses and factors, as we show in Example \ref{ex:mln}.

	\begin{figure}[t]
		\centering
		\tikzset{every picture/.style={line width=0.75pt}} 

\begin{tikzpicture}[x=0.75pt,y=0.75pt,yscale=-1,xscale=1]

\draw   (279,162.11) .. controls (279,152.22) and (310.62,144.21) .. (345.5,144.21) .. controls (380.39,144.21) and (411,152.22) .. (411,162.11) .. controls (411,171.99) and (380.39,180) .. (345.5,180) .. controls (310.62,180) and (279,171.99) .. (279,162.11) -- cycle ;

\draw   (138,27.89) .. controls (138,18.01) and (171.67,10) .. (206.55,10) .. controls (241.43,10) and (273,18.01) .. (273,27.89) .. controls (273,37.78) and (241.43,45.79) .. (206.55,45.79) .. controls (171.67,45.79) and (138,37.78) .. (138,27.89) -- cycle ;

\draw   (-6,162.11) .. controls (-6,152.22) and (26.39,144.21) .. (61.28,144.21) .. controls (96.16,144.21) and (128,152.22) .. (128,162.11) .. controls (128,171.99) and (96.16,180) .. (61.28,180) .. controls (26.39,180) and (-6,171.99) .. (-6,162.11) -- cycle ;

\draw   (138,162.11) .. controls (138,152.22) and (171.67,144.21) .. (206.55,144.21) .. controls (241.43,144.21) and (273,152.22) .. (273,162.11) .. controls (273,171.99) and (241.43,180) .. (206.55,180) .. controls (171.67,180) and (138,171.99) .. (138,162.11) -- cycle ;

\draw   (219.18,99.47) .. controls (219.18,89.59) and (238.98,81.58) .. (263.39,81.58) .. controls (287.81,81.58) and (307.61,89.59) .. (307.61,99.47) .. controls (307.61,109.36) and (287.81,117.37) .. (263.39,117.37) .. controls (238.98,117.37) and (219.18,109.36) .. (219.18,99.47) -- cycle ;

\draw   (99.17,99.47) .. controls (99.17,89.59) and (118.97,81.58) .. (143.39,81.58) .. controls (167.81,81.58) and (187.6,89.59) .. (187.6,99.47) .. controls (187.6,109.36) and (167.81,117.37) .. (143.39,117.37) .. controls (118.97,117.37) and (99.17,109.36) .. (99.17,99.47) -- cycle ;

\draw   (17.06,27.89) .. controls (17.06,18.01) and (36.86,10) .. (61.28,10) .. controls (85.7,10) and (105.49,18.01) .. (105.49,27.89) .. controls (105.49,37.78) and (85.7,45.79) .. (61.28,45.79) .. controls (36.86,45.79) and (17.06,37.78) .. (17.06,27.89) -- cycle ;

\draw   (301.35,27.89) .. controls (301.35,18.01) and (321.14,10) .. (345.56,10) .. controls (369.98,10) and (389.77,18.01) .. (389.77,27.89) .. controls (389.77,37.78) and (369.98,45.79) .. (345.56,45.79) .. controls (321.14,45.79) and (301.35,37.78) .. (301.35,27.89) -- cycle ;

\draw    (99.17,99.47) -- (61.28,144.21) ;
\draw    (61.28,45.79) -- (99.17,99.47) ;
\draw    (345.5,45.79) -- (307.61,99.47) ;
\draw    (345.5,144.21) -- (307.61,99.47) ;
\draw    (206.55,45.79) -- (143.39,81.58) ;
\draw    (206.55,45.79) -- (263.39,81.58) ;
\draw    (143.39,117.37) -- (206.55,144.21) ;
\draw    (263.39,117.37) -- (206.55,144.21) ;
\draw    (187.6,99.47) -- (219.18,99.47) ;

\draw (270.77,155.26) node [anchor=north west][inner sep=0.75pt]   [align=left] {\begin{minipage}[lt]{109.82pt}\setlength\topsep{0pt}
\begin{center}
{\small \tt  influences(mary,mary)}
\end{center}

\end{minipage}};
\draw (131.81,21.05) node [anchor=north west][inner sep=0.75pt]   [align=left] {\begin{minipage}[lt]{109.82pt}\setlength\topsep{0pt}
\begin{center}
{\small \tt  influences(mary,john)}
\end{center}

\end{minipage}};
\draw (-13.46,155.26) node [anchor=north west][inner sep=0.75pt]   [align=left] {\begin{minipage}[lt]{109.82pt}\setlength\topsep{0pt}
\begin{center}
{\small \tt  influences(john,john)}
\end{center}

\end{minipage}};
\draw (131.81,155.26) node [anchor=north west][inner sep=0.75pt]   [align=left] {\begin{minipage}[lt]{109.82pt}\setlength\topsep{0pt}
\begin{center}
{\small \tt  influences(john,mary)}
\end{center}

\end{minipage}};
\draw (215.21,91.42) node [anchor=north west][inner sep=0.75pt]   [align=left] {\begin{minipage}[lt]{66.98pt}\setlength\topsep{0pt}
\begin{center}
{\small \tt  smokes(mary)}
\end{center}

\end{minipage}};
\draw (95.2,91.42) node [anchor=north west][inner sep=0.75pt]   [align=left] {\begin{minipage}[lt]{66.98pt}\setlength\topsep{0pt}
\begin{center}
{\small \tt  smokes(john)}
\end{center}

\end{minipage}};
\draw (13.09,19.84) node [anchor=north west][inner sep=0.75pt]   [align=left] {\begin{minipage}[lt]{66.98pt}\setlength\topsep{0pt}
\begin{center}
{\small \tt  stress(john)}
\end{center}

\end{minipage}};
\draw (297.37,19.84) node [anchor=north west][inner sep=0.75pt]   [align=left] {\begin{minipage}[lt]{66.98pt}\setlength\topsep{0pt}
\begin{center}
{\small \tt  stress(mary)}
\end{center}

\end{minipage}};

\end{tikzpicture}
		\caption{The Markov Field corresponding to the Markov logic network in Example \ref{ex:mln}}
		\label{fig:mln}
	\end{figure}
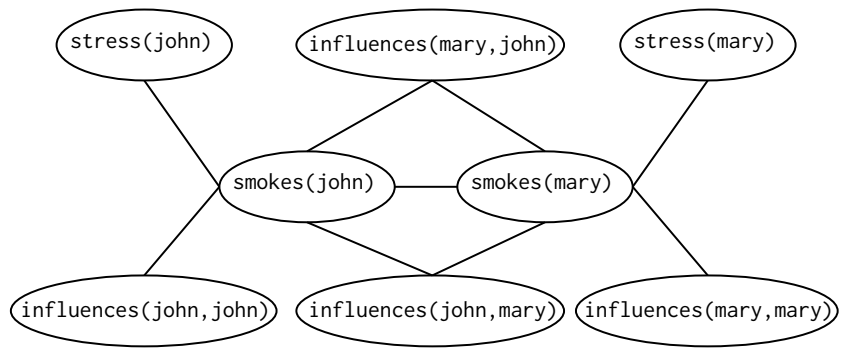
	
	\begin{boxedexample}[label = ex:mln]{Markov Logic Networks}
		
		We show a probabilistic extension (adapted from \cite{richardson2006mln}) of the theory in Example~\ref{ex:model_based} using the formalism of Markov Logic Networks. We use a First Order language with domain $D = \{\textit{john},\textit{mary}\}$ and weighted clauses $\alpha_1$ and $\alpha_2$, i.e.:
		\begin{align*}
		\alpha_1: \quad &\mathtt{
			2.0::smokes(Y) \leftarrow smokes(X), influences(X,Y)} \\
		\alpha_2: \quad &\mathtt{0.5::smokes(X) \leftarrow  stress(X)}
		\end{align*}
		
		In Figure~\ref{fig:mln}, we show the corresponding Markov field.
		
		A Markov Logic Network defines a probability distribution over possible worlds as follows.
		Let $ A = [\alpha_1, \cdots, \alpha_n]$ be a set of logical clauses and let $B = [\beta_1, \cdots, \beta_n]$ the corresponding positive weights. Let $\theta_j$ be a possible assignment of constants (from the domain $D$) to the variables (e.g. $\mathtt{X,Y}$) of the clause $\alpha_i$, that is, a substitution. Let  $\alpha_i\theta_j$ the grounded clause where all variables in $\theta_j$ have been replaced by their corresponding constants. Finally, let $\mathbbm{1}(\omega,\alpha_i\theta_i)$ be an indicator function, evaluating to 1 if the ground clause is \textit{True} in $\omega$, 0 otherwise. 
		
		The probabilistic semantics of Markov Logic is the distribution (with $Z$ the normalization constant):
		
		\begin{equation*}
		p(\omega) = \frac{1}{Z} \exp \big(\sum_i \beta_i \sum_{j} \mathbbm{1}(\omega, \alpha_i\theta_j) \big)
		\end{equation*}
		
		Intuitively, in MLNs, a world is more probable if it makes many of its ground instances \textit{True}. Notice that MLNs are usually defined on first-order clause theories, with variables and domains. We will further investigate this issue in Section \ref{sec:syntax}.

		
		
	\end{boxedexample}

	\subsection{Implications for NeSy}
	
	\rev{The distinction between proof vs models and inference rules vs constraints, turns out to be fundamental for neurosymbolic systems as well.}
	
	\begin{evidencebox}
		\rev{In neurosymbolic AI, weighted clauses are not used to construct a probabilistic graphical model, but they are likewise used to construct neural models. More specifically, NeSy systems that exploit a proof-theoretic approach use the proofs to build the \textit{architecture} of the neural net. On the other side of the spectrum, NeSy systems that exploit a model-theoretic approach use the constraints to build a \textit{loss function} for the neural net. }
	\end{evidencebox}
	
	\rev{Both choices are extremely natural.
     Proof trees capture the structure of the inference process in a graphical representation. Therefore, they can be used as the structure of the neural network computation, which corresponds to their architecture.  On the other hand, the desired behaviour of the variables is expressed in terms of constraints. 
     Loss functions are the de facto standard to enforce desired behaviours on the output variables of a neural network.}
%

\rev{First, we survey proof-based NeSy models, which use  theorem proving for  logical inference and  proofs to template the neural architecture.  In particular, when proving a specific query atom, they keep track of all the used rules in a proof tree, such as the one  shown in Example \ref{ex:logic_program}.	Weights on facts and rules are then used to label leaves or edges of the tree, respectively, while real valued activation functions are used to label the AND and OR nodes. The result is a computational graph that can be executed (or evaluated) bottom-up, starting from the leaves up to the root. Generally speaking, the output of the computational graph is a \textit{score} for the query atom. Different semantics  can be exploited in building the computational graph, ranging from relaxations of truth values (such as in fuzzy logic) to probabilities (see Section \ref{sec:semantics}). The connection between the proof tree and the neural network suggests schemes for learning the parameters of these models. Indeed, the obtained computational graph is always differentiable. Thus, given a set of atoms that are known to be \textit{True} (resp. \textit{False}), one can maximize (resp. minimize) their score using the corresponding computational graphs.
	Inference in these models is turned into \textit{evaluation} of the computational graph. The direction of the rules indicates the direction of the evaluation, in the same way as it indicates the direction of inference in logic programming. 
	Among this category are} systems based on Prolog or Datalog, such as  TensorLog \cite{cohen2017tensorlog}, Neural Theorem Provers (NTPs) \cite{rocktaschel2017ntp}, NLProlog \cite{weber2019nlprolog}, DeepProbLog \cite{manhaeve2018deepproblog}, NLog \cite{tsamoura2021neural} and DiffLog \cite{si2019difflog}.
	Lifted Relational Neural Networks (LRNNs) \cite{sourek2018lrnn} and $\partial$ILP \cite{evans:dilp} are other examples of non-probabilistic directed models, where weighted definite clauses are compiled into a neural network architecture in a forward chaining fashion.
	The systems that imitate logical reasoning with tensor calculus, Neural Logic Programming (NeuralLP) \cite{Cohen_NeuralLP} and Neural Logic Machines (NLM) \cite{NLM},  are likewise instances of directed logic. An example of a proof-based NeSy model is given in Example \ref{ex:kbann}.

	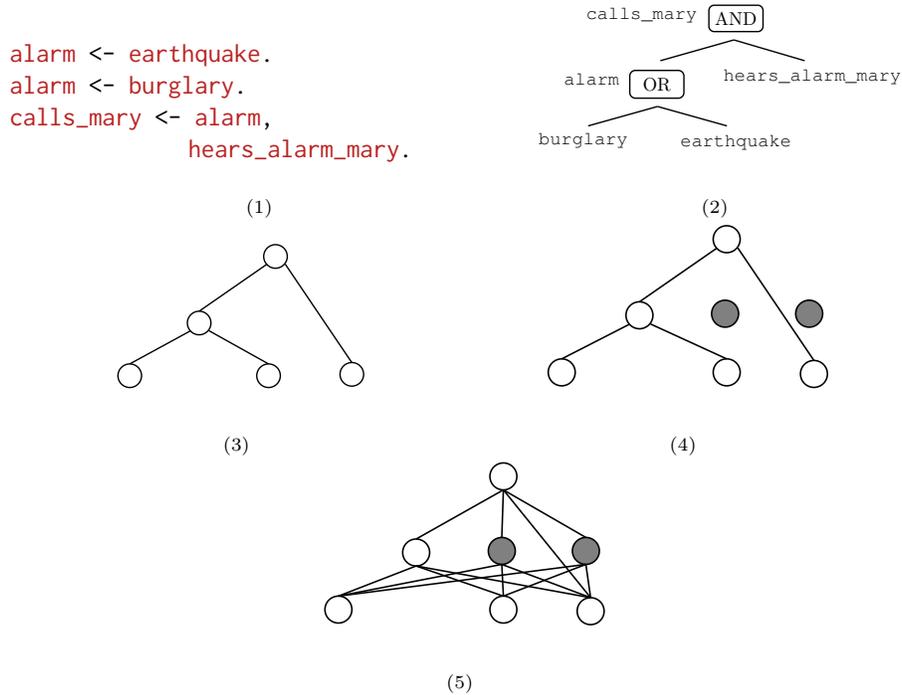
\begin{figure}[t]
		\centering
		\begin{subfigure}[b]{0.55\textwidth}
			\centering
			\begin{prolog}
alarm <- earthquake. 
alarm <- burglary.
calls_mary <- alarm, 
               hears_alarm_mary.
			\end{prolog}
			\caption{}
		\end{subfigure}
		\begin{subfigure}[b]{0.43\textwidth}
			\centering
			\scalebox{0.7}{
\tikzset{every picture/.style={line width=0.75pt}} 

\begin{tikzpicture}[x=0.75pt,y=0.75pt,yscale=-1,xscale=1]

\draw   (292,74.8) .. controls (292,72.7) and (293.7,71) .. (295.8,71) -- (327.2,71) .. controls (329.3,71) and (331,72.7) .. (331,74.8) -- (331,86.2) .. controls (331,88.3) and (329.3,90) .. (327.2,90) -- (295.8,90) .. controls (293.7,90) and (292,88.3) .. (292,86.2) -- cycle ;
\draw   (235,122) .. controls (235,119.79) and (236.79,118) .. (239,118) -- (270,118) .. controls (272.21,118) and (274,119.79) .. (274,122) -- (274,134) .. controls (274,136.21) and (272.21,138) .. (270,138) -- (239,138) .. controls (236.79,138) and (235,136.21) .. (235,134) -- cycle ;
\draw    (310,96) -- (257,111) ;
\draw    (361,111) -- (310,96) ;
\draw    (255,144) -- (205,158) ;
\draw    (305,158) -- (255,144) ;

\draw (202,71) node [anchor=north west][inner sep=0.75pt]   [align=left] {{\fontfamily{pcr}\selectfont calls\_mary}};
\draw (311.5,80.5) node   [align=left] {AND};
\draw (186,118) node [anchor=north west][inner sep=0.75pt]   [align=left] {{\fontfamily{pcr}\selectfont alarm}};
\draw (301,115) node [anchor=north west][inner sep=0.75pt]   [align=left] {{\fontfamily{pcr}\selectfont hears\_alarm\_mary}};
\draw (254.5,128) node   [align=left] {OR};
\draw (168,161) node [anchor=north west][inner sep=0.75pt]   [align=left] {{\fontfamily{pcr}\selectfont burglary}};
\draw (270,163) node [anchor=north west][inner sep=0.75pt]   [align=left] {{\fontfamily{pcr}\selectfont earthquake}};

\end{tikzpicture}
}
			\caption{}
		\end{subfigure}
		\begin{subfigure}[b]{0.48\textwidth}
			\centering
			\scalebox{0.7}{
\tikzset{every picture/.style={line width=0.75pt}} 

\begin{tikzpicture}[x=0.75pt,y=0.75pt,yscale=-1,xscale=1]

\draw    (304,86) -- (255,120) ;
\draw    (365,157) -- (317,86) ;
\draw    (249,134) -- (205,158) ;
\draw    (305,158) -- (262,134) ;
\draw   (196.5,166.5) .. controls (196.5,161.81) and (200.31,158) .. (205,158) .. controls (209.69,158) and (213.5,161.81) .. (213.5,166.5) .. controls (213.5,171.19) and (209.69,175) .. (205,175) .. controls (200.31,175) and (196.5,171.19) .. (196.5,166.5) -- cycle ;
\draw   (296.5,166.5) .. controls (296.5,161.81) and (300.31,158) .. (305,158) .. controls (309.69,158) and (313.5,161.81) .. (313.5,166.5) .. controls (313.5,171.19) and (309.69,175) .. (305,175) .. controls (300.31,175) and (296.5,171.19) .. (296.5,166.5) -- cycle ;
\draw   (246.5,128.5) .. controls (246.5,123.81) and (250.31,120) .. (255,120) .. controls (259.69,120) and (263.5,123.81) .. (263.5,128.5) .. controls (263.5,133.19) and (259.69,137) .. (255,137) .. controls (250.31,137) and (246.5,133.19) .. (246.5,128.5) -- cycle ;
\draw   (301.5,80.5) .. controls (301.5,75.81) and (305.31,72) .. (310,72) .. controls (314.69,72) and (318.5,75.81) .. (318.5,80.5) .. controls (318.5,85.19) and (314.69,89) .. (310,89) .. controls (305.31,89) and (301.5,85.19) .. (301.5,80.5) -- cycle ;
\draw   (356.5,165.5) .. controls (356.5,160.81) and (360.31,157) .. (365,157) .. controls (369.69,157) and (373.5,160.81) .. (373.5,165.5) .. controls (373.5,170.19) and (369.69,174) .. (365,174) .. controls (360.31,174) and (356.5,170.19) .. (356.5,165.5) -- cycle ;
\end{tikzpicture}
}
			\caption{}
		\end{subfigure}
		\begin{subfigure}[b]{0.48\textwidth}
			\centering
			\scalebox{0.8}{
\tikzset{every picture/.style={line width=0.75pt}} 

\begin{tikzpicture}[x=0.75pt,y=0.75pt,yscale=-1,xscale=1]

\draw    (304,86) -- (255,120) ;
\draw    (365,157) -- (317,86) ;
\draw    (249,134) -- (206,156) ;
\draw    (310,156) -- (262,134) ;
\draw   (197.5,164.5) .. controls (197.5,159.81) and (201.31,156) .. (206,156) .. controls (210.69,156) and (214.5,159.81) .. (214.5,164.5) .. controls (214.5,169.19) and (210.69,173) .. (206,173) .. controls (201.31,173) and (197.5,169.19) .. (197.5,164.5) -- cycle ;
\draw   (301.5,164.5) .. controls (301.5,159.81) and (305.31,156) .. (310,156) .. controls (314.69,156) and (318.5,159.81) .. (318.5,164.5) .. controls (318.5,169.19) and (314.69,173) .. (310,173) .. controls (305.31,173) and (301.5,169.19) .. (301.5,164.5) -- cycle ;
\draw   (246.5,128.5) .. controls (246.5,123.81) and (250.31,120) .. (255,120) .. controls (259.69,120) and (263.5,123.81) .. (263.5,128.5) .. controls (263.5,133.19) and (259.69,137) .. (255,137) .. controls (250.31,137) and (246.5,133.19) .. (246.5,128.5) -- cycle ;
\draw   (301.5,80.5) .. controls (301.5,75.81) and (305.31,72) .. (310,72) .. controls (314.69,72) and (318.5,75.81) .. (318.5,80.5) .. controls (318.5,85.19) and (314.69,89) .. (310,89) .. controls (305.31,89) and (301.5,85.19) .. (301.5,80.5) -- cycle ;
\draw   (356.5,165.5) .. controls (356.5,160.81) and (360.31,157) .. (365,157) .. controls (369.69,157) and (373.5,160.81) .. (373.5,165.5) .. controls (373.5,170.19) and (369.69,174) .. (365,174) .. controls (360.31,174) and (356.5,170.19) .. (356.5,165.5) -- cycle ;
\draw  [fill={rgb, 255:red, 128; green, 128; blue, 128 }  ,fill opacity=1 ] (300.5,127.5) .. controls (300.5,122.81) and (304.31,119) .. (309,119) .. controls (313.69,119) and (317.5,122.81) .. (317.5,127.5) .. controls (317.5,132.19) and (313.69,136) .. (309,136) .. controls (304.31,136) and (300.5,132.19) .. (300.5,127.5) -- cycle ;
\draw  [fill={rgb, 255:red, 128; green, 128; blue, 128 }  ,fill opacity=1 ] (353.5,127.5) .. controls (353.5,122.81) and (357.31,119) .. (362,119) .. controls (366.69,119) and (370.5,122.81) .. (370.5,127.5) .. controls (370.5,132.19) and (366.69,136) .. (362,136) .. controls (357.31,136) and (353.5,132.19) .. (353.5,127.5) -- cycle ;

\end{tikzpicture}
}
			\caption{}
		\end{subfigure}
		\begin{subfigure}[b]{0.48\textwidth}
			\centering
			\scalebox{0.8}{
\tikzset{every picture/.style={line width=0.75pt}} 

\begin{tikzpicture}[x=0.75pt,y=0.75pt,yscale=-1,xscale=1]

\draw    (310,89) -- (255,120) ;
\draw    (365,157) -- (310,89) ;
\draw    (255,137) -- (206,156) ;
\draw    (310,156) -- (255,137) ;
\draw   (197.5,164.5) .. controls (197.5,159.81) and (201.31,156) .. (206,156) .. controls (210.69,156) and (214.5,159.81) .. (214.5,164.5) .. controls (214.5,169.19) and (210.69,173) .. (206,173) .. controls (201.31,173) and (197.5,169.19) .. (197.5,164.5) -- cycle ;
\draw   (301.5,164.5) .. controls (301.5,159.81) and (305.31,156) .. (310,156) .. controls (314.69,156) and (318.5,159.81) .. (318.5,164.5) .. controls (318.5,169.19) and (314.69,173) .. (310,173) .. controls (305.31,173) and (301.5,169.19) .. (301.5,164.5) -- cycle ;
\draw   (246.5,128.5) .. controls (246.5,123.81) and (250.31,120) .. (255,120) .. controls (259.69,120) and (263.5,123.81) .. (263.5,128.5) .. controls (263.5,133.19) and (259.69,137) .. (255,137) .. controls (250.31,137) and (246.5,133.19) .. (246.5,128.5) -- cycle ;
\draw   (301.5,80.5) .. controls (301.5,75.81) and (305.31,72) .. (310,72) .. controls (314.69,72) and (318.5,75.81) .. (318.5,80.5) .. controls (318.5,85.19) and (314.69,89) .. (310,89) .. controls (305.31,89) and (301.5,85.19) .. (301.5,80.5) -- cycle ;
\draw   (356.5,165.5) .. controls (356.5,160.81) and (360.31,157) .. (365,157) .. controls (369.69,157) and (373.5,160.81) .. (373.5,165.5) .. controls (373.5,170.19) and (369.69,174) .. (365,174) .. controls (360.31,174) and (356.5,170.19) .. (356.5,165.5) -- cycle ;
\draw  [fill={rgb, 255:red, 128; green, 128; blue, 128 }  ,fill opacity=1 ] (300.5,127.5) .. controls (300.5,122.81) and (304.31,119) .. (309,119) .. controls (313.69,119) and (317.5,122.81) .. (317.5,127.5) .. controls (317.5,132.19) and (313.69,136) .. (309,136) .. controls (304.31,136) and (300.5,132.19) .. (300.5,127.5) -- cycle ;
\draw  [fill={rgb, 255:red, 128; green, 128; blue, 128 }  ,fill opacity=1 ] (353.5,127.5) .. controls (353.5,122.81) and (357.31,119) .. (362,119) .. controls (366.69,119) and (370.5,122.81) .. (370.5,127.5) .. controls (370.5,132.19) and (366.69,136) .. (362,136) .. controls (357.31,136) and (353.5,132.19) .. (353.5,127.5) -- cycle ;
\draw    (206,156) -- (309,136) ;
\draw    (310,156) -- (309,136) ;
\draw    (365,157) -- (309,136) ;
\draw    (362,136) -- (310,156) ;
\draw    (206,156) -- (362,136) ;
\draw    (365,157) -- (362,136) ;
\draw    (255,137) -- (365,157) ;
\draw    (309,119) -- (310,89) ;
\draw    (362,119) -- (310,89) ;

\end{tikzpicture}
}
			\caption{}
		\end{subfigure}
		\caption{Knowledge-Based Artificial Neural Network. Network creation process. (1) the initial logic program; (2) the AND-OR tree for the query \textit{calls\_mary}; (3) mapping the tree into a neural network; (4) adding hidden neurons, (5) adding interlayer connections.}
		\label{fig:kbann}
	\end{figure}

\begin{boxedexample}[label = ex:kbann]{Knowledge-Based Artificial Neural Networks}

	\noindent Knowledge-Based Artificial Neural Networks (KBANN)\cite{kbann} is the first method to use definite clausal logic and theorem proving to template the architecture of a neural network. 
	
	\noindent KBANN turns a program into a neural network in several steps:
	\begin{enumerate}
		\item KBANN starts from a definite clause program and a set of queries.
		\item The program is turned into an AND-OR tree using the proofs for the queries.
		\item The AND-OR tree is turned into a neural network with a similar structure. Nodes are divided into layers. The weights and the biases are set such that evaluating the network returns the same outcome of querying the program.
		\item New hidden units are added. Hidden units play the role of unknown rules that need to be learned. They are initialized with zero weights; i.e. they are inactive.
		\item New links are added from each layer to the next one, obtaining the final neural network. 
	\end{enumerate}
	
	\noindent An example of this process is shown in Figure~\ref{fig:kbann}. KBANN needs some restrictions over the kind of rules. In particular, the rules are assumed to be conjunctive, non-recursive, and variable-free (or propositional). Many of these restrictions are removed by more recent systems.
	
\end{boxedexample}

\rev{We now survey the second class of NeSy systems, the  model-based ones. These systems use logic to define a loss function (usually a regularization term) for  neural networks. The networks compute scores for the set of atoms that  correspond to the output neurons. At each training step, the logic-based loss function determines the degree to which the assigned scores violate the logical theory and uses this to determine the penalty. 
Logical inference is turned into a learning problem (i.e. ``learning to satisfy'') and it is usually cast in a variational optimization scheme.\footnote{This is reminiscent of the variational approach to probabilistic inference in probabilistic graphical models and constitutes a further parallel between the fields.}
As a consequence, in constraint-based models, the neural network has to solve two tasks at the same time: solving a subsymbolic learning problem (e.g. perception) as well as approximating the logical inference process  \cite{phdthesis_giuseppe}.}
A large group of NeSy approaches, including Semantic Based Regularization (SBR) \cite{diligenti2017sbr}, Logic Tensor Networks (LTN) \cite{ltn2021aij},  Semantic Loss (SL) \cite{xu2018semantic} and DL2 \cite{fisher2019training},  exploits logical knowledge as a soft regularization constraint that favours solutions that satisfy the logical constraints. SBR and LTN compute atom (fuzzy) truth assignments as the output of the neural network and translate the provided logical formulas into a real valued regularization loss term using fuzzy logic.  SL uses marginal probabilities of the target atoms to define the regularization term and relies on  arithmetic circuits \cite{darwiche2011sdd} to evaluate it efficiently, as detailed in Example \ref{ex:sl}.  DL2 defines a numerical loss providing no specific fuzzy or probabilistic semantics, which allows for including numerical variables in the formulas (e.g. by using a logical term $x > 1.5$).  Another group of approaches, including Neural Markov Logic Networks (NMLN) \cite{marra2019nmln} and Relational Neural Machines (RNM) \cite{marra2020relational} extend MLNs,  allowing factors of exponential distributions to be implemented as neural architectures.  Finally, \cite{rocktaschel2015injecting,demeester2016lifted} compute ground atoms scores as dot products between relation and entity embeddings; implication rules are then translated into a logical loss through a continuous relaxation of the implication operator.
	
	\begin{boxedexample}[label = ex:sl]{Semantic Loss}
		
		\noindent The Semantic Loss \cite{xu2018semantic} is an example of an undirected model where (probabilistic) logic is exploited as a \textit{regularization} term in training a neural model.
		
		\noindent Let $p = [p_1,\dots,p_n]$ be a vector of probabilities for a list of propositional variables $X = [X_1,\dots,X_n]$. In particular, $p_i$ denotes the probability of variable $X_i$ being \textit{True} and corresponds to a single output of a neural net having $n$ outputs.
		Let  $\alpha$ be a logic sentence defined over $X$.
		
		\noindent Then, the \textit{semantic loss} between $\alpha$ and $p$ is:
		\begin{equation*}
		Loss(\alpha,p) \propto - \log \,\, \sum_{x \models \alpha} \,\,\, \prod_{i: x \models X_i} p_i \,\,\, \prod_{i: x \models \neg X_i}  (1-p_i).
		\end{equation*}
		
		\noindent The authors provide the intuition behind this loss: 
		
		\begin{quote}
			\textit{The semantic loss is proportional to the negative logarithm of the probability of generating a state that satisfies the constraint  when sampling values according to $p$.}
		\end{quote} 
		
		\noindent Suppose you want to solve a multi-class classification task (example adapted from \cite{xu2018semantic}), where each input example must be assigned to a single class. Then, one would like to enforce \textit{mutual exclusivity} among the classes. This can be easily done on supervised examples, by coupling a softmax activation layer with a cross entropy loss. However, there is no standard way to impose this constraint for unlabeled data, which can be useful in a semi-supervised setting.
		
		\noindent The solution provided by the Semantic Loss framework is to encode mutual exclusivity into the propositional constraint $\beta$:
		\begin{equation*}
		\beta = (X_1 \land \neg X_2 \land \neg X_3) \lor
		(\neg X_1 \land  X_2 \land \neg X_3) \lor
		(\neg X_1 \land  \neg X_2 \land  X_3)
		\end{equation*}
		
		Consider a neural network classifier with three outputs $ p =[ p_1, p_2, p_3]$.  Then, for each input example (whether labeled or unlabeled), we can build the semantic loss term:
		
		\begin{equation*}
		L(\beta,p) = p_1(1 - p_2)(1 - p_3) +
		(1 - p_1)p_2(1 - p_3) +
		(1 - p_1)(1 - p_2)p_3
		\end{equation*}
		
		\noindent It can be summed up to the standard cross-entropy term for the labeled examples.
		
		Unlike for directed methods such as KBANN (Example \ref{ex:kbann}) and TensorLog, 
     the logic is turned into a loss-function that is used during training. The function constrains the underlying probabilities, but there are no directed or causal relationships among them.
	Moreover, during inference only the probabilities $p$ are used while the logic formula $\beta$ is not used anymore. On the contrary, in KBANN, the logic is compiled into the architecture of the network and, therefore, it it is also exploited at evaluation time. 
		
	\end{boxedexample}

	\rev{To conclude, let us stress a key difference between 
 the two classes of NeSy systems w.r.t. \textit{the way they incorporate the knowledge expressed in the logical clauses}. Proof-based, directed models use logic to define the architecture of a neural symbolic network. 
 Thus, logic is part of the inference of the model and acts as a structural constraint. The designer has full control of where and how the logic is used inside the network. Thus, logical knowledge can easily be extended or modified at test-time, without the need to retrain, leading to a high degree of modularity and out-of-distribution generalization \cite{misino2022vael}. On the other hand, when logic is only encoded in an objective function, the neural net learns to (approximately) satisfy it. Therefore, the  knowledge is only latently encoded in the weights of the network, which leads to a loss of control and interpretability. However, the latter techniques are often much more scalable, especially at inference time. The balance between control and interpretability, on the one hand, and scalability, on the other hand, is an open and important research question in the \nesy{} community. }

	\section{Logic - Syntax}
	\label{sec:syntax}
	In Section \ref{sec:proof_vs_model}, we have introduced clausal logic, without paying much attention to the structure of the atoms and literals. This structure and  its consequences for StarAI and \nesy{} models are the topic of the present section. Consider the following example: 
	
	\begin{boxedexample}[label = ex:prop_clausal_logic]{Propositional Clausal logic}
		
		Consider the following set of clauses. 
		\begin{lstlisting}
		mortal_socrates $\leftarrow$ human_socrates.
		mortal_aristotle $\leftarrow$ human_aristotle.
		
		human_socrates.
		human_aristotle.
		\end{lstlisting}
	\end{boxedexample} 
	Here, the literals do not possess any internal structure. \rev{They are  {\em propositions}, which are atoms that we can only assign the value $True$ or $False$. We say that we are working in \textit{propositional logic}.  }
	
	This contrasts with
	\textit{first-order} logic in which the literals take the form  $p(t_1, ... , t_m)$, with $p$ a predicate of arity $m$ and the $t_i$  terms, that is, constants, variables, or structured terms of the form $f(t_1, ..., t_q)$, where $f$ is a functor and the $t_i$ are again terms. 
	Intuitively, constants represent objects or entities, functors represent functions, variables make abstraction of specific  objects, and predicates specify relationships amongst objects.
	The subset of first order logic where there are no functors is called {\em relational logic}.

	\begin{boxedexample}[label=ex:fol_clausal_logic]{First order clausal logic}
		
		In contrast to the previous example, we  now
		write the theory in a more compact manner using \textit{first order logic}.  By convention, constants start with a lowercase letter, while variables start with an uppercase.
		Essential is the use of the variable \textit{X}, which is implicitly universally quantified.
		\begin{lstlisting}
		mortal(X) $\leftarrow$ human(X). 
		
		human(socrates).
		human(aristotle).
		\end{lstlisting}
	\end{boxedexample}


	\rev{It is interesting to understand  the  connection between propositional, relational and first-order logic. To this end, we introduce the concept of grounding.} 
	When an expression (i.e, clause, atom or term) does not contain any variable it is called {\em ground}. 
	A substitution $\theta$ is an expression of the form $\{X_1/c_1, ..., X_k/c_k \}$ with the $X_i$ different variables, the $c_i$ terms.   Applying a substitution $\theta$ to an expression $e$ (term, atom or clause) yields the instantiated expression $e\theta$ where all variables $X_i$ in $e$ have been simultaneously replaced by their corresponding terms $c_i$ in $\theta$.  We can take for instance the clause $mortal(X) \leftarrow human(X)$ and apply the substitution $\{X/socrates\}$ to yield $mortal(socrates) \leftarrow human(socrates)$.  Grounding is the process whereby all possible substitutions that ground the clauses are applied. Notice that grounding a first order logical theory may result in an infinite set of ground clauses (when there are functors), and a polynomially larger set of clauses (when working with finite domains).
	
	\rev{Finite domains are the focus in both StarAI and \nesy{}. In such domains,} any problem expressed in first-order logic can be equivalently expressed in relational logic and any problem expressed in relational logic can likewise be expressed in propositional logic by grounding out the clauses \cite{luc:book,flach:simplylogical}.

	
	\subsection{Implications for StarAI}
	StarAI typically focus on first order logic ~\cite{raedt:problog,prism,richardson2006mln}. \rev{In Section \ref{sec:proof_vs_model}, we have seen how StarAI models can be easily interpreted in terms of  probabilistic graphical models (PGM). Here, we want to show that FOL is a powerful tool for building such  models.} 
		
	\begin{evidencebox}
		\rev{FOL allows for knowledge in the form of logical rules to be interpreted as a template for defining the  graphical models. Grounding the theory  then corresponds to \textit{unrolling} the template. At the same time, first order logic has also an important statistical and learning advantage: a FOL rule leads to parameter sharing in the model as the parameters of a single FOL rule are tied to all its groundings. Parameter sharing compresses the representation of the corresponding probabilistic model, resulting in more efficient learning and better generalization.}
	\end{evidencebox}

	\rev{These properties are reminiscent of plate notation for probabilistic graphical models, bringing logical reasoning into the picture \cite{de2015probabilistic}. In Example \ref{ex:mln}, we have used only two first order rules but we obtained a larger graphical model with six factors (see Figure \ref{fig:mln}) by grounding (i.e. unrolling) the rules over the domain. All  factors corresponding to the same rule share the same weight.}

	\subsection{Implications for  \nesy{}}
	
	\rev{NeSy  exploits the internal structure of literals, resulting in many relational and first-order systems. System exploiting propositional logic are  Semantic Loss (SL) ~\cite{xu2018semantic} and DL2~\cite{fisher2019training}.  Relational logic-based systems are  DiffLog~\cite{si2019difflog}, $\theta$ILP \cite{evans:dilp}, Lifted Relational Neural Networks (LRNN) \cite{sourek2018lrnn},  Neural Theorem Provers (NTP) \cite{rocktaschel2017ntp} 
	and NeurASP~\cite{yang2020neurasp}.
		Finally, many systems are based on  first-order logic or first-order logic programs, such as 
		DeepProbLog \cite{manhaeve2018deepproblog}, NLog \cite{tsamoura2021neural}, NLProlog \cite{weber2019nlprolog}, 
		DeepStochLog \cite{winters2021deepstochlog},
		Logic Tensor Networks \cite{ltn2021aij}, Semantic Based Regularization \cite{diligenti2017sbr}, Relational Neural Machines \cite{marra2020relational} and Logical Neural Networks \cite{riegel2020logical}.}
	
	\rev{The focus in \nesy{} on structured terms is strongly related to that in StarAI and plays a fundamental role in \nesy{}. 
	 In fact, grounding a relational or first-order theory can often be seen as unrolling either the architecture (e.g., DeepStochLog\cite{winters2021deepstochlog}, LRNN \cite{sourek2018lrnn}) or the loss function (e.g., SBR \cite{diligenti2017sbr}, LTN \cite{ltn2021aij}) of the corresponding neural model. Unrolling fixed modules over multiple elements of a complex data structure is fundamental to neural networks on sequences (recurrent nets, RNN), trees (recursive nets,  RvNN) and graphs (graph nets, GNN).  \nesy{} can be regarded as unrolling more complex logical structures, with similar benefits in terms of model capacity, modularization and generalization, and strong control due to the formal semantics.}
	
	\rev{Moreover, first-order  \nesy{} models can explicitly deal with how subsymbolic data (e.g. images or audio) are fed to the neural components of the system.
 In fact, \nesy{} systems often use subsymbolic data samples as elements of the domain of discourse. For example, the element \textit{mary} can be used to refer to an image, e.g. $mary = \smallimg{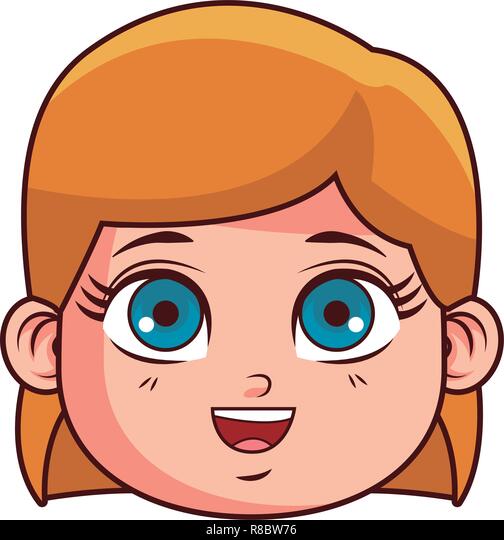}$. Feeding such samples as input to a neural network can then be naturally encoded as  grounding  a  predicate over the domain of interest. When the internal structure of the literals is absent, as in SL, this mapping must be handled outside the logical framework.}
	
While both relational logic and first-order logic have their advantages, there is a noteworthy distinction in the latter. 
First-order logic allows representing real valued functions through the use of functors. For example, segmentation can be modeled as a functor returning the bounding box of an object inside an image, e.g. \textit{location(mary,image)} \cite{desmet2023deepseaproblog}. Therefore, FOL-based systems can address regression tasks, diverging from the conventional classification tasks associated with relational logic systems.

	\section{Logic - Semantics}
	\label{sec:semantics}

\subsection{Model-theoretic  semantics}
\rev{The semantics of logical, probabilistic logical and neurosymbolic systems is defined in terms of a model theoretic semantics.}
\rev{In the present section, we will restrict our attention to Herbrand interpretations and models as is usual in logic programming and statistical relational AI (see Section \ref{sec:proof_vs_model})}.

\rev{We can distinguish three different levels of semantics, which are also closely tied to the used syntax of the underlying logic.}

\rev{First, when the logical theory consists of definite clauses only, the semantics is given by the least Herbrand model. The least Herbrand model of a definite clause theory
is unique and it is the smallest w.r.t. set inclusion. It contains all ground facts (from the Herbrand domain) that are logically entailed by the theory.
For instance, considering the facts $a$ and $b$ and the rules $d \leftarrow a, b$ and $b \leftarrow c$  would give the least Herbrand model $\{a, b, d\}$.}

\rev{Second, when the logical theory can contain any set of clauses, the semantics is given by the set of all possible Herbrand models. For instance, considering the clause $a \vee b $
yields the models $\{a\}, \{b\}$ and $\{a, b\}$. So there is not necessarily a unique model, not even when considering only  minimal models, where
we have $\{a\}, \{b\}$.} 

\rev{Third, while Horn-clauses are the basis for "pure" Prolog and logic programs, there exist several extensions to this formalism to accommodate
negated literals in the condition part of rules or disjunction in the head. A popular framework in this regard is answer set programming (ASP). 
In ASP the clause $a \vee b$ could be represented by two clauses $a \leftarrow \neg b$ and $b \leftarrow \neg a$ which would  have two stable models $\{a\}$ and $\{b\}$.}


\subsection{Fuzzy semantics}
\rev{The previous three levels of semantics are based on Boolean models, i.e. models where each atom is either present (i.e. \textit{True}) or absent (i.e. \textit{False})}. Differently, \textit{fuzzy logic}, and in particular t-norm fuzzy logic, assigns a truth value to atoms in the \textit{continuous} real interval $[0,1]$. Logical operators are then turned into real-valued functions, mathematically grounded in the t-norm theory. A \textit{t-norm} $t(x, y)$ is a real function $t:[0,1] \times [0,1] \to [0,1]$ that models the logical AND and from which the other operators can be derived. Table~\ref{tab:t-norms} shows well-known t-norms and the functions corresponding to their  connectives. A fuzzy logic formula is mapped to a real valued function of its input atoms, as we show in Example~\ref{ex:fuzzy}.  Fuzzy logic generalizes Boolean logic to continuous values. All the different t-norms are coherent with Boolean logic in the endpoints of the interval $[0,1]$, which correspond to \textit{completely true} and \textit{completely false} values. The concept of model in fuzzy logic can be easily recovered from an extension of the model-theoretic semantics of the Boolean logic. Any \textit{fuzzy} interpretation is a model of a formula if the formula evaluates to 1.

\begin{boxedexample}[label=ex:fuzzy]{Fuzzy logic}
		
		Let us consider the following propositions: \textit{alarm}, \textit{burglary} and \textit{earthquake}. Defining a fuzzy semantics for this language requires one to assign truth degrees to each of the propositions and selecting a particular t-norm to implement the connectives. 
		
		Let us consider the \L ukasiewicz t-norm and the following interpretation of the language:

		\begin{minipage}{0.40\linewidth}
			\begin{align*}
			\mathcal{I} = \{&\texttt{alarm} = 0.7,  \\
			&\texttt{burglary}=0.6, \\ &\texttt{earthquake}=0.3\}
			\end{align*}
		\end{minipage}
		\begin{minipage}{0.60\linewidth}
			\begin{align*}
			t_{\vee}(x,y)  & =  \min ( 1, x + y)\\
			t_{\rightarrow}(x,y) & = \min(1, 1 - x + y)
			\end{align*}
		\end{minipage}
		\newline
		
		Once we have defined the semantics of the language, we can evaluate logic sentences, e.g.:
		\begin{align*}
		& \texttt{alarm} \leftarrow (\texttt{burglary} \vee \texttt{earthquake}) =   \\
		& \min(1,1 - \min(1, \texttt{burglary} + \texttt{earthquake}) +\texttt{alarm}) = 0.8
		\end{align*}
		
		This evaluation can be performed automatically by parsing the logical sentence in the corresponding \textit{expression tree} and then compiling the expression tree using the corresponding t-norm operation:  
		\begin{center}
			\tikzset{every picture/.style={line width=0.75pt}} 

\begin{tikzpicture}[x=0.75pt,y=0.75pt,yscale=-1,xscale=1]

\draw   (265,8.8) .. controls (265,6.7) and (266.7,5) .. (268.8,5) -- (300.2,5) .. controls (302.3,5) and (304,6.7) .. (304,8.8) -- (304,20.2) .. controls (304,22.3) and (302.3,24) .. (300.2,24) -- (268.8,24) .. controls (266.7,24) and (265,22.3) .. (265,20.2) -- cycle ;
\draw    (316,41) -- (282,25) ;
\draw    (282,25) -- (246,42) ;
\draw   (297,45.3) .. controls (297,43.2) and (298.7,41.5) .. (300.8,41.5) -- (332.2,41.5) .. controls (334.3,41.5) and (336,43.2) .. (336,45.3) -- (336,56.7) .. controls (336,58.8) and (334.3,60.5) .. (332.2,60.5) -- (300.8,60.5) .. controls (298.7,60.5) and (297,58.8) .. (297,56.7) -- cycle ;
\draw    (348,77.5) -- (314,61.5) ;
\draw    (314,61.5) -- (278,78.5) ;
\draw   (77,8.8) .. controls (77,6.7) and (78.7,5) .. (80.8,5) -- (112.2,5) .. controls (114.3,5) and (116,6.7) .. (116,8.8) -- (116,20.2) .. controls (116,22.3) and (114.3,24) .. (112.2,24) -- (80.8,24) .. controls (78.7,24) and (77,22.3) .. (77,20.2) -- cycle ;
\draw    (128,41) -- (94,25) ;
\draw    (94,25) -- (58,42) ;
\draw   (109,45.3) .. controls (109,43.2) and (110.7,41.5) .. (112.8,41.5) -- (144.2,41.5) .. controls (146.3,41.5) and (148,43.2) .. (148,45.3) -- (148,56.7) .. controls (148,58.8) and (146.3,60.5) .. (144.2,60.5) -- (112.8,60.5) .. controls (110.7,60.5) and (109,58.8) .. (109,56.7) -- cycle ;
\draw    (160,77.5) -- (126,61.5) ;
\draw    (126,61.5) -- (90,78.5) ;

\draw (284.5,12.5) node   [align=left] {$\displaystyle t_{\leftarrow}$};
\draw (234,43) node [anchor=north west][inner sep=0.75pt]   [align=left] {0.7};
\draw (265,80) node [anchor=north west][inner sep=0.75pt]   [align=left] {0.6};
\draw (337,79) node [anchor=north west][inner sep=0.75pt]   [align=left] {0.3};
\draw (339,41) node [anchor=north west][inner sep=0.75pt]   [align=left] {(0.9)};
\draw (308,5) node [anchor=north west][inner sep=0.75pt]   [align=left] {(0.8)};
\draw (316.5,51) node   [align=left] {$\displaystyle t_{\vee}$};
\draw (96.5,12.5) node   [align=left] {$\displaystyle \leftarrow $};
\draw (29,42) node [anchor=north west][inner sep=0.75pt]   [align=left] {alarm};
\draw (50,80) node [anchor=north west][inner sep=0.75pt]   [align=left] {burglary};
\draw (147,80) node [anchor=north west][inner sep=0.75pt]   [align=left] {earthquake};
\draw (128.5,51) node   [align=left] {$\displaystyle \vee$};

\end{tikzpicture}
			
		\end{center}
		The resulting circuit represents a differentiable function and the truth degree of the sentence is computed by evaluating the circuit bottom-up.

	\end{boxedexample}

	\begin{table}[tb]
		\centering
		\begin{tabular}{c|c|c|c}
			
			& Product & \L ukasiewicz  &  G\"{o}del \\ 
			\hline
			$x \land y$ & $x \cdot y$ & $\max(0, x + y -1$) & $\min(x, y$) \\
			\hline
			$x \lor y$ & $x + y - x \cdot y$ &  $\min(1, x + y)$ & $\max(x, y)$ \\
			\hline
			$\lnot x$ & $1 - x$ &  $1 - x$ & $1 - x$ \\
			\hline
			$x \Rightarrow y \;\; (x>y)$ & $y/x$ &  $\min(1,1-x+y)$ & $y$ \\
			\hline
		\end{tabular}
		\caption{Logical connectives on the inputs $x,y$ when using the fundamental t-norms.}
		\label{tab:t-norms}
	\end{table}

\subsection{Implications for StarAI}

Statistical Relational AI has extended the previous semantics by defining probability distributions $p(\omega)$ over models, or \textit{possible worlds}\footnote{In this paper, we use the distribution semantics as representative of the probabilistic approach to logic. While this is the most common solution in StarAI, many other solutions exist \cite{muggleton1996stochastic,halpern1990analysis}, whose description is out of the scope of the current survey. A detailed overview of the different flavours of formal reasoning about uncertainty can be found in \cite{halpern2017reasoning}.}.

The goal is to reason about the uncertainty of logical statements. In particular, the probability that a certain formula $\alpha$ holds is computed as the sum of the probabilities of the possible worlds that are models of $\alpha$ (i.e. where $\alpha$ is True):
	\begin{equation}
	\label{eq:wmc}
	p(\alpha) = \sum_{\omega \models \alpha} p(\omega) 
	\end{equation}
	This is an instance of the Weighted Model Counting (WMC) problem. In fact, we are counting how many worlds are models of $\alpha$ and we are weighting each of them by its  probability according to the distribution $p(\omega)$.

	\begin{table}
		\centering
		\begin{tabular}{cccc|l}
			B & E & J & M & $p(\omega)$ \\
			\hline
			F&F&F&F& 0.2394 \\
			F&F&F&T& 0.1026 \\
			F&F&T&F& 0.3591 \\
			F&F&T&T& 0.1539 \\
			F&T&F&F& 0.0126 \\
			F&T&F&T& 0.0054 \\
			F&T&T&F& 0.0189 \\
			F&T&T&T& 0.0081 \\
			T&F&F&F& 0.0266 \\
			T&F&F&T& 0.0114 \\
			T&F&T&F& 0.0399 \\
			T&F&T&T& 0.0171  \\
			T&T&F&F& 0.0014 * \\
			T&T&F&T& 0.0006 * \\
			T&T&T&F& 0.0021 * \\
			T&T&T&T& 0.0009 * 
		\end{tabular}
		\caption{A distribution over possible worlds for the four propositional variables $burglary$ (B), $earthquake$ (E), $hears\_alarm\_john$ (J) and  $hears\_alarm\_mary$ (M). The $*$ indicates those worlds where $burglary \wedge earthquake$ is \textit{True}.}
		\label{tab:distribution_semantics}
	\end{table}
	
	\begin{boxedexample}[label = ex:distr_semantics]{Probabilistic Logic}
		
		Let us consider the following set of propositions $B = burglary$, $E = earthquake$, $J = hears\_alarm\_john$ and $M = hears\_alarm\_mary$. In probabilistic logic, a probability distribution over all the possible worlds is defined. For example, Table \ref{tab:distribution_semantics} represents a valid distribution.
		
		Suppose we want to compute the probability of the formula $burglary \wedge earthquake$. This is done by summing up the probabilities of all the worlds where both $burglary$ and $earthquake$ are \textit{True} (indicated by a $*$ in Table \ref{tab:distribution_semantics}). 
	\end{boxedexample}

    The StarAI community has provided several formalisms to define such  probability distributions over possible worlds using labeled logic theories. Probabilistic Logic Programs (cf. Example \ref{ex:problog}) and Markov logic networks (cf. Example \ref{ex:mln}) are two prototypical frameworks. For example, the distribution in Table~\ref{tab:distribution_semantics} is the one modeled by the ProbLog program in Example \ref{ex:problog}.

\rev{It is interesting to compare Markov Logic (Example \ref{ex:mln}) to ProbLog (Example \ref{ex:problog}) in terms of their model-theoretic semantics. Markov Logic is defined as a set of weighted full clauses, i.e. as an unnormalized  probability distribution over full clausal theories.
This means that, given any subset of the theory, there can be many possible models. For instance, the theory $a \vee b$, has three possible models. To obtain a probability distribution over models, Markov Logic needs to distribute the probability mass over its models. 
To do this, the maximum entropy principle is used, which results in equal distributions of the probability mass.} 
\rev{Conversely, ProbLog defines a probability distribution over definite clause theories, each obtained as subsets of the provided probabilistic facts. However, since each of these theories has a unique least Herbrand model, the probability mass corresponding to the selected facts is assigned to the corresponding unique Herbrand model. This means that when working with definite clauses only, there is no need to distribute the probability mass to multiple models and, therefore, no extra assumptions such as maximum entropy are necessary.}

	


Probabilistic inference (i.e. weighted model counting) is generally intractable. That is why, in StarAI, techniques such as \textit{knowledge compilation} (KC)~\cite{DarwicheMarquis} are used.
	Knowledge compilation transforms a logical formula $\alpha$ into a new representation in an  offline step, which can be computationally expensive. Using this new representation a particular set of queries can be answered efficiently (i.e. in poly-time in the size of the new representation). 
	From a probabilistic point of view, this translation solves the disjoint-sum problem, which states that one cannot simply sum up the probability 
 of two disjuncts but also has to subtract the probability of the intersection.
 After the translation, the probabilities of any conjunction and of any disjunction can be simply computed by multiplying, resp. summing up, the probabilities of their operands. Thus a  logical formula $\alpha$ can be compiled into an arithmetic circuit $ac(\alpha)$. The weighted model count of the query formula can then simply be computed by  evaluating  the corresponding arithmetic circuit bottom up; i.e.  $p(\alpha) = ac(\alpha)$.

	\begin{figure}[t]
		\centering
		\noindent
		\begin{subfigure}{0.4\linewidth}
			\begin{tikzpicture}[x=0.75pt,y=0.75pt,yscale=-1,xscale=1]

\draw   (291,76.3) .. controls (291,74.2) and (292.7,72.5) .. (294.8,72.5) -- (326.2,72.5) .. controls (328.3,72.5) and (330,74.2) .. (330,76.3) -- (330,87.7) .. controls (330,89.8) and (328.3,91.5) .. (326.2,91.5) -- (294.8,91.5) .. controls (292.7,91.5) and (291,89.8) .. (291,87.7) -- cycle ;
\draw    (309,97.5) -- (284,123.5) ;
\draw    (337,123.5) -- (309,97.5) ;
\draw   (265,130.3) .. controls (265,128.2) and (266.7,126.5) .. (268.8,126.5) -- (300.2,126.5) .. controls (302.3,126.5) and (304,128.2) .. (304,130.3) -- (304,141.7) .. controls (304,143.8) and (302.3,145.5) .. (300.2,145.5) -- (268.8,145.5) .. controls (266.7,145.5) and (265,143.8) .. (265,141.7) -- cycle ;
\draw    (283,151.5) -- (258,177.5) ;
\draw    (311,177.5) -- (283,151.5) ;
\draw   (230,183.3) .. controls (230,181.2) and (231.7,179.5) .. (233.8,179.5) -- (265.2,179.5) .. controls (267.3,179.5) and (269,181.2) .. (269,183.3) -- (269,194.7) .. controls (269,196.8) and (267.3,198.5) .. (265.2,198.5) -- (233.8,198.5) .. controls (231.7,198.5) and (230,196.8) .. (230,194.7) -- cycle ;
\draw    (248,204.5) -- (223,230.5) ;
\draw    (262,284.5) -- (248,204.5) ;
\draw   (294,183.3) .. controls (294,181.2) and (295.7,179.5) .. (297.8,179.5) -- (329.2,179.5) .. controls (331.3,179.5) and (333,181.2) .. (333,183.3) -- (333,194.7) .. controls (333,196.8) and (331.3,198.5) .. (329.2,198.5) -- (297.8,198.5) .. controls (295.7,198.5) and (294,196.8) .. (294,194.7) -- cycle ;
\draw    (312,204.5) -- (287,230.5) ;
\draw    (340,230.5) -- (312,204.5) ;
\draw   (269,237.3) .. controls (269,235.2) and (270.7,233.5) .. (272.8,233.5) -- (304.2,233.5) .. controls (306.3,233.5) and (308,235.2) .. (308,237.3) -- (308,248.7) .. controls (308,250.8) and (306.3,252.5) .. (304.2,252.5) -- (272.8,252.5) .. controls (270.7,252.5) and (269,250.8) .. (269,248.7) -- cycle ;
\draw    (287,258.5) -- (262,284.5) ;
\draw    (315,284.5) -- (287,258.5) ;

\draw (328,125) node [anchor=north west][inner sep=0.75pt]   [align=left] {hears\_alarm(mary)};
\draw (161,228) node [anchor=north west][inner sep=0.75pt]   [align=left] {$\displaystyle \neg $burglary};
\draw (212,287) node [anchor=north west][inner sep=0.75pt]   [align=left] {earthquake};
\draw (310.5,82) node   [align=left] {AND};
\draw (284.5,136) node   [align=left] {OR};
\draw (249.5,189) node   [align=left] {AND};
\draw (313.5,189) node   [align=left] {AND};
\draw (288.5,243) node   [align=left] {OR};
\draw (332,234) node [anchor=north west][inner sep=0.75pt]   [align=left] {burglary};
\draw (308,285) node [anchor=north west][inner sep=0.75pt]   [align=left] {$\displaystyle \neg $earthquake};

\end{tikzpicture}
		\end{subfigure}
		\hfill
		\begin{subfigure}{0.4\linewidth}
			\begin{tikzpicture}[x=0.75pt,y=0.75pt,yscale=-1,xscale=1]

\draw   (291,76.3) .. controls (291,74.2) and (292.7,72.5) .. (294.8,72.5) -- (326.2,72.5) .. controls (328.3,72.5) and (330,74.2) .. (330,76.3) -- (330,87.7) .. controls (330,89.8) and (328.3,91.5) .. (326.2,91.5) -- (294.8,91.5) .. controls (292.7,91.5) and (291,89.8) .. (291,87.7) -- cycle ;
\draw    (309,97.5) -- (284,123.5) ;
\draw    (337,123.5) -- (309,97.5) ;
\draw   (265,130.3) .. controls (265,128.2) and (266.7,126.5) .. (268.8,126.5) -- (300.2,126.5) .. controls (302.3,126.5) and (304,128.2) .. (304,130.3) -- (304,141.7) .. controls (304,143.8) and (302.3,145.5) .. (300.2,145.5) -- (268.8,145.5) .. controls (266.7,145.5) and (265,143.8) .. (265,141.7) -- cycle ;
\draw    (283,151.5) -- (258,177.5) ;
\draw    (311,177.5) -- (283,151.5) ;
\draw   (230,183.3) .. controls (230,181.2) and (231.7,179.5) .. (233.8,179.5) -- (265.2,179.5) .. controls (267.3,179.5) and (269,181.2) .. (269,183.3) -- (269,194.7) .. controls (269,196.8) and (267.3,198.5) .. (265.2,198.5) -- (233.8,198.5) .. controls (231.7,198.5) and (230,196.8) .. (230,194.7) -- cycle ;
\draw    (248,204.5) -- (223,230.5) ;
\draw    (262,284.5) -- (248,204.5) ;
\draw   (294,183.3) .. controls (294,181.2) and (295.7,179.5) .. (297.8,179.5) -- (329.2,179.5) .. controls (331.3,179.5) and (333,181.2) .. (333,183.3) -- (333,194.7) .. controls (333,196.8) and (331.3,198.5) .. (329.2,198.5) -- (297.8,198.5) .. controls (295.7,198.5) and (294,196.8) .. (294,194.7) -- cycle ;
\draw    (312,204.5) -- (287,230.5) ;
\draw    (340,230.5) -- (312,204.5) ;
\draw   (269,237.3) .. controls (269,235.2) and (270.7,233.5) .. (272.8,233.5) -- (304.2,233.5) .. controls (306.3,233.5) and (308,235.2) .. (308,237.3) -- (308,248.7) .. controls (308,250.8) and (306.3,252.5) .. (304.2,252.5) -- (272.8,252.5) .. controls (270.7,252.5) and (269,250.8) .. (269,248.7) -- cycle ;
\draw    (287,258.5) -- (262,284.5) ;
\draw    (315,284.5) -- (287,258.5) ;

\draw (328,125) node [anchor=north west][inner sep=0.75pt]   [align=left] {0.3};
\draw (198,232) node [anchor=north west][inner sep=0.75pt]   [align=left] {1 - 0.1};
\draw (240,287) node [anchor=north west][inner sep=0.75pt]   [align=left] {0.05};
\draw (310.5,82) node   [align=left] {*};
\draw (284.5,136) node   [align=left] {+};
\draw (249.5,189) node   [align=left] {*};
\draw (313.5,189) node   [align=left] {*};
\draw (288.5,243) node   [align=left] {+};
\draw (332,234) node [anchor=north west][inner sep=0.75pt]   [align=left] {0.1};
\draw (308,285) node [anchor=north west][inner sep=0.75pt]   [align=left] {1-0.05};
\draw (177,180) node [anchor=north west][inner sep=0.75pt]   [align=left] {(0.045)};
\draw (339,179) node [anchor=north west][inner sep=0.75pt]   [align=left] {(0.1)};
\draw (211,127) node [anchor=north west][inner sep=0.75pt]   [align=left] {(0.145)};
\draw (233,73) node [anchor=north west][inner sep=0.75pt]   [align=left] {(0.0435)};

\end{tikzpicture}
		\end{subfigure}
		\caption{dDNNF (left) and arithmetic circuit (right) corresponding to the ProbLog program in Example \ref{ex:problog}}
		\label{fig:kc}
	\end{figure}
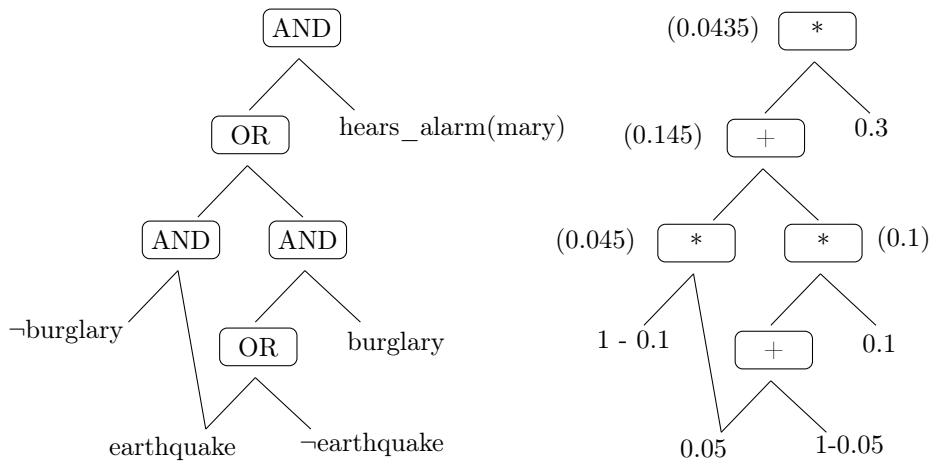

	\begin{boxedexample}[label = ex:kc]{Knowledge Compilation}

		Let us consider the ProbLog program in Example \ref{ex:problog} and the corresponding tabular representation in Table \ref{tab:distribution_semantics}. Let us consider the query $q = calls(mary)$. Now we can use Equation~\ref{eq:wmc} to compute the probability $p(q)$. To do this, we iterate over the table and we sum all the probabilities of the worlds where $calls(mary)$ is True, which we know from Example \ref{ex:logic_program} are those where either $burglary=T$ or $earthquake=T$ and where $hears\_alarm(mary)=T$. This yields $p(q) = 0.0435$. This method would require us to iterate over $2^N$ terms (where $N$ is the number of probabilistic facts).
		
		Knowledge compilation compiles $\alpha$ into some normal form that is logically equivalent. In Figure \ref{fig:kc}, the target representation is a decomposable, deterministic negative normal form (d-DNNF)~\cite{ddnnf}, for which weighted model counting is poly-time in the size of the formula. Decomposability means that, for every conjunction, the two conjuncts do not share any variables. Deterministic means that, for every disjunction, the two disjuncts are \rev{mutually exclusive, i.e., only one of the disjuncts can be true at the same time}. 
		The formula in d-DNNF can then be straightforwardly turned into an arithmetic circuit by substituting AND nodes with multiplication and OR nodes by summation. 
  In Figure \ref{fig:kc}, we show the d-DNNF and the arithmetic circuit of the distribution defined by the ProbLog program in Example \ref{ex:problog}. The bottom-up evaluation of this arithmetic circuit computes the correct marginal probability $p(\alpha)$ much more efficiently than the naive iterative sum that we have computed before.

		
	\end{boxedexample}
	
	Even though probabilistic Boolean logic is the most common choice in StarAI, some approaches use probabilistic fuzzy logic. The most prominent approach is Probabilistic Soft Logic (PSL) \cite{bach2017psl}, illustrated in Example~\ref{ex:psl}. Similarly to Markov logic networks, Probabilistic Soft Logic (PSL) defines log linear models where features are represented by ground clauses. However, PSL uses a fuzzy semantics of the logical theory. Therefore, atoms are mapped to real valued  variables and ground clauses to real valued factors.
	
	\begin{boxedexample}[label = ex:psl]{Probabilistic Soft Logic}
		
		Let us consider the logical rule $\alpha =  smokes(X) \leftarrow stress(X)$ with weight $\beta$.
		
		As we have seen in Example~\ref{ex:mln}, Markov Logic translates the formula into a discrete factor by using the indicator functions $\mathbbm{1}(\omega, \alpha\theta)$:
		
		\begin{equation*}
		\phi^{MLN}(\omega, \alpha) = \beta \mathbbm{1}(\omega, \alpha\{X/mary\})  + \beta \mathbbm{1}(\omega, \alpha\{X/john\})
		\end{equation*}
		
		Instead of discrete indicator functions, PSL \cite{bach2017psl} translates the formula into a continuous t-norm based function:
		
		\begin{equation*}
		t(\omega, \alpha) = \min(1, 1 - stress(X) + smokes(X))
		\end{equation*}
		
		and the corresponding potential is then translated into the continuous and differentiable function:
		\begin{equation*}
		\phi^{PSL}(\omega, \alpha) = \beta t(\omega, \alpha\{X/mary\})  + \beta t(\omega, \alpha\{X/john\})
		\end{equation*}

		Another important task in StarAI is MAP inference. In MAP inference, given the distribution $p(\omega)$, one is interested in finding the interpretation $\omega^\star$ where $p$ is maximal, i.e.
		
		\begin{equation}
		\label{eq:map}
		\omega^\star = \text{arg}\max_\omega p(\omega)
		\end{equation}
		
		When the $\omega$ is a boolean interpretation, i.e. $\omega \in \{0,1\}^n$, like in ProbLog or MLNs, this problem is \rev{related to maxSAT, which is NP-hard}.  However, in PSL, $\omega$ is a fuzzy interpretation, i.e. $\omega \in [0,1]^n$ and $p(\omega) \propto \exp\big(\sum_i \beta_i \phi(\omega, \alpha_i)\big)$ is a continuous and differentiable function. The MAP inference problem can thus be \textit{approximated} more efficiently than its boolean counterpart using gradient-based techniques.
	\end{boxedexample}

	\subsection{Implications for NeSy}


We have seen that in StarAI, one can turn inference tasks into the evaluation (as in KC) or gradient-based optimization (as in PSL) of a differentiable parametric circuit. The parameters are scalar values (e.g. probabilities or truth degrees) that are attached to basic elements of a logical theory (facts or clauses).
	
	A natural way of carrying over the StarAI approach to NeSy is the reparameterization method. Reparameterization substitutes the scalar values assigned to facts or formulas with the output of a neural network. One can interpret this substitution in terms of a different parameterization of the original model. Many probabilistic methods parameterize the underlying distribution in terms of neural components. In particular, as we show in Example \ref{ex:deepproblog}, DeepProbLog exploits neural predicates to compute the probabilities of probabilistic facts as the output of neural computations over vectorial representations of the constants, which is similar to SL in the propositional counterpart (see Example \ref{ex:sl}).   NeurASP also inherits the concept of a neural predicate from DeepProbLog.

\begin{figure}[t]
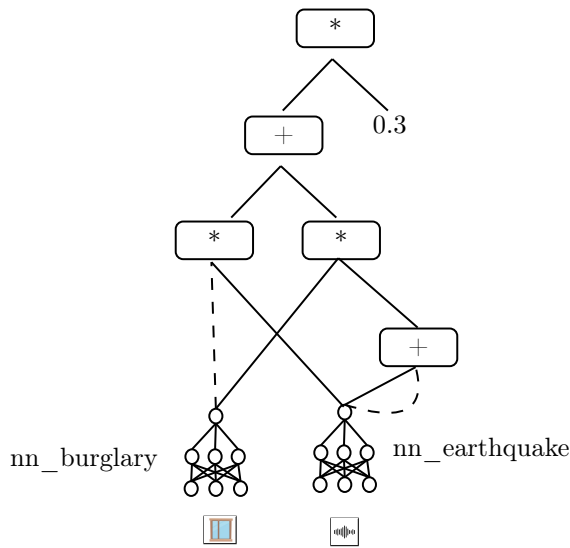

		\centering
		\include{to_include/arithmetic_circuit_dpl}
		\caption{A neural reparametrization of the arithmetic circuit in Example \ref{ex:kc} as done by DeepProbLog (cf. Example \ref{ex:deepproblog}). Dashed lines indicate a negative output, i.e 1 - x. \rev{We use a different notation for negation than in Figure \ref{fig:kc} to stress that both  leaves are parameterized by the same neural network}.}
		\label{fig:deepproblog}
	\end{figure}

	\begin{boxedexample}[label = ex:deepproblog]{Probabilistic semantics reparameterization in  DeepProbLog}
		
		DeepProbLog \cite{manhaeve2018deepproblog} is a neural extension of the probabilistic logic programming language ProbLog. DeepProbLog allows images or other subsymbolic representations as terms of the program.
		
		Let us consider a possible neural extension of the program in Example~\ref{ex:problog}. We could extend the predicate $calls(X)$ with two extra inputs, i.e. $calls(B,E,X)$. $B$ is supposed to contain an image of a security camera, while $E$ is supposed to contain the time-series of a seismic sensor. We would like to answer queries like $calls(\smallimg{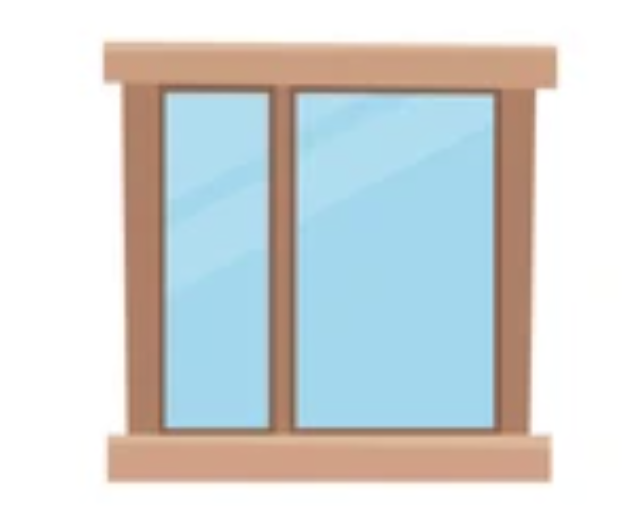},\smallimg{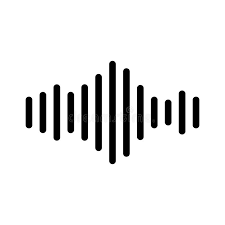},mary)$, i.e. what is the probability that $mary$ calls, given that the security camera has captured the image $\smallimg{burglary1.png}$ and the sensor the signal $\smallimg{earthquake1.png}$ .
		
		DeepProbLog can answer this query using  the following program:
		
		\begin{lstlisting}
		nn(nn_burglary, [B]) :: burglary(B).
		nn(nn_earthquake, [E]) :: earthquake(E).
		0.3::hears_alarm(mary). 
		0.6::hears_alarm(john). 
		alarm(B,_) <- burglary(B).
		alarm(_,E) <- earthquake(E).
		calls(B,E, X) <- alarm(B,E), hears_alarm(X).
		\end{lstlisting}
		
		Here, the program has been extended in two ways. First, new arguments (i.e. $B$ and $E$) have been introduced in order to deal with the subsymbolic inputs. Second, the probabilistic facts $burglary$ and $earthquake$ have been turned into \textit{neural predicates}. Neural predicates are special probabilistic facts that are annotated by neural networks instead of by scalar probabilities.
		
		Inference in DeepProbLog mimics that of ProbLog. Given the query and the program, knowledge compilation is used to build the arithmetic circuit in Figure \ref{fig:deepproblog}.

		Since the program is structurally identical to the purely symbolic one in Example \ref{ex:kc}, the arithmetic circuit is exactly the same. The only  only difference is that some leaves of the tree (i.e. capturing probabilities of  facts) can now also be neural networks.
		
		Given a set of queries that are \textit{True}, i.e.:
		\begin{align*}
		\mathcal{D} = \{&calls(\smallimg{burglary1.png},\smallimg{earthquake1.png},mary),\\  &calls(\smallimg{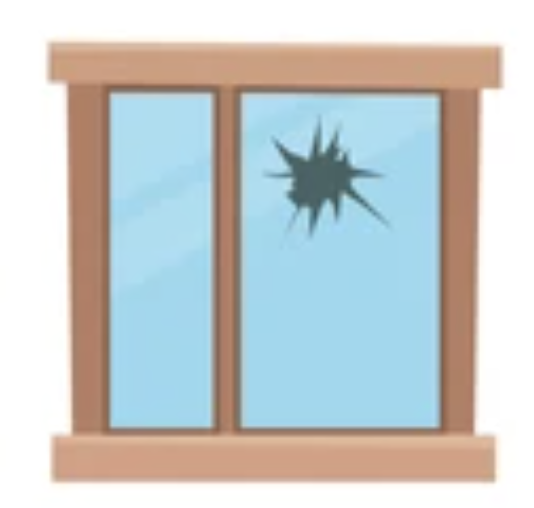},\smallimg{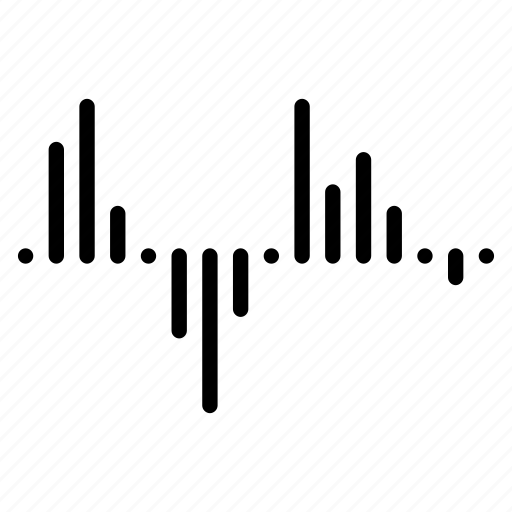},john), \\ &calls(\smallimg{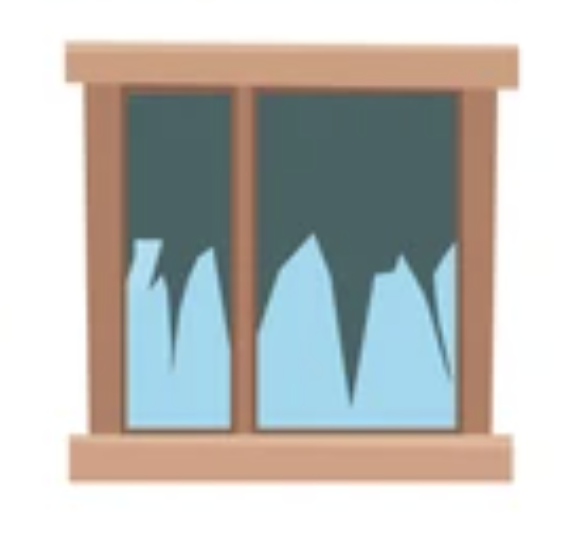},\smallimg{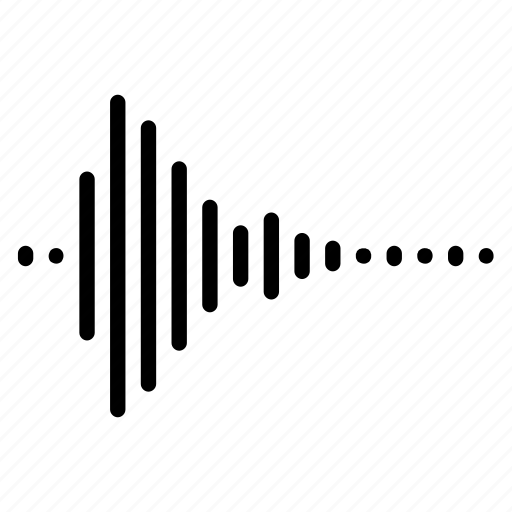},mary), ...\},
		\end{align*} we can train the parameters $\theta$ of the DeepProbLog program (both neural networks and scalar probabilities) by maximizing the log-likelihood of the training queries using gradient descent:
		
		\begin{equation*}
		\max_{\theta} \sum_{q \in \mathcal{D}} \log p(q)
		\end{equation*}

	\end{boxedexample}

	Similarly to DeepProbLog, NMLNs and RNMs use neural networks to parameterize the factors (or the weights) of a Markov Logic Network.
	\cite{rocktaschel2015injecting} computes marginal probabilities as logistic functions over similarity measures between embeddings of entities and relations. An alternative solution to exploit a probabilistic semantics is to use knowledge graphs (see also  \ref{sec:kge_gnn}) to define probabilistic priors to neural network predictions, as done in \cite{takeishi2018knowledge}.
	
	SBR\cite{diligenti2017sbr} and LTN\cite{ltn2021aij} reparametrize fuzzy atoms using neural networks that take as inputs the feature representation of the constants and return the corresponding truth value, as shown in Example~\ref{ex:sbr}. Logical rules are then relaxed into soft constraints using fuzzy logic. Many other systems exploit fuzzy logic to inject knowledge into neural models \cite{guo2016jointly,li2020augmenting}. These methods  can be regarded as variants of a unique conceptual framework as the differences are often minor and in the implementation details. 

	\begin{boxedexample}[label = ex:sbr]{Semantic-Based Regularization}

		\noindent Semantic-Based Regularization (SBR) \cite{diligenti2017sbr} is an example of an undirected model where fuzzy logic is exploited as a \textit{regularization} term when training a neural model. 
		
		Let us consider a possible grounding for the rule in Example \ref{ex:psl}:
		
		\begin{lstlisting}
		smokes(mary) $\leftarrow$ stress(mary)
		\end{lstlisting}

		For each grounded rule $r$, SBR builds a regularization loss term  $L(r)$ in the following way. First, it maps each constant $c$ (e.g. \textit{mary}) to a set of (perceptual) features $x_c$ (e.g. a tensor of pixel intensities $x_\texttt{mary}$). Each relation $r$ (e.g. \textit{smokes, stress}) is then mapped to a neural network $f_r(x)$, where $x$ is the tensor of features of the input constants and the output is a truth degree in $[0,1]$. For example, the atom \textit{smokes(mary)} is mapped to the function call $f_\texttt{smokes}(x_\texttt{mary})$.  
		Then, a fuzzy logic t-norm is selected and  logic connectives are mapped to the corresponding real valued functions. For example, when the \L ukasiewicz t-norm is selected, the implication is mapped to the binary real function $f(x,y) = \min(1, 1 - x + y)$.
		
		For the rule above, the Semantic-Based Regularization loss term is (for the \L ukasiewicz t-norm):
		
		\begin{equation*}
		L^{\text{\L}}(r) = \min \Big(1, 1 - f_\texttt{stress}(x_\texttt{mary}) + f_\texttt{smokes}(x_\texttt{mary}) \Big)
		\end{equation*}
		
		The aim of Semantic-Based Regularization is to use the regularization term together a with classical loss function for supervised learning to learn the functions associated to the relations (here $f_\texttt{stress}$ and $f_\texttt{smokes}$).

		It is worth comparing this method with the Semantic Loss  (Example \ref{ex:sl}). Both methods turn a logic formula (either propositional or first-order) to a real valued function that is used as a regularization term. However, because of the different semantics, these two methods have different properties. On the one hand, SL preserves the original logical semantics, by using probabilistic logic. However, due to the probabilistic assumption, the input formula cannot be compiled directly into a differentiable loss but needs to be first translated, i.e. compiled, into an equivalent deterministic and decomposable formula. While this step is necessary for the probabilistic model to be sound, the size of the resulting formula can be exponential in the size of the grounded theory. On the other hand, in SBR, the formula can be compiled directly into a differentiable loss, whose size is linear in the size of the grounded theory. However, in order to do so, the semantics of logic is altered, by turning it into fuzzy logic.
	\end{boxedexample}
	
	Fuzzy logic can also be used to relax rules. For example, in LRNN\cite{sourek2018lrnn}, $\partial$ILP\cite{evans:dilp}, DiffLog\cite{si2019difflog} and the approach of \cite{wang2019integrating}, the scores of the proofs are computed  using fuzzy logic connectives. 
	The theory  t-norms has  identifying parameterized (i.e. weighted) classes of t-norms \cite{vsourek2021beyond, riegel2020logical} that are very close to standard neural computation patterns (e.g. ReLU or sigmoidal layers). This creates an interesting, still not fully understood, connection between soft logical inference and inference in neural networks. A large class of methods \cite{minervini2017adversarial,demeester2016lifted,cohen2017tensorlog,weber2019nlprolog} relaxes logical statements numerically, without explicitly defining a specific semantics.
	Usually, the atoms are assigned scores in $\mathbb{R}$ computed by a neural scoring function over embeddings.
	Numerical approximations are then applied  either to combine these scores according to logical formulas or to aggregate proofs scores.
	The resulting neural architecture is usually differentiable and, thus, trained end-to-end.

	Some NeSy methods, such as PSL, have used mixed probabilistic and fuzzy semantics. In particular, Deep Logic Models (DLM)\cite{marra2019integrating} extend PSL by adding neurally parameterized factors to the Markov field, while \cite{hu2016harnessing} uses fuzzy logic to train posterior regularizers for standard deep networks using knowledge distillation  \cite{hinton2015distilling}. 

 \rev{The semantics of computational logic has also been explored and extended along other directions that have also been used within AI, for example, \textit{modal} and \textit{temporal} logics \cite{vardi1996temporal}. While their analysis is out of the scope of the paper, it is worth mentioning that also such formalisms have been  investigation from a neurosymbolic perspective \cite{garcez2006connectionist, garcez2007connectionist, hahn2021teaching}.}


	\section{Structure versus Parameter Learning}
	\label{sec:struct}
	
	Learning approaches in StarAI and NeSy are usually distinguished as to whether the structure~\cite{KokStruct} or the parameters of the model are learned~\cite{problogWeights,LowdWeights}.
	In structure learning, the learning task is to discover the logical theory, i.e., a set of logical clauses and their corresponding probabilities or weights that reliably explains the  examples. 
	What \textit{explaining the examples} exactly means  depends on the learning setting.
	In discriminative learning, we are interested in learning a theory that explains, or predicts, a specific target relation given background knowledge.
	In generative learning, there is no specific target relation; instead, we are interested in a theory that explains the interactions between all relations in a dataset.
	In contrast to structure learning, parameter learning  starts with a given logical theory and only learns the corresponding probabilities or weights.

	
	%
	%

	Structure learning is an inherently NP-complete problem of searching for the right combinatorial structure, whereas parameter learning can be achieved with any curve fitting technique, such as gradient descent or least-squares.
	While parameter learning is, in principle, an easier problem to solve, it comes with a strong dependency on the provided user input. If the provided clauses are of low quality, the resulting model will also be of low quality.
	Structure learning, on the other hand, is less dependent on the user provided input, but is an inherently more difficult problem.

	\subsection{Implications for StarAI}
	Both structure  and parameter learning are common in StarAI.
	Structure learning  in StarA is an instance of learning by search~\cite{Mitchell82} and is closely connected to program synthesis.
	The existing techniques are typically extensions of techniques originating in inductive logic programming (ILP) ~\cite{mugg:ilp94,luc:book},  which learn deterministic logical theories, and probabilistic graphical models (PGMs), which learn Bayesian or Markov networks from data.
	Being an instance of learning by search, the central components of a learning framework are a space of valid structures and a search procedure.
	In  ILP, valid structures are logical theories;  for Bayesian networks, valid structures are DAGs capturing their graph structure.
	The resulting search space is then traversed with generic search procedures.

	 StarAI structure learning techniques suffer from a combinatorial explosion.
	That is especially the case with ILP techniques, in which  the search space consists of programs containing  several clauses. 
	Therefore, it is necessary to  limit the search space to make learning tractable.
	The most common way to do this is to impose a \textit{language bias} -- a set of instructions on how to construct the search space, such that it is narrowed down to a subset of the space of all logical theories.
	Though language bias can make the problem more tractable, it requires special care:  too many restrictions might eliminate the target theory, while too few restrictions make the search space too large to traverse.
	Another  strategy is to leverage the compositionality of logic programs: 
 adding an additional clause to a program increases its coverage and does not affect the prediction of examples covered by the initial program.
	That is, we can learn a single clause at a time instead of simultaneously searching over theories containing multiple clauses.

	Learning clauses and their probabilities is usually treated as a two stage process.
	ILP-based StarAI learning techniques first identify useful (deterministic) clauses, and then learn the corresponding probabilities or weights via parameter learning.  Similarly, StarAI methods grounded primarily in PGMs, such as MLNs, search for frequently occurring cliques in data~\cite{Kok:2010}, lift them into logical clauses, and then learn the weights or probabilities.
	Parameter learning techniques are often also extensions of well known statistical approaches such as least-squares regression~\cite{problogWeights}, gradient descent~\cite{LowdWeights}, and expectation maximisation~\cite{GutmannTR11}.


	\begin{boxedexample}[label = ex:probfoil]{Structure learning with ProbFoil}

		As an illustration of structure learning techniques, we will focus on ProbFoil~\cite{probfoil}.
		Assume that we are interested in learning the definition of \textit{grandparent} from akinship data.
		That is, we are given a set of examples of grandparent relations
		\begin{lstlisting}
		grandparent(abe,lisa).
		grandparent(abe,bart).
		grandparent(jacqueline,lisa).
		grandparent(jacqueline,bart).
		\end{lstlisting}
		and background knowledge containing the following facts:
  \begin{lstlisting}
	father(homer,lisa). 
        father(homer,bart). 
        father(abe,homer).
	mother(jacqueline,marge). 
        mother(marge,bart).
	mother(marge,lisa).
		\end{lstlisting}
		
		ProbFoil iteratively searches for a single clause that covers as many examples as possible, until all examples are covered or it adding more clauses does not improve the results. 
		While searching for the best clause, it starts from the most general one, \textit{grandparent(X,Y)}, which effectively states that every pair of people forms a grandparent relationship.
		Then it gradually specialises the clause by adding literals to the body. For instance, extending  \textit{grandparent(X,Y)} with a \textit{mother/2} predicate results in the following clauses
		\begin{lstlisting}
		grandparent(X,Y) <- mother(X,Y).
		grandparent(X,Y) <- mother(X,X).
		grandparent(X,Y) <- mother(Y,X).
		grandparent(X,Y) <- mother(Y,Y).
		\end{lstlisting}
		Extending the initial clause with the \textit{father/2} results in similar clauses.
		Having the new candidate clauses, ProbFoil scores each candidate by counting how many positive and negative examples are covered.
		These candidate clauses would not cover any examples and ProbFoil continues to refine the candidates by adding another literal to the body.
		This would result in clauses of the following form:
		\begin{lstlisting}
		grandparent(X,Z) <- mother(X,Y), father(Y,Z).
		grandparent(X,Z) <- mother(X,Y), mother(Y,Z).
		grandparent(X,Z) <- father(X,Y), mother(Y,Z).
		...
	\end{lstlisting}
		Some of the new candidates will cover only positive examples, such as 
  \begin{lstlisting}
  grandparent(X,Z) <- mother(X,Y), mother(Y,Z)
  \end{lstlisting}
  that covers both examples
  \begin{lstlisting}
  grandparent(jacqueline,lisa).
  grandparent(jacqueline,bart).
   \end{lstlisting}
		Having found one clause, ProbFoil learns the corresponding probability labels and adds the clause to the theory.
		ProbFoil then repeats the same procedure, starting with the most general clause, to cover the remaining examples.
	\end{boxedexample}

	\subsection{Implications for  NeSy}
	
	While StarAI learning techniques are categorised exclusively as either structure or parameter learning, \nesy{} learning techniques combine both.
	We will now discuss four groups of NeSy learning approaches: neurally-guided search, structure learning via parameter learning, program sketching, and implicitly structure learning.

	\textit{Neurally guided structure search} \cite{NGSynth,ellis:libraries,ellis:repl,Valkov2018HOUDINILL} is the \nesy{} paradigm most similar to structure learning in StarAI.
	It addresses one of the major weaknesses of StarAI structure learning methods - uninformed search over valid theories.
	Instead, neurally guided search relies on a \textit{recognition model}, typically a neural network, to prioritise parts of the symbolic search space so that the target model can be found faster.
	Generally speaking, the recognition model predicts the probability of a certain structure, e.g. a predicate or an entire clause, to be a part of the target model.
	For instance, Deepcoder~\cite{deepcoder} uses input-output examples to predict the probability of each predicate appearing in the target model.
	Therefore, Deepcoder turns a systematic search into an informed one by introducing a ranking over predicates in the search space.
	Likewise, EC$^{2}$~\cite{ellis:libraries}  derives the probability of a program solving the task at hand.
	Several approaches push this direction  further and explore the idea of replacing an explicit symbolic model space with an implicit generative model over symbolic models ~\cite{nye:neurips20,mao2018the}. 
	For instance, in \cite{nye:neurips20}, the authors learn a generative model over grammar rules, conditioned on the examples.
	Structure learning is then performed by sampling grammar rules from the generative model, according to their probability, and evaluating them symbolically on the provided examples.

	These approaches clearly show how symbolic search can be made tractable by introducing various forms of guidance via neural models.
	These guidance-based approaches reduce, to a large extent, the most important weakness of symbolic structure learning approaches - the generation of many useless clauses or models.
	On the other hand, these approaches often need large amounts of data for training, sometimes millions of examples \cite{ellis:repl} even though creating data is relatively easy by enumerating random model structures and sampling examples from them~\cite{ellis:repl}.

	\begin{boxedexample}[label = ex:ngps]{Neurally-guided structure learning}
		
		\rev{To illustrate neurally-guided search, we use the approach of  Zhang et al. \citep{ngps}.
        StarAI techniques for structure learning typically perform a systematic search, which results in many useless models being tested.
        Given $N$ atoms, we can construct $N^l$ clauses of length $l$; this is an enormous space that is difficult to search efficiently.} \\

        \rev{Zhang et al. sidestep the systematic search by introducing a neural network that chooses which programs to explore next.
        This search space is made of clauses such that an empty clause is at the top and children are extensions of the empty clause with all possible predicates; their children are further extensions with all individual atoms.} \\

        \rev{The approach follows a top-down search strategy, exploring shorter clauses before longer ones, with a twist: instead of following a predefined order, the approach uses a neural network to decide which child to expand next. 
        The approach can be viewed as a best-first search with a heuristic function implemented by a neural model. To this end, the network  (1) encodes all literals in each clause separately, (2) scores all literals, (3) pools the scores of each literal per candidate, and (4) chooses the best candidate based on their scores.
        Ordering the search space in this way leads to substantial improvements in computation time, typically several orders of magnitude.}

	\end{boxedexample}

	An alternative way to reduce the combinatorial complexity of learning is to  learn only a part of the program.
	This is known as \textit{program sketching:} a user provides an almost complete target model with certain parts being unspecified (known as \textit{holes}). 
	For instance, when learning a model in the form of a (logic) program for sorting numbers or strings, the user might leave the comparison operator unspecified and provide the rest of the program.
	The learning task is then  to fill  in the holes.
	Examples of \nesy{} systems based on sketching are DeepProbLog and $\partial$4, which fill in the holes in a (symbolic) program via neural networks.

	The advantage of sketching is that it provides a nice interface for \nesy{} systems, as the holes can be filled either symbolically or neurally.
	Holes provide a clear interface in terms of inputs and outputs and are agnostic to the specific implementation.
	The disadvantage of sketching is that the user still needs to know, at least approximatively, the structure of the program.
	The provided structure, the sketch,  acts as a strong  bias.
	Deciding which functionality is left as a hole is a non-trivial issue: as the sketch becomes less strict,  the search space becomes larger.

	\textit{Structure learning via parameter learning} (Example \ref{ex:difflog}) is arguably the most prominent learning paradigm in \nesy{}, positioned in between the two StarAI learning paradigms.
	Structure learning via parameter learning is technically equivalent to parameter learning in that the learning tasks consists of learning the probabilities of a fixed set of clauses.
	However, in contrast to StarAI in which the user carefully selects the informative clauses, the clauses are typically enumerated from  user-provided templates of predefined complexity.
	Constructed in this way, the majority of clauses are noisy and erroneous  and are  of little use.
	They would receive very low, but non-zero, probabilities.
	 Approaches that follow this learning principle include NTPs \cite{rocktaschel2017ntp}, $\partial$ILP \cite{evans:dilp}, DeepProbLog\cite{manhaeve2018deepproblog}, NeuralLP \cite{Cohen_NeuralLP} and DiffLog \cite{si2019difflog}.

	The advantage of structure learning via parameter learning is that it removes the combinatorial search from the learning.
	However, the number of clauses that needs to be considered is still extremely large, which  leads to difficult optimisation problems (cf. \cite{evans:dilp}). Furthermore, irrelevant clauses are never removed from the model and are thus always considered during inference.
	This can lead to spurious interactions even when low probabilities are associated to irrelevant clauses: as the number of irrelevant clauses is extremely large, their cumulative effect can be substantial.

	\begin{boxedexample}[label = ex:difflog]{Structure learning via parameter learning}

		As an illustration of structure learning via parameter learning, we focus on DiffLog \cite{si2019difflog}.
		DiffLog expects the candidate clauses to be provided by the user.
		The user can either provide the rules she knows are useful or construct them by using a clause template and instantiating it \cite{metagol}.. 

		Given a set of positive examples, DiffLog proceeds by constructing \textit{derivation trees} for each example.
		Consider the problem of learning the connectivity relation over a graph.
		The input tuples (background knowledge in StarAI terminology) specify edges in a graph
		\begin{lstlisting}
	edge(a,b).  edge(b,c).  edge(b,d).  edge(d,e). edge(c,f).
		\end{lstlisting}
		The examples indicate the connectivity relations among the nodes in the graph (for simplicity, consider only the following two examples)
		\begin{lstlisting}
		connected(a,b). connected(a,c). 
		\end{lstlisting}
		Also assume that the candidate clause set contains the following clauses (with $p_1$ and $p_2$ their weights):
		\begin{lstlisting}
		$p_1$::connected(X,Y) <- edge(X,Y).      
		$p_2$::connected(X,Y) <- edge(X,Z), connected(Z,Y).
		\end{lstlisting}
		Derivation trees are essentially proofs of individual examples that correspond to branches in the SLD-tree \cite{lloyd:book}.
		For instance, the example \textit{connected(a,b)} can be proven using the first clause, whereas the example \textit{connected(a,c)} can be proven by chaining the two clauses ($connected(a,c) \leftarrow edge(a,b), connected(b,c)$ and $connected(b,c) \leftarrow edge(b,c)$).

		DiffLog uses derivation trees to formulate the learning problem as numerical optimisation over the weights associated with the rules.
		More precisely, DiffLog defines the probability of deriving an example as the product of the weights associated to the clauses used in the derivation tree of the corresponding example.
		For instance, DiffLog would formulate the learning problem for the two examples as follows
		$$\min_{p_1, p_2} \  \underbrace{(1 - p_1)}_{\small \tt connected(a,b)} + \underbrace{(1 - p_1\times p_2)}_{\small \tt connected(a,c)}.$$

	\end{boxedexample}

	The last group of approaches learns the structure of a program only \textit{implicitly}.
	For instance, Neural Markov Logic Networks (NMLN) \cite{marra2019nmln}, a generalisation of MLNs, extract structural features from relational data.
	Whereas MLNs define potentials only over cliques defined by the structure (logical formulas) of a model, NMLNs add potentials over \textit{fragments} of data (projected over a subset of constants).
	NMLNs thus do not necessarily depend on the symbolic structure of the model, be it learned or provided by a user, but can still learn to exploit relational patterns present in data.
	Moreover, NMLNs can incorporate embeddings of constants.
	The benefit of this approach is that it removes combinatorial search from learning and performs learning via more scalable gradient-based methods.
	However, one loses the ability to inspect and interpret the discovered structure.
	Additionally, to retain tractability, NMLNs limit the size of fragments which  imposes limits on the complexity of the discovered relational structure.

	\section{Symbolic vs subsymbolic representations}
 \label{sec:symb_vs_subsymb}
	
	In neurosymbolic artificial intelligence,  approaches can be characterized by the  way they represents entities and relationships in two classes: symbolic methods, where entities are represented using symbols such as strings and natural numbers, and subsymbolic methods, where entities are represented using numerical or distributed representations.

	Symbolic representations include constants $(an, bob)$, numbers $(4, -3.5)$, variables $(X, Y)$ and structured terms $f(t_1, ... ,t_n)$ where $f$ is a functor and the $t_i$ are \rev{constants}, variables or structured terms. 
    \rev{Structured terms are a powerful construct that can represent arbitrary structures over entities, such as relations, lists or trees.}
	subsymbolic AI systems, such as neural networks, require that entities are represented numerically using vectors, matrices or tensors. Throughout this this section, we will call these subsymbolic representations or subsymbols. 
	subsymbolic AI systems usually require that these representations have a fixed size and dimensionality. Exceptions require special architectures and are still the subject of active research (e.g. RNNs for list-like inputs or GCNs~\cite{gcn} for graph-type inputs).
	
	\paragraph{Comparing representations}
	A powerful and elegant mechanism for reasoning with symbols in  logic is  {\em unification}. Essentially, it calculates the most general substitution that makes two symbols syntactically equal, if it exists. \rev{This does not allow one to compare two different entities, but allows one to find what two structured terms have in common.}
	For instance,  the  terms $p(a,Y)$ and $p(X,b)$ can be unified using the substitution $\{X=a, Y=b\}$. 
	Conversely, due to their numerical nature, calculating the similarity between subsymbols is straightforward. 
	Similarity metrics such as the radial-basis function or distance metrics such as the L1 and L2 norm can be used. However, it is not clear when to decide that two subsymbolically represented entities are the same.
	
	\paragraph{Translating between representations}
	Many systems need to translate back and forth between symbolic and subsymbolic representations. In fact, a lot of research on deep learning is devoted to efficiently representing symbols so that neural networks can properly leverage them.
	A straightforward example is to translate symbols to a subsymbolic representation that can serve as  input for a neural network. Generally, these symbols are replaced by a one-hot encoding or by learned embeddings. 
    \rev{Note, however, that this does not imply that the system can perform symbolic manipulation on this input. Rather, it serves as an index to a set of learned, latent embeddings.}
	A more interesting example is encoding relations in subsymbolic space. The wide variety of methods~\cite{bordes2013translating, trouillon2016complex, distmult} developed for this purpose indicates that this is far from a solved problem. 
	Different encodings have different benefits. For example, TransE~\cite{bordes2013translating} encodes relations as vector translations from subject to object embeddings. A disadvantage is that symmetric relations are represented by the null vector, and entities in symmetric relations are pushed towards each other.
	More complex structures are even harder to represent. For example, there is currently a lot of research in how to utilize graph-structured data in neural networks (cf. \ref{sec:kge_gnn}).
	
	Translating from a subsymbolic representation back to a symbolic one happens, for  example, at the end of a neural network classifier. Here, a subsymbolic vector needs to be translated to discrete classes. Generally, this happens through the use of a final layer with a soft-max activation function which then models the confidence scores of these classes as a categorical distribution. However, other options are possible. For example, some methods are only interested in the most likely class, and will use an arg-max instead. Alternatively, a Gumbel-softmax activation can be used as a differentiable approximation of sampling from the categorical distribution.
	
	\subsection{Implications for  StarAI and NeSy}
	In StarAI systems, the input, intermediate and output representations are all using the same symbolic representations. \rev{Although there are StarAI systems that can support numerical values, these are still treated as symbols, which is  different than a latent, subsymbolic representation.}
	In neural systems, the input and intermediate representations are subsymbolic. The output representation can be either symbolic (e.g. classifiers) or subsymbolic (e.g. auto-encoders, GANs).
	The most important aspect of neurosymbolic systems is that they combine symbolic and subsymbolic representations.
	NeSy systems can be categorized by how they do this. 
 We distinguish several approaches.
	
	In the first approach, the inputs are symbolic, but they are translated to subsymbols in a single translation step, after which  the intermediate representations used during reasoning are purely subsymbolic. This approach is followed by the majority of NeSy methods. Some examples include Logic Tensor Networks~\cite{ltn2021aij}, Semantic-based Regularization~\cite{diligenti2017sbr}, Neural Logic Machines~\cite{NLM} and TensorLog~\cite{cohen2017tensorlog}.

	\begin{boxedexample}[label = ex:ltn]{Logic Tensor Networks}
		
		Logic tensor networks~\cite{ltn2021aij} make this translation step explicit. The authors introduce the concept of a \textit{grounding} (not to be confused with the term grounding used in logic). Here, a grounding is a mapping of all symbolic entities onto their subsymbolic counterpart. More formally, the authors define a grounding as a mapping $\mathcal{G}$ where:
		\begin{itemize}
			\item $\mathcal{G}(c) \in \mathbb{R}^n$ for every constant symbol $c$
			\item $\mathcal{G}(f) \in \mathbb{R}^{n.m} \rightarrow \mathbb{R}^n$ for every function $f$ of arity $n$
			\item $\mathcal{G}(p) \in \mathbb{R}^{n.m} \rightarrow [0,1]$ for every predicate $p$ of arity $n$
		\end{itemize}
		The grounding of a clause is then performed by combining the aforementioned groundings using a t-norm.
	\end{boxedexample}
	
	In the second approach, intermediate representations are both symbolic and subsymbolic, but not simultaneously. This means that some parts of the reasoning work on the subsymbolic representation, and other parts deal with the symbolic representation, but not at the same time.
	This is indicative of NeSy methods that implement an interface between the logic and neural aspect.
	This approach is more natural for systems that originate from a logical framework  such as DeepProbLog~\cite{manhaeve2018deepproblog}, NeurASP~\cite{yang2020neurasp}), ABL~\cite{dai2019abl} and NLog~\cite{tsamoura2021neural}.
	
	\begin{boxedexample}[label = ex:abl]
{ABL}
		In ABL~\cite{dai2019abl} there are three components that function in an alternating fashion. There is a perception model, a consistency checking component and an abductive reasoning component.
		Take for example the task where there are 3 MNIST images that need to be recognized such that the last is the result of applying an operation on  the first two (e.g. $\digit{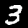}+\digit{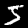}=\digit{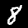}$). The structure of the expression is given as background knowledge, but the exact operation (addition) needs to be abduced.
		First, the perception model classifies the images into pseudo-labels, using the most likely prediction (i.e. arg-max). 
		The abductive reasoning component then tries to abduce a logically consistent hypothesis. 
		For example, if the digits are correctly classified as $3$, $5$ and $8$, the only logically consistent hypothesis is that the operation is an addition.
		If this is not possible, there is an error in the pseudo-labels. 
		A heuristic function is then used to determine which pseudo-labels are wrong.
		The reasoning module then searches for logically consistent pseudo-labels. These revised pseudo-labels are then used to retrain the perception model.
	\end{boxedexample}
	
	In the final approach, intermediate representations are considered simultaneously as symbolic and subsymbolic by the  reasoning mechanism.
 This is implemented in only a few methods, such as the NTP\cite{rocktaschel2017ntp} and the CTP\cite{minervini2020learning}. 
	\begin{boxedexample}[label = ex:ntp]{Neural Theorem Prover}
		
		In the Neural Theorem Prover\cite{rocktaschel2017ntp}, two entities can be unified if they are similar, and not just if they are identical. As such, the NTP   interweaves both symbols and subsymbols during inference. For each symbol $S$, there is a learnable subsymbol $T_S$.
		Soft-unification happens by applying the normal unification procedure where possible. However, if two symbols $S_1$ and $S_2$ can not be unified, the comparison is assigned a score based on the similarity between $T_{S_1}$ and $T_{S_2}$. The similarity is calculated using a radial basis function $\varphi(||x-y||_2)$.
		
		For example, to unify   \texttt{mother(an,bob)} and \texttt{parent(X,bob)}, soft-unification proceeds as follows:
		\begin{align*}
		\{\mathtt{mother(an,bob)} &= \mathtt{parent(X,bob)}\} \\
		&\Downarrow \quad \varphi(\mathtt{mother}, \mathtt{parent}) \\
		\{\mathtt{an} = \mathtt{X}&, \mathtt{bob} = \mathtt{bob}\} \\
		&\Downarrow \quad X = an \\
		\{\mathtt{bob} &= \mathtt{bob}\} \\
		&\Downarrow\\
		&\{~\}
		\end{align*}
		
		Soft-unification is not only used to learn  which constants and predicates are similar, but can also be used to perform rule learning. By adding new, parameterized rules with unique predicates, soft-unification allows these new predicates to become very similar to other predicates and as such behave as newly introduced rules.
		For example, consider the  program consisting of the fact \textit{mother(an,bob)} and a single parameterized rule $r1(X,Y) \leftarrow r2(Y,X)$. The Neural Theorem Prover can answer the query \textit{child(bob,an)} as follows: 

		\begin{center}
			\tikzset{every picture/.style={line width=0.75pt}} 

\begin{tikzpicture}[x=0.75pt,y=0.75pt,yscale=-1,xscale=1]

\draw    (240,60) -- (140,100) ;
\draw    (240,60) -- (340,100) ;
\draw    (340,140) -- (340,180) ;

\draw (201,32) node [anchor=north west][inner sep=0.75pt]   [align=left] {$\mathtt{child(bob,an)}$};
\draw (299,111) node [anchor=north west][inner sep=0.75pt]   [align=left] {$\mathtt{r2(ann,bob)}$};
\draw (335,191) node [anchor=north west][inner sep=0.75pt]   [align=left] { $\mathtt{T}$ };
\draw (131,112) node [anchor=north west][inner sep=0.75pt]   [align=left] { $\mathtt{T}$ };
\draw (151,62) node [anchor=north west][inner sep=0.75pt]   [align=left] {(1)};
\draw (299,62) node [anchor=north west][inner sep=0.75pt]   [align=left] {(2)};
\draw (349,151) node [anchor=north west][inner sep=0.75pt]   [align=left] {(3)};

\end{tikzpicture}
		\end{center}
		
		\begin{align*}
		(1) ~& \mathtt{child} = \mathtt{mother} &\varphi(||T_{child}-T_{mother}||_2)\\
		& \mathtt{bob} = \mathtt{an}  &\varphi(||T_{an}-T_{bob}||_2)\\
		(2)~ & \mathtt{child} = \mathtt{r1} &\varphi(||T_{child}-T_{r1}||_2)\\
		(3) ~& \mathtt{child} = \mathtt{r2}  &\varphi(||T_{r2}-T_{mother}||_2)
		\end{align*}
		
		The figure above shows the two possible derivations the neural theorem prover can make to infer \textit{child(bob, an)}. One the one hand, it can soft-unify with the fact \textit{mother(an, bob)}, where \textit{mother} unifies with \textit{child} and \textit{an} with \textit{bob}. On the other hand, it can use the parameterized rule which encodes an inverse relation. In that case, \textit{mother} unifies with \textit{r1} and \textit{r2} with \textit{child}. If we optimize the subsymbolic embeddings for the latter, this will be equivalent to learning the rule $mother(X,Y) \leftarrow child(Y,X)$. This example  also shows that soft-unification potentially adds a lot of different proofs, which can result in computational problems. This problem was solved in later iterations of the system \cite{minervini2019differentiable}.
	\end{boxedexample}

		\section{Logic vs Probability vs Neural}
	\label{sec:paradigms}
	
	When two or more paradigms are integrated, examining which of the base paradigms are preserved, and to which extent, tells us a lot about the strengths and weaknesses of the resulting paradigm.
	It has been argued \cite{de2019neuro} that when combining different perspectives in one model or framework, such as logic, probabilistic and neural ones, it is  desirable  to have the original paradigms as a special case.

	\rev{In this section, we analyze to which extent different models in  StarAI and NeSy  preserve the three basic paradigms. Intuitively, with preserving we mean to which extent one can exactly replicate the model and  inference algorithm of the original paradigm. We will use the  capital letters \textit{L, P} and \textit{N} to label systems where the logic, probability and neural paradigms can be recovered in full.  We will use lowercase letters (i.e. \textit{l, p} and \textit{n}) when a method only partially recovers these paradigms, i.e. retain some but not all of the features.} The absence of a letter means that the paradigm is not considered by an approach.
	
	\subsection{StarAI: Logic + Probability}
	
	Traditionally, StarAI focused on the integration of  logic and probability.
	
	\textbf{\textit{lP:}}
	The classical knowledge-based model construction approach uses logic only to generate a probabilistic graphical model. 
	Thus the graphical model can be used to define the semantics of the model and also to perform inference.
	This can make it harder to understand the effects of applying logical inference rules to the model. 
	For instance, in MLNs, the addition of the resolvent of two weighted rules makes it hard to predict the effect on the distribution.

	\textbf{\textit{Lp}:} On the other hand, the opposite holds for probabilistic logic programs (PLPs) and their variants. 
	While the effect of a logical operation is clear, it is harder to identify and exploit properties such as conditional or contextual independencies, that are needed for efficient probabilistic inference. 
	

	
	
	\subsection{NeSy: Logic + Probability + Neural}
	In NeSy, we consider a third paradigm: neural computation.
	With neural computation, we refer mainly to the set of models and techniques that allows for exploiting (deep) latent spaces to learn intermediate representations. This includes dealing with perceptual inputs and also dealing directly with embeddings of symbols. 
	
	\textbf{\textit{lN}}: Many \nesy{} approaches  focus on the neural aspect (i.e., they originated as a neural method to which logical components have been added). For example, LTNs and SBRs turn the logic into a regularization function to provide a penalty whenever the  logical constraints are violated. At test time the logical loss component is dropped and only the network is used to make predictions. Moreover, by using fuzzy logic, these methods do not integrate the probabilistic paradigm.

	\textbf{\textit{Ln}}: Another class of \nesy{} methods does retain the focus on logic. 
	These methods usually expand an existing logical framework into a differentiable version. Examples include LRNNs \cite{sourek2018lrnn}, TensorLog \cite{cohen2017tensorlog}, DiffLog \cite{si2019difflog}, $\partial$ILP \cite{evans:dilp}, $\partial$4 \cite{BosnjakDFI} and NTPs \cite{rocktaschel2017ntp}.
	The key inference concepts are mapped onto an analogous concept that behaves identically for the edge cases but is continuous and differentiable in non-deterministic cases. 
	As  described in the previous sections,  many such systems cast logical inference  as forward or backward chaining. The focus on logic is clear if one considers that logical inference is performed symbolically  to build the network and the semantics is relaxed only in a subsequent stage to learn the parameters. \rev{While the architecture mimics the logical reasoning, it is often far from the deep-stacked architecture of neural networks.}
 
    \textbf{\textit{LN}}: \rev{It is worth mentioning a later iteration of LRNN, where the framework has been extended to allow for tensorial weights on  atoms and custom aggregation functions \cite{vsourek2021beyond}. In that framework, it is shown how specifying logic rules can be regarded as specifying the layers of a deep architecture. This provides a nice  and complete integration between forward-chaining logical reasoning and neural networks
    that is able to implement any existing neural architecture.
    }
 

	\textbf{\textit{lPN} and \textit{LpN}} There are two final classes of methods that start from existing StarAI methods, \textit{lP} and \textit{Lp} respectively, and extend them with primitives that can be interfaced with neural networks and allow for differentiable operations.
	In the \textit{lPN} class,  NeSy methods such as SL, RNMs and NMLNs follow the knowledge-based model construction paradigm.  In the \textit{LpN} class, methods such as DeepProbLog and NeurASP  extend PLP. 
	
	
There is usually a trade-off that one must make: 	
systems in the \textit{lN} or \textit{Ln} classes are usually more scalable but  \textit{(i)} do not model a probability distribution and \textit{(ii)} often relax the logic. On the contrary, \textit{LpN} or \textit{lPN} systems preserve the original paradigms but at the cost of more complex inference (e.g. they usually resort to exact probabilistic inference).
	
	An aspect that significantly aids in developing a common framework, and analysing its properties, is the development of an intermediate representation language that can serve as a kind of \emph{assembly language} \cite{pedroassembly}. 
	One such idea concerns performing probabilistic inference by mapping it onto a  weighted model counting (WMC) problem. 
	This can then in turn be solved by compiling it into a structure (e.g. an arithmetic circuit) that allows for efficient inference.
	This has the added benefit that this structure is differentiable, which  facilitates the integration between logic based systems and neural networks. 
	 StarAI based systems often use this approach. 

\section{Tasks}
\label{sec:tasks}
	
	\rev{In this Section, we analyze the learning tasks to which the \nesy{} models considered in this paper have been  applied.}

        \paragraph{Distant Supervision} \rev{A classical task in NeSy is to use logic as distant supervision for a learning model.  Here, input $X$ is paired with label $y$. However, instead of using a single model to map $X$ to $y$, the input $X$ is firstly mapped to a set of intermediate concepts $C$ by a (set of) neural networks. Then, these concepts are used to compute $y$ in a symbolic way. Logic programs are usually exploited to map the concepts $C$, represented as logical atoms, to the label $y$, which represents the logical query.  Therefore,  the neural networks are not directly supervised (on $C$) but they are only distantly supervised through the label $y$ and the knowledge contained in the logic program. The intuition  is that when the label $y$ is only weakly linked to the input, it is  more convenient to break the task in several easier subtasks and then compose them using background knowledge in the form of a logic program. Notice that the logic program is fundamental for the inference. Without the program, the networks will not be able to solve their subtasks, as there is no direct supervision. Moreover, by splitting the task into subtasks, the inference done by the composite system (neural + logic) is far more explainable than a corresponding end-to-end neural network. A classical example is the MNIST addition \cite{manhaeve2018deepproblog}, shown in Example \ref{ex:mnist_addition}. Distant supervision tasks are very common in prototypical systems such as DeepProbLog, DeepStochLog, NLog, NeurASP, SATNet \cite{wang2019satnet}. A downside of such tasks is that, to enable learning of untrained neural subtasks, the logic has to consider all possible combinations of concepts that are compatible with the label $y$, even though only few (or one) are correct. The challenge is to balance the exploration of multiple combinations with  a greedy strategy for scaling to larger problems \cite{tsamoura2021neural, manhaeve2021approximate, mandi2022decision}. Other problems falling in this category are scene parsing, image segmentation and semantic image interpretation \cite{donadello2017ltn, semanticreferee}}

        \begin{boxedexample}[label = ex:mnist_addition]{MNIST Addition} 
        Given the classical MNIST dataset, $\mathcal{D} = \{(x_i,y_i)\}$, with $x_i$ an MNIST image, and $y_i$ its numeric label, the MNIST addition dataset is built by mapping pairs of images to the label representing their sum. In particular, $\mathcal{D}_{\text{add}} = \{(x_i,x_j,z_{ij}) : z_{ij} = y_i + y_j \land (x_i,y_i),(x_j,y_j) \in \mathcal{D}\}$. The idea is to learn to classify the digits without direct supervision on their labels, but only using distant supervision about sums of such images. 
     The task is often also coupled to  background knowledge of what  addition is, e.g. in Prolog syntax:

        \begin{lstlisting}
addition(X1, X2, Z) <- digit(X1,Y1), digit(X2,Y2), Z is Y1 + Y2. 
        \end{lstlisting}

        Such knowledge is used to reason about the (most-likely) pairs \texttt{Y1,Y2} that sum to the provided label \texttt{Z}. Logic is then used to link the actual outputs of the learning model \texttt{Y1,Y2} to the distant supervision \texttt{Z}.

        \end{boxedexample}
	
	\paragraph{Semi-supervised Classification} \rev{A related class of tasks is semi-supervised classification  \cite{chapelle2006semi}} with knowledge. Here, the starting point is a standard classification task, where a set of inputs $X$ is mapped by a neural model to a set of labels $C$. However, we are also provided with some additional knowledge $y$ related to the labels $C$ of the inputs. This knowledge is often expressed in terms of logical rules and programs. The setting is very similar to  distant supervision, where we have three levels: inputs $X$, concepts $C$ and additional labels $y$. However, in this case, we have also access to supervision for some (usually few) concepts $C$. Although this task could be tackled in a purely supervised way by discarding the information contained in $y$, \nesy{} approaches can improve the predictions of several input patterns using the external knowledge. When the external knowledge is  relating  concepts $C$ of multiple patterns, the task is called collective classification \cite{sen2008collective}, as one can improve the accuracy on multiple patterns by collectively predicting their classes. A classical example in this setting is document classification in citation networks, cf. Example \ref{ex:citation}. By treating the information contained in $y$ as extra knowledge, these tasks are often tackled using regularization based systems, like SBR\cite{diligenti2017sbr}, DLM\cite{marra2019integrating}, RNM\cite{marra2020relational} or Semantic Loss\cite{xu2018semantic}. However, logic programs can also be  used to simulate a label-passing scheme along the citation network, as done in DeepStochLog \cite{winters2021deepstochlog}. A characteristic of this class of tasks is that the additional information $y$ is often very noisy (e.g. the manifold rule in the citation network is not always valid). While this task is closely related to  distant supervision, there is an important difference: in semi-supervised classification, the additional knowledge $y$ is meant to provide an additional signal, which, however, would not suffice in the absence of direct supervision on the concepts $C$. 

    \begin{boxedexample}[label = ex:citation]{Document classification in citation networks}

    In document classification in citation networks, we are provided with both labelled and unlabelled scientific papers. A label is often the domain area of the paper (e.g. Machine Learning, Artificial Intelligence, Databases, etc.). 
    However, a network of citations between papers is also provided, linking papers between domains.

    The idea of the document classification task is that in many domains, a paper cited by other papers with a certain label is likely to belong to the same domain. When classifying a document,  one has to balance  the signal coming from the features of the document (i.e. words) and that coming from neighbors in the citation network to  provide a \textit{collective} prediction.

    In \nesy{} systems, this is usually done by coupling the subsymbolic model with a rule of the following type:

    \begin{lstlisting}
 w:: domain(X,Y) <- cite(X,X1), domain(X1,Y). 
    \end{lstlisting}

    The rules get a different weight according to the  domain to account for the differences between them.

    \end{boxedexample}

    \paragraph{Knowledge Graph Completion}
    \rev{Another  common task in \nesy{} is knowledge graph completion (KGC) or link prediction. A knowledge graph (KG) is a pair of $(E,R)$, where $N$ is the set of entities and $R$ the set of edges. In a KG, an edge is a triple $(e_1, r, e_2)$, where $e_1$ and $e_2$ are the head and tail of the edge and $r$ is the  relation between them. In a KGC task, the goal is to predict missing edges in the input graph. Link prediction has been one of the key tasks in StarAI \cite{Getoor07:book}, and more recently also in NeSy as NeSy allows to merge symbolic reasoning (from StarAI) with the recent geometric deep learning approaches based on Knowledge Graph Embeddings (KGE) \cite{wang2017knowledge} and Graph Neural Networks \cite{scarselli2008graph}. \nesy{} systems focusing on this task include NTPs \cite{rocktaschel2017ntp}, NMLN \cite{marra2019nmln}, DLM \cite{marra2019integrating}, DiffLog \cite{si2019difflog}, TensorLog \cite{cohen2017tensorlog}.}

    \paragraph{Generative Tasks} Most previously mentioned tasks can be described as classification\footnote{Even though, many of them use a generative model to tackle the classification task, instead of a conditional one.}.  \nesy{} has recently also focused on tasks concerned with modeling the input data distribution as accurately as possible. The goal is then to sample new patterns from the learned distribution. The idea behind \nesy{} generative approaches is that one can learn important features  from data  using deep generative models (e.g. variational auto-encoders or Markov Chain Monte Carlo methods). Combining symbolic features with logic reasoning can be used to control,  stratify and simplify the inference. The generative modeling can either refer to the relational structure, e.g. molecule generation in NMLNs \cite{marra2019nmln}, or to the subsymbolic space, e.g. image generation in VAEL \cite{misino2022vael} or \cite{skryagin2020sum}.

    \paragraph{Knowledge Induction} \rev{Rather than exploiting symbolic knowledge  predictive tasks, one can also induce  
     \textit{symbolic knowledge}. In all  previous tasks,  symbolic knowledge is provided by the user as part of the input. However, as explored in Section \ref{sec:struct}, we can still apply several neurosymbolic techniques by learning the symbolic knowledge when this is not the case. The unknown symbolic knowledge is then the actual target to be learned. A classical example is \textit{program synthesis}, where the goal is to learn the program from positive and negative examples of the desired input-output behaviour. Ideally, all positive pairs and none of the negatives should be covered. Many systems learn logic programs, i.e. NTPs \cite{rocktaschel2017ntp}, $\partial$ILP \cite{evans:dilp}, DeepProbLog\cite{manhaeve2018deepproblog}, NeuralLP\cite{Cohen_NeuralLP}, DiffLog\cite{si2019difflog}, DeepCoder\cite{deepcoder}.
     Sometimes, the input-output pairs are not part of the training dataset, but are actually generated by a black-box neural model. The induced programs  then explain the behaviour of the model, which relates NeSy to the domain of \textit{explainability} \cite{ciravegna2023logic}.}

	\section{Open Challenges}
	To conclude, we list some interesting challenges for \nesy{}.

	\paragraph{Semantics} The statistical relational AI  and  probabilistic graphical model communities have devoted a lot of attention to the semantics of its models. This has resulted in several clear choices (such as directed vs. undirected, trace-based vs. possible world \cite{russell2015unifying}), with corresponding strengths and weaknesses that clarify the relationships between the different models. Workshops have been held on this topic\footnote{For instance, \url{https://pps2018.luddy.indiana.edu/}}. Furthermore, some researchers have investigated how to transform one type of model into another \cite{jaeger2008model}. At the same time,  weighted model counting has emerged as a common assembly language for inference. The situation in neurosymbolic computation today is very much like that of the early days in statistical relational learning, in which there were many competing formalisms, sometimes characterized as the statistical relational learning alphabet soup. It would be great to get more insight into the semantics of neurosymbolic approaches and their relationships.  This survey hopes to contribute towards this goal.
	
	\paragraph{Probabilistic reasoning}
	Although relatively few methods explore the integration of  logical and neural methods from a probabilistic perspective, we believe that a probabilistic approach is a very natural way to integrate the two, since it has been shown \cite{de2019neuro} how one can recover the single methods as special cases. However, many open questions remain. Probabilistic inference is computationally expensive, usually requiring approximations. It would be interesting to determine exactly how probabilistic approximate inference compares with other approximations based on relaxations of the logic, like fuzzy logic.

	
	\paragraph{Fuzzy semantics}
	The selection of the t-norm fuzzy logic and the corresponding translation of the connectives is very heterogeneous in the literature. It is often unclear which properties of Boolean logic a model is preserving, while there is a tendency to consider fuzzy logic as a continuous surrogate of Boolean logic without considering implications for the semantics. There is a clear need for further studies in this field. On the one hand, one may want to define new models which are natively fuzzy, thus not requiring a translation from  Boolean logic. On the other hand, an interesting research direction concerns the characterisation of what are appropriate fuzzy approximations of Boolean logic relative to a set of properties that one wants to preserve (see Section \ref{sec:semantics}).

	\paragraph{Structure learning}
	While significant progress has been made on learning the structure of purely relational models (without probabilities), learning StarAI models remains a major challenge due to the complexity of inference and the combinatorial nature of the problem.
	Incorporating neural aspects complicates the problem even more.
	\nesy{} methods have certainly shown potential for addressing this problem (Section \ref{sec:struct}), but the existing methods are still limited and mostly domain-specific which impedes their wide application.
	For instance, the current systems that support structure learning require user effort to specify the clause templates or write a sketch of a model.
	
	
	\paragraph{Scaling inference}
	Scalable inference is a major challenge for StarAI and therefore also for \nesy{} approaches with an explicit logical or probabilistic reasoning component.
	Investigating to what extent neural methods can help with this challenge by means of lifted (exploiting symmetries in models) or approximate inference, as well as reasoning from  intermediate representations \cite{abboud2020learning}, are promising future research directions.
	
	\paragraph{Data efficiency}
	A major advantage of StarAI methods, as compared to neural ones, is their data efficiency -- StarAI methods can efficiently learn from small amounts of data, whereas neural methods are data hungry.
	On the other hand, StarAI methods do not scale to big data sets, while neural methods can easily handle them.
	We believe that understanding how these methods can help each other to overcome their complementary weaknesses, is a promising research direction.

	\paragraph{Symbolic representation learning}
	The effectiveness of deep learning  comes from the ability to change the representation of the data so that the target task becomes easier to solve.
	The ability to change the representation also at the symbolic level would significantly increase the capabilities of \nesy{} systems.
	This is a major open challenge for which neurally inspired methods could help achieve progress \cite{crop:playgol,dumancic:encoding}.
	
	
	

	\section*{Acknowledgements}
	This work has received funding from the Research Foundation-Flanders (FWO)
	(G. Marra: 1239422N, S. Dumančić: 12ZE520N, R. Manhaeve: 1S61718N). 
	Luc De Raedt has received funding from the Flemish Government (AI Research Program), from the FWO, from the European Research Council (ERC) under the European Union’s Horizon 2020 research and innovation programme (grant agreement No 694980 SYNTH: Synthesising Inductive Data Models) and the Wallenberg AI, Autonomous Systems and Software Program (WASP) funded by the Knut and Alice Wallenberg Foundation. This work was also supported by TAILOR, a project funded by EU Horizon 2020 research and innovation programme under GA No 952215.

	\bibliographystyle{plain}
	\bibliography{main}

\begin{thebibliography}{100}

\bibitem{abboud2020learning}
Ralph Abboud, {\.I}smail~{\.I}lkan Ceylan, and Thomas Lukasiewicz.
\newblock Learning to reason: Leveraging neural networks for approximate {DNF}
  counting.
\newblock In {\em 34th Conference on Artificial Intelligence, {AAAI} 2020, New
  York, NY, USA, February 7-12, 2020}, pages 3097--3104. {AAAI} Press, 2020.

\bibitem{semanticreferee}
Marjan Alirezaie, Martin L{\"{a}}ngkvist, Michael Sioutis, and Amy Loutfi.
\newblock Semantic referee: {A} neural-symbolic framework for enhancing
  geospatial semantic segmentation.
\newblock {\em Semantic Web}, 10(5):863--880, 2019.

\bibitem{bach2017psl}
Stephen~H. Bach, Matthias Broecheler, Bert Huang, and Lise Getoor.
\newblock Hinge-loss markov random fields and probabilistic soft logic.
\newblock {\em J. Mach. Learn. Res.}, 18:109:1--109:67, 2017.

\bibitem{bader2005dimensions}
Sebastian Bader and Pascal Hitzler.
\newblock Dimensions of neural-symbolic integration - {A} structured survey.
\newblock In Sergei~N. Art{\"{e}}mov, Howard Barringer, Artur~S. d'Avila
  Garcez, Lu{\'{\i}}s~C. Lamb, and John Woods, editors, {\em We Will Show Them!
  Essays in Honour of Dov Gabbay, Volume One}, pages 167--194. College
  Publications, 2005.

\bibitem{ltn2021aij}
Samy Badreddine, Artur~S. d'Avila Garcez, Luciano Serafini, and Michael
  Spranger.
\newblock Logic tensor networks.
\newblock {\em Artif. Intell.}, 303:103649, 2022.

\bibitem{deepcoder}
Matej Balog, Alexander~L. Gaunt, Marc Brockschmidt, Sebastian Nowozin, and
  Daniel Tarlow.
\newblock Deepcoder: Learning to write programs.
\newblock In {\em 5th International Conference on Learning Representations,
  {ICLR} 2017, Toulon, France, April 24-26, 2017}. OpenReview.net, 2017.

\bibitem{barcelo2019logical}
Pablo Barcel{\'{o}}, Egor~V. Kostylev, Mika{\"{e}}l Monet, Jorge P{\'{e}}rez,
  Juan~L. Reutter, and Juan~Pablo Silva.
\newblock The logical expressiveness of graph neural networks.
\newblock In {\em 8th International Conference on Learning Representations,
  {ICLR} 2020, Addis Ababa, Ethiopia, April 26-30, 2020}. OpenReview.net, 2020.

\bibitem{battaglia2018relational}
Peter~W. Battaglia, Jessica~B. Hamrick, Victor Bapst, Alvaro
  Sanchez{-}Gonzalez, Vin{\'{\i}}cius~Flores Zambaldi, Mateusz Malinowski,
  Andrea Tacchetti, David Raposo, Adam Santoro, Ryan Faulkner, {\c{C}}aglar
  G{\"{u}}l{\c{c}}ehre, H.~Francis Song, Andrew~J. Ballard, Justin Gilmer,
  George~E. Dahl, Ashish Vaswani, Kelsey~R. Allen, Charles Nash, Victoria
  Langston, Chris Dyer, Nicolas Heess, Daan Wierstra, Pushmeet Kohli, Matthew
  Botvinick, Oriol Vinyals, Yujia Li, and Razvan Pascanu.
\newblock Relational inductive biases, deep learning, and graph networks.
\newblock {\em CoRR}, abs/1806.01261, 2018.

\bibitem{bvelohlavek2017fuzzy}
Radim B{\v{e}}lohl{\'a}vek, Joseph~W Dauben, and George~J Klir.
\newblock {\em Fuzzy logic and mathematics: a historical perspective}.
\newblock Oxford University Press, 2017.

\bibitem{aidebate}
Yoshua Bengio and Gary Marcus.
\newblock Ai debate: The best way forward for ai
  \url{https://montrealartificialintelligence.com/aidebate/}, (checked on 13
  december 2021), 2020.

\bibitem{Dagstuhl17192}
Tarek~R. Besold, Artur d'Avila Garcez, and Luis~C. Lamb.
\newblock {Human-Like Neural-Symbolic Computing (Dagstuhl Seminar 17192)}.
\newblock {\em Dagstuhl Reports}, 7(5):56--83, 2017.

\bibitem{besold2017neural}
Tarek~R. Besold, Artur~S. d'Avila Garcez, Sebastian Bader, Howard Bowman,
  Pedro~M. Domingos, Pascal Hitzler, Kai{-}Uwe K{\"{u}}hnberger, Lu{\'{\i}}s~C.
  Lamb, Daniel Lowd, Priscila Machado~Vieira Lima, Leo de~Penning, Gadi Pinkas,
  Hoifung Poon, and Gerson Zaverucha.
\newblock Neural-symbolic learning and reasoning: {A} survey and
  interpretation.
\newblock {\em CoRR}, abs/1711.03902, 2017.

\bibitem{bordes2013translating}
Antoine Bordes, Nicolas Usunier, Alberto Garc{\'{\i}}a{-}Dur{\'{a}}n, Jason
  Weston, and Oksana Yakhnenko.
\newblock Translating embeddings for modeling multi-relational data.
\newblock In {\em Advances in Neural Information Processing Systems 27, NeurIPS
  2013, December 5-8, 2013, Lake Tahoe, Nevada}, pages 2787--2795, 2013.

\bibitem{BosnjakDFI}
Matko Bosnjak, Tim Rockt{\"{a}}schel, Jason Naradowsky, and Sebastian Riedel.
\newblock Programming with a differentiable forth interpreter.
\newblock In Doina Precup and Yee~Whye Teh, editors, {\em 34th International
  Conference on Machine Learning, {ICML} 2017, Sydney, NSW, Australia, 6-11
  August 2017}, volume~70 of {\em Proceedings of Machine Learning Research},
  pages 547--556. {PMLR}, 2017.

\bibitem{chapelle2006semi}
Olivier Chapelle, Bernhard Sch{\"o}lkopf, and Alexander Zien.
\newblock {\em Semi-supervised learning}.
\newblock The MIT Press Cambridge, MA, USA:, 2006.

\bibitem{chaudhuri2021neurosymbolic}
Swarat Chaudhuri, Kevin Ellis, Oleksandr Polozov, Rishabh Singh, Armando
  Solar{-}Lezama, and Yisong Yue.
\newblock Neurosymbolic programming.
\newblock {\em Found. Trends Program. Lang.}, 7(3):158--243, 2021.

\bibitem{ciravegna2023logic}
Gabriele Ciravegna, Pietro Barbiero, Francesco Giannini, Marco Gori, Pietro
  Li{\`{o}}, Marco Maggini, and Stefano Melacci.
\newblock Logic explained networks.
\newblock {\em Artif. Intell.}, 314:103822, 2023.

\bibitem{cohen2017tensorlog}
William~W. Cohen, Fan Yang, and Kathryn Mazaitis.
\newblock Tensorlog: Deep learning meets probabilistic dbs.
\newblock {\em CoRR}, abs/1707.05390, 2017.

\bibitem{crop:playgol}
Andrew Cropper.
\newblock Playgol: Learning programs through play.
\newblock In Sarit Kraus, editor, {\em 28th International Joint Conference on
  Artificial Intelligence, {IJCAI} 2019, Macao, China, August 10-16, 2019},
  pages 6074--6080. ijcai.org, 2019.

\bibitem{metagol}
Andrew Cropper and Stephen~H. Muggleton.
\newblock Metagol system.
\newblock https://github.com/metagol/metagol, 2016.

\bibitem{cussens2001parameter}
James Cussens.
\newblock Parameter estimation in stochastic logic programs.
\newblock {\em Mach. Learn.}, 44(3):245--271, 2001.

\bibitem{dai2019abl}
Wang{-}Zhou Dai, Qiu{-}Ling Xu, Yang Yu, and Zhi{-}Hua Zhou.
\newblock Bridging machine learning and logical reasoning by abductive
  learning.
\newblock In Hanna~M. Wallach, Hugo Larochelle, Alina Beygelzimer, Florence
  d'Alch{\'{e}}{-}Buc, Emily~B. Fox, and Roman Garnett, editors, {\em Advances
  in Neural Information Processing Systems 32, NeurIPS 2019, December 8-14,
  2019, Vancouver, BC, Canada}, pages 2811--2822, 2019.

\bibitem{ddnnf}
Adnan Darwiche.
\newblock On the tractable counting of theory models and its application to
  truth maintenance and belief revision.
\newblock {\em J. Appl. Non Class. Logics}, 11(1-2):11--34, 2001.

\bibitem{darwiche2011sdd}
Adnan Darwiche.
\newblock {SDD:} {A} new canonical representation of propositional knowledge
  bases.
\newblock In Toby Walsh, editor, {\em 22nd International Joint Conference on
  Artificial Intelligence, {IJCAI} 2011, Barcelona, Catalonia, Spain, July
  16-22, 2011}, pages 819--826. {IJCAI/AAAI}, 2011.

\bibitem{DarwicheMarquis}
Adnan Darwiche and Pierre Marquis.
\newblock A knowledge compilation map.
\newblock {\em Journal of Artificial Intelligence Research}, 17:229--264, 2002.

\bibitem{dash2021tell}
Tirtharaj Dash, Sharad Chitlangia, Aditya Ahuja, and Ashwin Srinivasan.
\newblock How to tell deep neural networks what we know.
\newblock {\em CoRR}, abs/2107.10295, 2021.

\bibitem{Dagstuhl14381}
Artur~S. d'Avila Garcez, Marco Gori, Pascal Hitzler, and Lu{\'{\i}}s~C. Lamb.
\newblock Neural-symbolic learning and reasoning (dagstuhl seminar 14381).
\newblock {\em Dagstuhl Reports}, 4(9):50--84, 2014.

\bibitem{garcez2019neural}
Artur~S. d'Avila Garcez, Marco Gori, Lu{\'{\i}}s~C. Lamb, Luciano Serafini,
  Michael Spranger, and Son~N. Tran.
\newblock Neural-symbolic computing: An effective methodology for principled
  integration of machine learning and reasoning.
\newblock {\em {FLAP}}, 6(4):611--632, 2019.

\bibitem{garcez2006connectionist}
Artur~S. d'Avila Garcez and Lu{\'{\i}}s~C. Lamb.
\newblock A connectionist computational model for epistemic and temporal
  reasoning.
\newblock {\em Neural Comput.}, 18(7):1711--1738, 2006.

\bibitem{garcez2007connectionist}
Artur~S. d'Avila Garcez, Lu{\'{\i}}s~C. Lamb, and Dov~M. Gabbay.
\newblock Connectionist modal logic: Representing modalities in neural
  networks.
\newblock {\em Theor. Comput. Sci.}, 371(1-2):34--53, 2007.

\bibitem{raedt:problog}
Luc {De Raedt}, Angelika Kimmig, and Hannu Toivonen.
\newblock Problog: {A} probabilistic prolog and its application in link
  discovery.
\newblock In {\em 20th International Joint Conference on Artificial
  Intelligence, {IJCAI} 2007, Hyderabad, India, January 6-12, 2007}, pages
  2462--2467, 2007.

\bibitem{demeester2016lifted}
Thomas Demeester, Tim Rockt{\"{a}}schel, and Sebastian Riedel.
\newblock Lifted rule injection for relation embeddings.
\newblock In Jian Su, Xavier Carreras, and Kevin Duh, editors, {\em Conference
  on Empirical Methods in Natural Language Processing, {EMNLP} 2016, Austin,
  Texas, USA, November 1-4, 2016}, pages 1389--1399. The Association for
  Computational Linguistics, 2016.

\bibitem{diligenti2017sbr}
Michelangelo Diligenti, Marco Gori, and Claudio Sacc{\`{a}}.
\newblock Semantic-based regularization for learning and inference.
\newblock {\em Artif. Intell.}, 244:143--165, 2017.

\bibitem{donadello2017ltn}
Ivan Donadello, Luciano Serafini, and Artur~S. d'Avila Garcez.
\newblock Logic tensor networks for semantic image interpretation.
\newblock In Carles Sierra, editor, {\em 26th International Joint Conference on
  Artificial Intelligence, {IJCAI} 2017, Melbourne, Australia, August 19-25,
  2017}, pages 1596--1602. ijcai.org, 2017.

\bibitem{NLM}
Honghua Dong, Jiayuan Mao, Tian Lin, Chong Wang, Lihong Li, and Denny Zhou.
\newblock Neural logic machines.
\newblock In {\em 7th International Conference on Learning Representations,
  {ICLR} 2019, New Orleans, LA, USA, May 6-9, 2019}. OpenReview.net, 2019.

\bibitem{dumancic:encoding}
Sebastijan Dumancic, Tias Guns, Wannes Meert, and Hendrik Blockeel.
\newblock Learning relational representations with auto-encoding logic
  programs.
\newblock In Sarit Kraus, editor, {\em 28th International Joint Conference on
  Artificial Intelligence, {IJCAI} 2019, Macao, China, August 10-16, 2019},
  pages 6081--6087. ijcai.org, 2019.

\bibitem{ellis:libraries}
Kevin Ellis, Lucas Morales, Mathias Sabl{\'{e}}{-}Meyer, Armando
  Solar{-}Lezama, and Josh Tenenbaum.
\newblock Learning libraries of subroutines for neurally-guided bayesian
  program induction.
\newblock In Samy Bengio, Hanna~M. Wallach, Hugo Larochelle, Kristen Grauman,
  Nicol{\`{o}} Cesa{-}Bianchi, and Roman Garnett, editors, {\em Advances in
  Neural Information Processing Systems 31, NeurIPS 2018, December 3-8, 2018,
  Montr{\'{e}}al, Canada}, pages 7816--7826, 2018.

\bibitem{ellis:repl}
Kevin Ellis, Maxwell~I. Nye, Yewen Pu, Felix Sosa, Josh Tenenbaum, and Armando
  Solar{-}Lezama.
\newblock Write, execute, assess: Program synthesis with a {REPL}.
\newblock {\em CoRR}, abs/1906.04604, 2019.

\bibitem{evans:dilp}
Richard Evans and Edward Grefenstette.
\newblock Learning explanatory rules from noisy data.
\newblock {\em J. Artif. Intell. Res.}, 61:1--64, 2018.

\bibitem{fisher2019training}
Marc Fischer, Mislav Balunovic, Dana Drachsler{-}Cohen, Timon Gehr, Ce~Zhang,
  and Martin~T. Vechev.
\newblock {DL2:} training and querying neural networks with logic.
\newblock In Kamalika Chaudhuri and Ruslan Salakhutdinov, editors, {\em 36th
  International Conference on Machine Learning, {ICML} 2019, 9-15 June 2019,
  Long Beach, California, {USA}}, volume~97 of {\em Proceedings of Machine
  Learning Research}, pages 1931--1941. {PMLR}, 2019.

\bibitem{flach:simplylogical}
Peter~A. Flach.
\newblock {\em Simply logical - intelligent reasoning by example}.
\newblock Wiley professional computing. Wiley, 1994.

\bibitem{frasconi2014klog}
Paolo Frasconi, Fabrizio Costa, Luc~De Raedt, and Kurt~De Grave.
\newblock klog: {A} language for logical and relational learning with kernels.
\newblock {\em Artif. Intell.}, 217:117--143, 2014.

\bibitem{friedman1999learning}
Nir Friedman, Lise Getoor, Daphne Koller, and Avi Pfeffer.
\newblock Learning probabilistic relational models.
\newblock In Thomas Dean, editor, {\em 16th International Joint Conference on
  Artificial Intelligence, {IJCAI} 99, Stockholm, Sweden, July 31 - August 6,
  1999. 2 Volumes, 1450 pages}, pages 1300--1309. Morgan Kaufmann, 1999.

\bibitem{Getoor07:book}
L.~Getoor and B.~Taskar, editors.
\newblock {\em An Introduction to Statistical Relational Learning}.
\newblock {MIT} {P}ress, 2007.

\bibitem{giannini2018convex}
Francesco Giannini, Michelangelo Diligenti, Marco Gori, and Marco Maggini.
\newblock On a convex logic fragment for learning and reasoning.
\newblock {\em {IEEE} Trans. Fuzzy Syst.}, 27(7):1407--1416, 2019.

\bibitem{gilmer2017neural}
Justin Gilmer, Samuel~S. Schoenholz, Patrick~F. Riley, Oriol Vinyals, and
  George~E. Dahl.
\newblock Neural message passing for quantum chemistry.
\newblock In Doina Precup and Yee~Whye Teh, editors, {\em 34th International
  Conference on Machine Learning, {ICML} 2017, Sydney, NSW, Australia, 6-11
  August 2017}, volume~70 of {\em Proceedings of Machine Learning Research},
  pages 1263--1272. {PMLR}, 2017.

\bibitem{grohe2021logic}
Martin Grohe.
\newblock The logic of graph neural networks.
\newblock In {\em 36th Annual {ACM/IEEE} Symposium on Logic in Computer
  Science, {LICS} 2021, Rome, Italy, June 29 - July 2, 2021}, pages 1--17.
  {IEEE}, 2021.

\bibitem{guo2016jointly}
Shu Guo, Quan Wang, Lihong Wang, Bin Wang, and Li~Guo.
\newblock Jointly embedding knowledge graphs and logical rules.
\newblock In Jian Su, Xavier Carreras, and Kevin Duh, editors, {\em Conference
  on Empirical Methods in Natural Language Processing, {EMNLP} 2016, Austin,
  Texas, USA, November 1-4, 2016}, pages 192--202. The Association for
  Computational Linguistics, 2016.

\bibitem{problogWeights}
Bernd Gutmann, Angelika Kimmig, Kristian Kersting, and Luc~De Raedt.
\newblock Parameter learning in probabilistic databases: {A} least squares
  approach.
\newblock In Walter Daelemans, Bart Goethals, and Katharina Morik, editors,
  {\em European Conference on Machine Learning and Knowledge Discovery in
  Databases, {ECML/PKDD} 2008, Antwerp, Belgium, September 15-19, 2008,
  Proceedings, Part {I}}, volume 5211 of {\em Lecture Notes in Computer
  Science}, pages 473--488. Springer, 2008.

\bibitem{GutmannTR11}
Bernd Gutmann, Ingo Thon, and Luc~De Raedt.
\newblock Learning the parameters of probabilistic logic programs from
  interpretations.
\newblock In Dimitrios Gunopulos, Thomas Hofmann, Donato Malerba, and Michalis
  Vazirgiannis, editors, {\em European Conference on Machine Learning, {ECML}
  {PKDD} 2011, Athens, Greece, September 5-9, 2011. Proceedings, Part {I}},
  volume 6911 of {\em Lecture Notes in Computer Science}, pages 581--596.
  Springer, 2011.

\bibitem{hahn2021teaching}
Christopher Hahn, Frederik Schmitt, Jens~U. Kreber, Markus~Norman Rabe, and
  Bernd Finkbeiner.
\newblock Teaching temporal logics to neural networks.
\newblock In {\em 9th International Conference on Learning Representations,
  {ICLR} 2021, Virtual Event, Austria, May 3-7, 2021}. OpenReview.net, 2021.

\bibitem{halpern1990analysis}
Joseph~Y. Halpern.
\newblock An analysis of first-order logics of probability.
\newblock {\em Artif. Intell.}, 46(3):311--350, 1990.

\bibitem{halpern2017reasoning}
Joseph~Y. Halpern.
\newblock {\em Reasoning about uncertainty}.
\newblock {MIT} Press, 2005.

\bibitem{hamilton2018embedding}
William~L. Hamilton, Payal Bajaj, Marinka Zitnik, Dan Jurafsky, and Jure
  Leskovec.
\newblock Embedding logical queries on knowledge graphs.
\newblock In Samy Bengio, Hanna~M. Wallach, Hugo Larochelle, Kristen Grauman,
  Nicol{\`{o}} Cesa{-}Bianchi, and Roman Garnett, editors, {\em Advances in
  Neural Information Processing Systems 31, NeurIPS 2018, December 3-8, 2018,
  Montr{\'{e}}al, Canada}, pages 2030--2041, 2018.

\bibitem{hinton2015distilling}
Geoffrey~E. Hinton, Oriol Vinyals, and Jeffrey Dean.
\newblock Distilling the knowledge in a neural network.
\newblock {\em CoRR}, abs/1503.02531, 2015.

\bibitem{hochreiter2022toward}
Sepp Hochreiter.
\newblock Toward a broad {AI}.
\newblock {\em Commun. {ACM}}, 65(4):56--57, 2022.

\bibitem{hu2016harnessing}
Zhiting Hu, Xuezhe Ma, Zhengzhong Liu, Eduard~H. Hovy, and Eric~P. Xing.
\newblock Harnessing deep neural networks with logic rules.
\newblock In {\em 54th Annual Meeting of the Association for Computational
  Linguistics, {ACL} 2016, August 7-12, 2016, Berlin, Germany, Volume 1: Long
  Papers}. The Association for Computer Linguistics, 2016.

\bibitem{huang2021scallop}
Jiani Huang, Ziyang Li, Binghong Chen, Karan Samel, Mayur Naik, Le~Song, and
  Xujie Si.
\newblock Scallop: From probabilistic deductive databases to scalable
  differentiable reasoning.
\newblock pages 25134--25145, 2021.

\bibitem{jaeger2008model}
Manfred Jaeger.
\newblock Model-theoretic expressivity analysis.
\newblock In Luc~De Raedt, Paolo Frasconi, Kristian Kersting, and Stephen~H.
  Muggleton, editors, {\em Probabilistic Inductive Logic Programming - Theory
  and Applications}, volume 4911 of {\em Lecture Notes in Computer Science},
  pages 325--339. Springer, 2008.

\bibitem{NGSynth}
Ashwin Kalyan, Abhishek Mohta, Oleksandr Polozov, Dhruv Batra, Prateek Jain,
  and Sumit Gulwani.
\newblock Neural-guided deductive search for real-time program synthesis from
  examples.
\newblock In {\em 6th International Conference on Learning Representations,
  {ICLR} 2018, Vancouver, BC, Canada, April 30 - May 3, 2018}. OpenReview.net,
  2018.

\bibitem{kersting20071}
Kristian Kersting and Luc De~Raedt.
\newblock Bayesian logic programming: Theory and tool.
\newblock In L.~Getoor and B.~Taskar, editors, {\em An introduction to
  Statistical Relational Learning}. MIT Press, 2007.

\bibitem{kingma2014adam}
Diederik~P. Kingma and Jimmy Ba.
\newblock Adam: {A} method for stochastic optimization.
\newblock 2015.

\bibitem{gcn}
Thomas~N. Kipf and Max Welling.
\newblock Semi-supervised classification with graph convolutional networks.
\newblock In {\em 5th International Conference on Learning Representations,
  {ICLR} 2017, Toulon, France, April 24-26, 2017}. OpenReview.net, 2017.

\bibitem{KokStruct}
Stanley Kok and Pedro~M. Domingos.
\newblock Learning the structure of markov logic networks.
\newblock In Luc~De Raedt and Stefan Wrobel, editors, {\em 22nd International
  Conference on Machine Learning, {(ICML} 2005), Bonn, Germany, August 7-11,
  2005}, volume 119 of {\em {ACM} International Conference Proceeding Series},
  pages 441--448. {ACM}, 2005.

\bibitem{Kok:2010}
Stanley Kok and Pedro~M. Domingos.
\newblock Learning markov logic networks using structural motifs.
\newblock In Johannes F{\"{u}}rnkranz and Thorsten Joachims, editors, {\em 27th
  International Conference on Machine Learning (ICML-10), June 21-24, 2010,
  Haifa, Israel}, pages 551--558. Omnipress, 2010.

\bibitem{koller2009probabilistic}
Daphne Koller and Nir Friedman.
\newblock {\em Probabilistic Graphical Models - Principles and Techniques}.
\newblock {MIT} Press, 2009.

\bibitem{lamb2020graph}
Lu{\'{\i}}s~C. Lamb, Artur~S. d'Avila Garcez, Marco Gori, Marcelo O.~R. Prates,
  Pedro H.~C. Avelar, and Moshe~Y. Vardi.
\newblock Graph neural networks meet neural-symbolic computing: {A} survey and
  perspective.
\newblock In Christian Bessiere, editor, {\em 29th International Joint
  Conference on Artificial Intelligence, {IJCAI} 2020}, pages 4877--4884.
  ijcai.org, 2020.

\bibitem{li2020augmenting}
Tao Li and Vivek Srikumar.
\newblock Augmenting neural networks with first-order logic.
\newblock {\em CoRR}, abs/1906.06298, 2019.

\bibitem{lloyd:book}
John~W. Lloyd.
\newblock {\em Foundations of Logic Programming, 2nd Edition}.
\newblock Springer, 1987.

\bibitem{LowdWeights}
Daniel Lowd and Pedro~M. Domingos.
\newblock Efficient weight learning for markov logic networks.
\newblock In Joost~N. Kok, Jacek Koronacki, Ram{\'{o}}n~L{\'{o}}pez
  de~M{\'{a}}ntaras, Stan Matwin, Dunja Mladenic, and Andrzej Skowron, editors,
  {\em 11th European Conference on Principles and Practice of Knowledge
  Discovery in Databases, {PKDD} 2007, Warsaw, Poland, September 17-21, 2007,
  Proceedings}, volume 4702 of {\em Lecture Notes in Computer Science}, pages
  200--211. Springer, 2007.

\bibitem{mandi2022decision}
Jayanta Mandi, V{\'{\i}}ctor Bucarey, Maxime Mulamba~Ke Tchomba, and Tias Guns.
\newblock Decision-focused learning: Through the lens of learning to rank.
\newblock In Kamalika Chaudhuri, Stefanie Jegelka, Le~Song, Csaba
  Szepesv{\'{a}}ri, Gang Niu, and Sivan Sabato, editors, {\em 39th
  International Conference on Machine Learning, {ICML} 2022, 17-23 July 2022,
  Baltimore, Maryland, {USA}}, volume 162 of {\em Proceedings of Machine
  Learning Research}, pages 14935--14947. {PMLR}, 2022.

\bibitem{manhaeve2018deepproblog}
Robin Manhaeve, Sebastijan Dumancic, Angelika Kimmig, Thomas Demeester, and
  Luc~De Raedt.
\newblock Deepproblog: Neural probabilistic logic programming.
\newblock In Samy Bengio, Hanna~M. Wallach, Hugo Larochelle, Kristen Grauman,
  Nicol{\`{o}} Cesa{-}Bianchi, and Roman Garnett, editors, {\em Advances in
  Neural Information Processing Systems 31, NeurIPS 2018, December 3-8, 2018,
  Montr{\'{e}}al, Canada}, pages 3753--3763, 2018.

\bibitem{manhaeve2021approximate}
Robin Manhaeve, Giuseppe Marra, and Luc~De Raedt.
\newblock Approximate inference for neural probabilistic logic programming.
\newblock In Meghyn Bienvenu, Gerhard Lakemeyer, and Esra Erdem, editors, {\em
  18th International Conference on Principles of Knowledge Representation and
  Reasoning, {KR} 2021, Online event, November 3-12, 2021}, pages 475--486,
  2021.

\bibitem{mao2018the}
Jiayuan Mao, Chuang Gan, Pushmeet Kohli, Joshua~B. Tenenbaum, and Jiajun Wu.
\newblock The neuro-symbolic concept learner: Interpreting scenes, words, and
  sentences from natural supervision.
\newblock In {\em 7th International Conference on Learning Representations,
  {ICLR} 2019, New Orleans, LA, USA, May 6-9, 2019}. OpenReview.net, 2019.

\bibitem{phdthesis_giuseppe}
Giuseppe Marra.
\newblock {\em Bridging symbolic and subsymbolic reasoning with MiniMax Entropy
  models.}
\newblock PhD thesis, University of Florence, 2 2020.

\bibitem{marra2020relational}
Giuseppe Marra, Michelangelo Diligenti, Francesco Giannini, Marco Gori, and
  Marco Maggini.
\newblock Relational neural machines.
\newblock In Giuseppe~De Giacomo, Alejandro Catal{\'{a}}, Bistra Dilkina,
  Michela Milano, Sen{\'{e}}n Barro, Alberto Bugar{\'{\i}}n, and
  J{\'{e}}r{\^{o}}me Lang, editors, {\em 24th European Conference on Artificial
  Intelligence, {ECAI} 2020, 29 August-8 September 2020, Santiago de
  Compostela, Spain, August 29 - September 8, 2020}, volume 325 of {\em
  Frontiers in Artificial Intelligence and Applications}, pages 1340--1347.
  {IOS} Press, 2020.

\bibitem{marra2019integrating}
Giuseppe Marra, Francesco Giannini, Michelangelo Diligenti, and Marco Gori.
\newblock Integrating learning and reasoning with deep logic models.
\newblock In Ulf Brefeld, {\'{E}}lisa Fromont, Andreas Hotho, Arno~J. Knobbe,
  Marloes~H. Maathuis, and C{\'{e}}line Robardet, editors, {\em European
  Conference on Machine Learning and Knowledge Discovery in Databases, {ECML}
  {PKDD} 2019, W{\"{u}}rzburg, Germany, September 16-20, 2019, Proceedings,
  Part {II}}, volume 11907 of {\em Lecture Notes in Computer Science}, pages
  517--532. Springer, 2019.

\bibitem{marra2019nmln}
Giuseppe Marra and Ondrej Kuzelka.
\newblock Neural markov logic networks.
\newblock In Cassio~P. de~Campos, Marloes~H. Maathuis, and Erik Quaeghebeur,
  editors, {\em 37th Conference on Uncertainty in Artificial Intelligence,
  {UAI} 2021, Virtual Event, 27-30 July 2021}, volume 161 of {\em Proceedings
  of Machine Learning Research}, pages 908--917. {AUAI} Press, 2021.

\bibitem{minervini2019differentiable}
Pasquale Minervini, Matko Bosnjak, Tim Rockt{\"{a}}schel, Sebastian Riedel, and
  Edward Grefenstette.
\newblock Differentiable reasoning on large knowledge bases and natural
  language.
\newblock In {\em 34th Conference on Artificial Intelligence, {AAAI} 2020, New
  York, NY, USA, February 7-12, 2020}, pages 5182--5190. {AAAI} Press, 2020.

\bibitem{minervini2017adversarial}
Pasquale Minervini, Thomas Demeester, Tim Rockt{\"{a}}schel, and Sebastian
  Riedel.
\newblock Adversarial sets for regularising neural link predictors.
\newblock In Gal Elidan, Kristian Kersting, and Alexander~T. Ihler, editors,
  {\em 33rd Conference on Uncertainty in Artificial Intelligence, {UAI} 2017,
  Sydney, Australia, August 11-15, 2017}. {AUAI} Press, 2017.

\bibitem{minervini2020learning}
Pasquale Minervini, Sebastian Riedel, Pontus Stenetorp, Edward Grefenstette,
  and Tim Rockt{\"{a}}schel.
\newblock Learning reasoning strategies in end-to-end differentiable proving.
\newblock In {\em 37th International Conference on Machine Learning, {ICML}
  2020, 13-18 July 2020, Virtual Event}, volume 119 of {\em Proceedings of
  Machine Learning Research}, pages 6938--6949. {PMLR}, 2020.

\bibitem{misino2022vael}
Eleonora Misino, Giuseppe Marra, and Emanuele Sansone.
\newblock {VAEL:} bridging variational autoencoders and probabilistic logic
  programming.
\newblock In S.~Koyejo, S.~Mohamed, A.~Agarwal, D.~Belgrave, K.~Cho, and A.~Oh,
  editors, {\em Advances in Neural Information Processing Systems 35, NeurIPS
  2022, November 29 December 4, 2022, New Orleans, Luisiana}, pages 4667--4679,
  2022.

\bibitem{Mitchell82}
Tom~M. Mitchell.
\newblock Generalization as search.
\newblock {\em Artif. Intell.}, 18(2):203--226, 1982.

\bibitem{morris2019weisfeiler}
Christopher Morris, Martin Ritzert, Matthias Fey, William~L. Hamilton, Jan~Eric
  Lenssen, Gaurav Rattan, and Martin Grohe.
\newblock Weisfeiler and leman go neural: Higher-order graph neural networks.
\newblock In {\em 33rd Conference on Artificial Intelligence, {AAAI} 2019,
  Honolulu, Hawaii, USA, January 27 - February 1, 2019}, pages 4602--4609.
  {AAAI} Press, 2019.

\bibitem{muggleton1996stochastic}
Stephen Muggleton.
\newblock Stochastic logic programs.
\newblock {\em Advances in inductive logic programming}, 32, 1996.

\bibitem{mugg:ilp94}
Stephen Muggleton and Luc~De Raedt.
\newblock Inductive logic programming: Theory and methods.
\newblock {\em J. Log. Program.}, 19/20:629--679, 1994.

\bibitem{nathani2019learning}
Deepak Nathani, Jatin Chauhan, Charu Sharma, and Manohar Kaul.
\newblock Learning attention-based embeddings for relation prediction in
  knowledge graphs.
\newblock In Anna Korhonen, David~R. Traum, and Llu{\'{\i}}s M{\`{a}}rquez,
  editors, {\em 57th Conference of the Association for Computational
  Linguistics, {ACL} 2019, Florence, Italy, July 28- August 2, 2019, Volume 1:
  Long Papers}, pages 4710--4723. Association for Computational Linguistics,
  2019.

\bibitem{kg_srl}
Maximilian Nickel, Kevin Murphy, Volker Tresp, and Evgeniy Gabrilovich.
\newblock A review of relational machine learning for knowledge graphs.
\newblock {\em Proc. {IEEE}}, 104(1):11--33, 2016.

\bibitem{nickel2011three}
Maximilian Nickel, Volker Tresp, and Hans{-}Peter Kriegel.
\newblock A three-way model for collective learning on multi-relational data.
\newblock In Lise Getoor and Tobias Scheffer, editors, {\em 28th International
  Conference on Machine Learning, {ICML} 2011, Bellevue, Washington, USA, June
  28 - July 2, 2011}, pages 809--816. Omnipress, 2011.

\bibitem{nye:neurips20}
Maxwell~I. Nye, Armando Solar{-}Lezama, Josh Tenenbaum, and Brenden~M. Lake.
\newblock Learning compositional rules via neural program synthesis.
\newblock In Hugo Larochelle, Marc'Aurelio Ranzato, Raia Hadsell,
  Maria{-}Florina Balcan, and Hsuan{-}Tien Lin, editors, {\em Advances in
  Neural Information Processing Systems 33, NeurIPS 2020, December 6-12, 2020,
  virtual}, 2020.

\bibitem{pearl1988probabilistic}
Judea Pearl.
\newblock {\em Probabilistic reasoning in intelligent systems - networks of
  plausible inference}.
\newblock Morgan Kaufmann series in representation and reasoning. Morgan
  Kaufmann, 1989.

\bibitem{poole1993}
David Poole.
\newblock Probabilistic horn abduction and bayesian networks.
\newblock {\em Artif. Intell.}, 64(1):81--129, 1993.

\bibitem{qu2019probabilistic}
Meng Qu and Jian Tang.
\newblock Probabilistic logic neural networks for reasoning.
\newblock In Hanna~M. Wallach, Hugo Larochelle, Alina Beygelzimer, Florence
  d'Alch{\'{e}}{-}Buc, Emily~B. Fox, and Roman Garnett, editors, {\em Advances
  in Neural Information Processing Systems 32, NeurIPS 2019, December 8-14,
  2019, Vancouver, BC, Canada}, pages 7710--7720, 2019.

\bibitem{luc:book}
Luc~De Raedt.
\newblock {\em Logical and relational learning}.
\newblock Cognitive Technologies. Springer, 2008.

\bibitem{probfoil}
Luc~De Raedt, Anton Dries, Ingo Thon, Guy~Van den Broeck, and Mathias Verbeke.
\newblock Inducing probabilistic relational rules from probabilistic examples.
\newblock In Qiang Yang and Michael~J. Wooldridge, editors, {\em 24th
  International Joint Conference on Artificial Intelligence, {IJCAI} 2015,
  Buenos Aires, Argentina, July 25-31, 2015}, pages 1835--1843. {AAAI} Press,
  2015.

\bibitem{DeRaedtKerstingEtAl16}
Luc~De Raedt, Kristian Kersting, Sriraam Natarajan, and David Poole.
\newblock {\em Statistical Relational Artificial Intelligence: Logic,
  Probability, and Computation}.
\newblock Synthesis Lectures on Artificial Intelligence and Machine Learning.
  Morgan {\&} Claypool Publishers, 2016.

\bibitem{de2015probabilistic}
Luc~De Raedt and Angelika Kimmig.
\newblock Probabilistic (logic) programming concepts.
\newblock {\em Mach. Learn.}, 100(1):5--47, 2015.

\bibitem{de2019neuro}
Luc~De Raedt, Robin Manhaeve, Sebastijan Dumancic, Thomas Demeester, and
  Angelika Kimmig.
\newblock Neuro-symbolic = neural + logical + probabilistic.
\newblock In Derek Doran, Artur~S. d'Avila Garcez, and Freddy
  L{\'{e}}cu{\'{e}}, editors, {\em International Workshop on Neural-Symbolic
  Learning and Reasoning (NeSy 2019), Annual workshop of the Neural-Symbolic
  Learning and Reasoning Association, Macao, China, August 12, 2019}, 2019.

\bibitem{ren2020beta}
Hongyu Ren and Jure Leskovec.
\newblock Beta embeddings for multi-hop logical reasoning in knowledge graphs.
\newblock In Hugo Larochelle, Marc'Aurelio Ranzato, Raia Hadsell,
  Maria{-}Florina Balcan, and Hsuan{-}Tien Lin, editors, {\em Advances in
  Neural Information Processing Systems 33, NeurIPS 2020, December 6-12, 2020,
  virtual}, 2020.

\bibitem{richardson2006mln}
Matthew Richardson and Pedro~M. Domingos.
\newblock Markov logic networks.
\newblock {\em Mach. Learn.}, 62(1-2):107--136, 2006.

\bibitem{riegel2020logical}
Ryan Riegel, Alexander~G. Gray, Francois P.~S. Luus, Naweed Khan, Ndivhuwo
  Makondo, Ismail~Yunus Akhalwaya, Haifeng Qian, Ronald Fagin, Francisco
  Barahona, Udit Sharma, Shajith Ikbal, Hima Karanam, Sumit Neelam, Ankita
  Likhyani, and Santosh~K. Srivastava.
\newblock Logical neural networks.
\newblock {\em CoRR}, abs/2006.13155, 2020.

\bibitem{rocktaschel2017ntp}
Tim Rockt{\"{a}}schel and Sebastian Riedel.
\newblock End-to-end differentiable proving.
\newblock In Isabelle Guyon, Ulrike von Luxburg, Samy Bengio, Hanna~M. Wallach,
  Rob Fergus, S.~V.~N. Vishwanathan, and Roman Garnett, editors, {\em Advances
  in Neural Information Processing Systems 30, NeurIPA 2017, December 4-9,
  2017, Long Beach, CA, {USA}}, pages 3788--3800, 2017.

\bibitem{rocktaschel2015injecting}
Tim Rockt{\"{a}}schel, Sameer Singh, and Sebastian Riedel.
\newblock Injecting logical background knowledge into embeddings for relation
  extraction.
\newblock In Rada Mihalcea, Joyce~Yue Chai, and Anoop Sarkar, editors, {\em
  Conference of the North American Chapter of the Association for Computational
  Linguistics: Human Language Technologies, {NAACL} {HLT} 2015 , Denver,
  Colorado, USA, May 31 - June 5, 2015}, pages 1119--1129. The Association for
  Computational Linguistics, 2015.

\bibitem{russell2015unifying}
Stuart~J. Russell.
\newblock Unifying logic and probability.
\newblock {\em Commun. {ACM}}, 58(7):88--97, 2015.

\bibitem{sato1995distributionsemantics}
Taisuke Sato.
\newblock A statistical learning method for logic programs with distribution
  semantics.
\newblock In Leon Sterling, editor, {\em 12nd International Conference on Logic
  Programming, Tokyo, Japan, June 13-16, 1995}, pages 715--729. {MIT} Press,
  1995.

\bibitem{prism}
Taisuke Sato and Yoshitaka Kameya.
\newblock {PRISM:} {A} language for symbolic-statistical modeling.
\newblock In {\em 15th International Joint Conference on Artificial
  Intelligence, {IJCAI} 97, Nagoya, Japan, August 23-29, 1997, 2 Volumes},
  pages 1330--1339. Morgan Kaufmann, 1997.

\bibitem{scarselli2008graph}
Franco Scarselli, Marco Gori, Ah~Chung Tsoi, Markus Hagenbuchner, and Gabriele
  Monfardini.
\newblock The graph neural network model.
\newblock {\em IEEE transactions on neural networks}, 20(1):61--80, 2008.

\bibitem{schlichtkrull2018modeling}
Michael~Sejr Schlichtkrull, Thomas~N. Kipf, Peter Bloem, Rianne van~den Berg,
  Ivan Titov, and Max Welling.
\newblock Modeling relational data with graph convolutional networks.
\newblock In Aldo Gangemi, Roberto Navigli, Maria{-}Esther Vidal, Pascal
  Hitzler, Rapha{\"{e}}l Troncy, Laura Hollink, Anna Tordai, and Mehwish Alam,
  editors, {\em 15th International Conference on the The Semantic Web, {ESWC}
  2018, Heraklion, Crete, Greece, June 3-7, 2018, Proceedings}, volume 10843 of
  {\em Lecture Notes in Computer Science}, pages 593--607. Springer, 2018.

\bibitem{sen2008collective}
Prithviraj Sen, Galileo Namata, Mustafa Bilgic, Lise Getoor, Brian Gallagher,
  and Tina Eliassi{-}Rad.
\newblock Collective classification in network data.
\newblock {\em {AI} Mag.}, 29(3):93--106, 2008.

\bibitem{shang2019end}
Chao Shang, Yun Tang, Jing Huang, Jinbo Bi, Xiaodong He, and Bowen Zhou.
\newblock End-to-end structure-aware convolutional networks for knowledge base
  completion.
\newblock In {\em 33rd Conference on Artificial Intelligence, {AAAI} 2019,
  Honolulu, Hawaii, USA, January 27 - February 1, 2019}, pages 3060--3067.
  {AAAI} Press, 2019.

\bibitem{Shindo2023}
Hikaru Shindo, Viktor Pfanschilling, Devendra~Singh Dhami, and Kristian
  Kersting.
\newblock {\(\alpha\)}{ILP}: thinking visual scenes as differentiable logic
  programs.
\newblock {\em Mach. Learn.}, 112(5):1465--1497, 2023.

\bibitem{si2019difflog}
Xujie Si, Mukund Raghothaman, Kihong Heo, and Mayur Naik.
\newblock Synthesizing datalog programs using numerical relaxation.
\newblock In Sarit Kraus, editor, {\em 28th International Joint Conference on
  Artificial Intelligence, {IJCAI} 2019, Macao, China, August 10-16, 2019},
  pages 6117--6124. ijcai.org, 2019.

\bibitem{skryagin2022slash}
Arseny Skryagin, Wolfgang Stammer, Daniel Ochs, Devendra~Singh Dhami, and
  Kristian Kersting.
\newblock Neural-probabilistic answer set programming.
\newblock In Gabriele Kern{-}Isberner, Gerhard Lakemeyer, and Thomas Meyer,
  editors, {\em 19th International Conference on Principles of Knowledge
  Representation and Reasoning, {KR} 2022, Haifa, Israel, July 31 - August 5,
  2022}, 2022.

\bibitem{skryagin2020sum}
Arseny Skryagin, Karl Stelzner, Alejandro Molina, Fabrizio Ventola, Zhongjie
  Yu, and Kristian Kersting.
\newblock Sum-product logic: integrating probabilistic circuits into
  deepproblog.
\newblock 2020.

\bibitem{desmet2023deepseaproblog}
Lennert~De Smet, Pedro Zuidberg~Dos Martires, Robin Manhaeve, Giuseppe Marra,
  Angelika Kimmig, and Luc~De Raedt.
\newblock Neural probabilistic logic programming in discrete-continuous
  domains.
\newblock In Robin~J. Evans and Ilya Shpitser, editors, {\em Uncertainty in
  Artificial Intelligence, {UAI} 2023, July 31 - 4 August 2023, Pittsburgh, PA,
  {USA}}, volume 216 of {\em Proceedings of Machine Learning Research}, pages
  529--538. {PMLR}, 2023.

\bibitem{sourek2018lrnn}
Gustav Sourek, Vojtech Aschenbrenner, Filip Zelezn{\'{y}}, Steven Schockaert,
  and Ondrej Kuzelka.
\newblock Lifted relational neural networks: Efficient learning of latent
  relational structures.
\newblock {\em J. Artif. Intell. Res.}, 62:69--100, 2018.

\bibitem{vsourek2021beyond}
Gustav Sourek, Filip Zelezn{\'{y}}, and Ondrej Kuzelka.
\newblock Beyond graph neural networks with lifted relational neural networks.
\newblock {\em Mach. Learn.}, 110(7):1695--1738, 2021.

\bibitem{takeishi2018knowledge}
Naoya Takeishi and Kosuke Akimoto.
\newblock Knowledge-based distant regularization in learning probabilistic
  models.
\newblock {\em CoRR}, abs/1806.11332, 2018.

\bibitem{kbann}
Geoffrey~G. Towell and Jude~W. Shavlik.
\newblock Knowledge-based artificial neural networks.
\newblock {\em Artif. Intell.}, 70(1-2):119--165, 1994.

\bibitem{trouillon2016complex}
Th{\'{e}}o Trouillon, Johannes Welbl, Sebastian Riedel, {\'{E}}ric Gaussier,
  and Guillaume Bouchard.
\newblock Complex embeddings for simple link prediction.
\newblock In Maria{-}Florina Balcan and Kilian~Q. Weinberger, editors, {\em
  33nd International Conference on Machine Learning, {ICML} 2016, New York
  City, NY, USA, June 19-24, 2016}, volume~48 of {\em {JMLR} Workshop and
  Conference Proceedings}, pages 2071--2080. JMLR.org, 2016.

\bibitem{tsamoura2021neural}
Efthymia Tsamoura, Timothy~M. Hospedales, and Loizos Michael.
\newblock Neural-symbolic integration: {A} compositional perspective.
\newblock In {\em 35th Conference on Artificial Intelligence, {AAAI} 2021,
  Virtual Event, February 2-9, 2021}, pages 5051--5060. {AAAI} Press, 2021.

\bibitem{Valkov2018HOUDINILL}
Lazar Valkov, Dipak Chaudhari, Akash Srivastava, Charles Sutton, and Swarat
  Chaudhuri.
\newblock {HOUDINI:} lifelong learning as program synthesis.
\newblock In Samy Bengio, Hanna~M. Wallach, Hugo Larochelle, Kristen Grauman,
  Nicol{\`{o}} Cesa{-}Bianchi, and Roman Garnett, editors, {\em Advances in
  Neural Information Processing Systems 31, NeurIPS 2018, December 3-8, 2018,
  Montr{\'{e}}al, Canada}, pages 8701--8712, 2018.

\bibitem{VanBekkum2021}
Michael van Bekkum, Maaike de~Boer, Frank van Harmelen, Andr{\'{e}}
  Meyer{-}Vitali, and Annette ten Teije.
\newblock Modular design patterns for hybrid learning and reasoning systems.
\newblock {\em Appl. Intell.}, 51(9):6528--6546, 2021.

\bibitem{van2020analyzing}
Emile van Krieken, Erman Acar, and Frank van Harmelen.
\newblock Analyzing differentiable fuzzy logic operators.
\newblock {\em Artif. Intell.}, 302:103602, 2022.

\bibitem{vardi1996temporal}
Moshe~Y. Vardi.
\newblock Why is modal logic so robustly decidable?
\newblock In Neil Immerman and Phokion~G. Kolaitis, editors, {\em Descriptive
  Complexity and Finite Models, {DIMACS} Workshop 1996, Princeton, New Jersey,
  USA, January 14-17, 1996}, volume~31 of {\em {DIMACS} Series in Discrete
  Mathematics and Theoretical Computer Science}, pages 149--183. {DIMACS/AMS},
  1996.

\bibitem{Vashishth2020Compositionbased}
Shikhar Vashishth, Soumya Sanyal, Vikram Nitin, and Partha~P. Talukdar.
\newblock Composition-based multi-relational graph convolutional networks.
\newblock In {\em 8th International Conference on Learning Representations,
  {ICLR} 2020, Addis Ababa, Ethiopia, April 26-30, 2020}. OpenReview.net, 2020.

\bibitem{wang2019satnet}
Po{-}Wei Wang, Priya~L. Donti, Bryan Wilder, and J.~Zico Kolter.
\newblock Satnet: Bridging deep learning and logical reasoning using a
  differentiable satisfiability solver.
\newblock In Kamalika Chaudhuri and Ruslan Salakhutdinov, editors, {\em 36th
  International Conference on Machine Learning, {ICML} 2019, 9-15 June 2019,
  Long Beach, California, {USA}}, volume~97 of {\em Proceedings of Machine
  Learning Research}, pages 6545--6554. {PMLR}, 2019.

\bibitem{wang2017knowledge}
Quan Wang, Zhendong Mao, Bin Wang, and Li~Guo.
\newblock Knowledge graph embedding: {A} survey of approaches and applications.
\newblock {\em {IEEE} Trans. Knowl. Data Eng.}, 29(12):2724--2743, 2017.

\bibitem{wang2019integrating}
Wenya Wang and Sinno~Jialin Pan.
\newblock Integrating deep learning with logic fusion for information
  extraction.
\newblock pages 9225--9232, 2020.

\bibitem{wang2014knowledge}
Zhen Wang, Jianwen Zhang, Jianlin Feng, and Zheng Chen.
\newblock Knowledge graph embedding by translating on hyperplanes.
\newblock In Carla~E. Brodley and Peter Stone, editors, {\em 28th Conference on
  Artificial Intelligence, {AAAI} 2014, July 27 -31, 2014, Qu{\'{e}}bec City,
  Qu{\'{e}}bec, Canada}, pages 1112--1119. {AAAI} Press, 2014.

\bibitem{weber2019nlprolog}
Leon Weber, Pasquale Minervini, Jannes M{\"{u}}nchmeyer, Ulf Leser, and Tim
  Rockt{\"{a}}schel.
\newblock Nlprolog: Reasoning with weak unification for question answering in
  natural language.
\newblock In Anna Korhonen, David~R. Traum, and Llu{\'{\i}}s M{\`{a}}rquez,
  editors, {\em 57th Conference of the Association for Computational
  Linguistics, {ACL} 2019, Florence, Italy, July 28- August 2, 2019, Volume 1:
  Long Papers}, pages 6151--6161. Association for Computational Linguistics,
  2019.

\bibitem{winters2021deepstochlog}
Thomas Winters, Giuseppe Marra, Robin Manhaeve, and Luc~De Raedt.
\newblock Deepstochlog: Neural stochastic logic programming.
\newblock In {\em 36th Conference on Artificial Intelligence, {AAAI} 2022,
  Virtual Event, February 22 - March 1, 2022}, pages 10090--10100. {AAAI}
  Press, 2022.

\bibitem{xu2018semantic}
Jingyi Xu, Zilu Zhang, Tal Friedman, Yitao Liang, and Guy~Van den Broeck.
\newblock A semantic loss function for deep learning with symbolic knowledge.
\newblock In Jennifer~G. Dy and Andreas Krause, editors, {\em 35th
  International Conference on Machine Learning, {ICML} 2018,
  Stockholmsm{\"{a}}ssan, Stockholm, Sweden, July 10-15, 2018}, volume~80 of
  {\em Proceedings of Machine Learning Research}, pages 5498--5507. {PMLR},
  2018.

\bibitem{xu2018how}
Keyulu Xu, Weihua Hu, Jure Leskovec, and Stefanie Jegelka.
\newblock How powerful are graph neural networks?
\newblock In {\em 7th International Conference on Learning Representations,
  {ICLR} 2019, New Orleans, LA, USA, May 6-9, 2019}. OpenReview.net, 2019.

\bibitem{xu2019dynamically}
Xiaoran Xu, Wei Feng, Yunsheng Jiang, Xiaohui Xie, Zhiqing Sun, and Zhi{-}Hong
  Deng.
\newblock Dynamically pruned message passing networks for large-scale knowledge
  graph reasoning.
\newblock In {\em 8th International Conference on Learning Representations,
  {ICLR} 2020, Addis Ababa, Ethiopia, April 26-30, 2020}. OpenReview.net, 2020.

\bibitem{distmult}
Bishan Yang, Wen{-}tau Yih, Xiaodong He, Jianfeng Gao, and Li~Deng.
\newblock Embedding entities and relations for learning and inference in
  knowledge bases.
\newblock In Yoshua Bengio and Yann LeCun, editors, {\em 3rd International
  Conference on Learning Representations, {ICLR} 2015, San Diego, CA, USA, May
  7-9, 2015}, 2015.

\bibitem{Cohen_NeuralLP}
Fan Yang, Zhilin Yang, and William~W. Cohen.
\newblock Differentiable learning of logical rules for knowledge base
  reasoning.
\newblock In Isabelle Guyon, Ulrike von Luxburg, Samy Bengio, Hanna~M. Wallach,
  Rob Fergus, S.~V.~N. Vishwanathan, and Roman Garnett, editors, {\em Advances
  in Neural Information Processing Systems 30, NeurIPS 2017, December 4-9,
  2017, Long Beach, CA, {USA}}, pages 2319--2328, 2017.

\bibitem{yang2020neurasp}
Zhun Yang, Adam Ishay, and Joohyung Lee.
\newblock Neurasp: Embracing neural networks into answer set programming.
\newblock In Christian Bessiere, editor, {\em 29th International Joint
  Conference on Artificial Intelligence, {IJCAI} 2020}, pages 1755--1762.
  ijcai.org, 2020.

\bibitem{zadeh1975fuzzy}
Lotfi~A Zadeh.
\newblock Fuzzy logic and approximate reasoning.
\newblock {\em Synthese}, 30(3-4):407--428, 1975.

\bibitem{ngps}
Lisa Zhang, Gregory Rosenblatt, Ethan Fetaya, Renjie Liao, William~E. Byrd,
  Matthew Might, Raquel Urtasun, and Richard~S. Zemel.
\newblock Neural guided constraint logic programming for program synthesis.
\newblock In Samy Bengio, Hanna~M. Wallach, Hugo Larochelle, Kristen Grauman,
  Nicol{\`{o}} Cesa{-}Bianchi, and Roman Garnett, editors, {\em Advances in
  Neural Information Processing Systems 31, NeurIPS 2018, December 3-8, 2018,
  Montr{\'{e}}al, Canada}, pages 1744--1753, 2018.

\bibitem{zhang2018link}
Muhan Zhang and Yixin Chen.
\newblock Link prediction based on graph neural networks.
\newblock In Samy Bengio, Hanna~M. Wallach, Hugo Larochelle, Kristen Grauman,
  Nicol{\`{o}} Cesa{-}Bianchi, and Roman Garnett, editors, {\em Advances in
  Neural Information Processing Systems 31, NeurIPS 2018, December 3-8, 2018,
  Montr{\'{e}}al, Canada}, pages 5171--5181, 2018.

\bibitem{zhang2020efficient}
Yuyu Zhang, Xinshi Chen, Yuan Yang, Arun Ramamurthy, Bo~Li, Yuan Qi, and
  Le~Song.
\newblock Efficient probabilistic logic reasoning with graph neural networks.
\newblock In {\em 8th International Conference on Learning Representations,
  {ICLR} 2020, Addis Ababa, Ethiopia, April 26-30, 2020}. OpenReview.net, 2020.

\bibitem{pedroassembly}
Pedro Zuidberg Dos~Martires, Vincent Derkinderen, Robin Manhaeve, Wannes Meert,
  Angelika Kimmig, and Luc De~Raedt.
\newblock Transforming probabilistic programs into algebraic circuits for
  inference and learning.
\newblock In {\em Program Transformations for ML Workshop at NeurIPS}, 2019.

\end{thebibliography}

	\appendix

  \section{Knowledge graphs embeddings and Graph Neural Networks for neurosymbolic AI}
	\label{sec:kge_gnn}

	In this appendix, we introduce two approaches, namely Knowledge Graph Embeddings and Graph Neural Networks,  which are commonly used in relational tasks in the deep learning community. We analyze them in the spirit of the seven dimensions introduced in the paper. In fact, it turns out that they share many features with neurosymbolic systems. In this way, we would like to suggest that NeSy can also be found at the intersection of statistical relational and geometric deep learning approaches.
	
	One of the most popular relational representations is that of a knowledge graph (KG). A KG is
	a multi-relational graph composed of entities (i.e. nodes) and relations (i.e. edges). It is common to represent an edge as a triple of the form: \textit{(head entity, relation, tail entity)}, e.g. ($homer$, $fatherOf$, $bart$).
	
	StarAI has been extensively used \cite{kg_srl} to solve many tasks on KGs: prediction of missing relationships (i.e. knowledge graph completion), prediction of properties of entities, or clustering entities based on their connectivity patterns. StarAI is particularly well suited to reasoning with knowledge graphs since it models explicitly the probabilistic dependencies among different relationships. 
	
	However, in order to scale to larger knowledge graphs, the probabilistic dependencies of StarAI models have been relaxed to give rise to a new class of scalable models based on latent features, which are particularly interesting from a neurosymbolic viewpoint.  The key intuition behind relational latent feature
	models is that the relationships between entities can be more efficiently predicted by modeling simpler interactions in a latent feature space. Knowledge Graph Embeddings and Graph Neural Networks represent two of the ways to encode such latent representations.
	

	\subsection{Knowledge Graph Embeddings}
	
	Knowledge Graph Embedding (KGE) models \cite{wang2017knowledge} assume that triples (i.e. relations) are conditionally independent  given a set of global latent variables, called embeddings, for entities and relations. Therefore, the existence of a triple can be predicted by a scoring function $f(e_h,e_r,e_t)$, where $e_i$ is the embedding of the corresponding object.
	
	KGE models mainly differ in terms of the scoring function  and the embedding space   they use. Translation or distance-based models \cite{bordes2013translating,wang2014knowledge} use a scoring function measuring to what extent the tail of a triple can be obtained by a relation-specific translation of the head, i.e. $f(e_h,e_r,e_t) = || e_h + e_r - e_t||$. Semantic matching methods \cite{nickel2011three} instead exploit similarity-based scoring
	functions, such as $f(e_h,W_r,e_t) = || e_hW_re_t^\top||$. It is interesting to note that standard multi-layer perceptrons are often used to learn this similarity measure, i.e. $f(e_h,W_r,e_t) = nn([e_h, e_t]; W_r)$. This is similar to neural interfaces typical of $lPN$ and $LpN$ models of Section \ref{sec:paradigms}, cf. the neural predicates of DeepProbLog in Example~\ref{ex:deepproblog}.
	
	KGE learn from ground relations only. They learn the embeddings of entities and relations as to maximize the score for a set of known \textit{True} triples.  However, some methods   incorporate higher level information, like first-order logical clauses \cite{guo2016jointly, demeester2016lifted} or logical queries \cite{hamilton2018embedding, ren2020beta}, bridging KGE and multiple NeSy systems (like SL\cite{xu2018semantic} or SBR\cite{diligenti2017sbr}). 
	
	It is interesting to analyze KGE methods in terms of the dimensions we described in our paper. KGE methods mostly work as model-based, undirected methods. They constrain the embeddings to be coherent with the logical facts and rules, which are no longer used after learning. They are heavily based on subsymbolic representations. These models learn the correct parameters (i.e. embeddings) for the task at hand. Even if no explicit semantics needs to be given to the scoring function $f(e_h,W_r,e_t)$,  a fuzzy logic interpretation is often used when injecting logical rules \cite{guo2016jointly, demeester2016lifted}
	
	\subsection{Graph Neural Networks}
	
	Graph Neural Networks (GNN) (\cite{scarselli2008graph, xu2018how}) are   deep learning models for dealing with graphs as input. Inference in these models can be cast as  \textit{message passing}  \cite{gilmer2017neural,battaglia2018relational}: at each inference step (i.e. GNN layer), a node sends a message to all its outgoing neighbors and  updates its state by aggregating all the messages coming from its ingoing neighbors. Messages are computed by standard neural networks. Each inference step computes a transformation of the representations of the nodes. In the last layer, these representations are used to classify the nodes or are aggregated to classify the entire graph.

	There are many connections between GNNs and neurosymbolic computation \cite{lamb2020graph} since both of them    apply neural methods to relational data: graphs for GNNs, and logic representations for NeSy. While   many works try to close the gap between these two representations \cite{frasconi2014klog}, the application of GNN techniques to logic-based graph structures is still very limited. The most related line of work is about using GNNs on knowledge graphs \cite{zhang2018link,schlichtkrull2018modeling,shang2019end,xu2019dynamically, nathani2019learning,Vashishth2020Compositionbased}. The underlying idea is to differentiate the messages exchanged by two nodes if they are related by different kinds of edges. For example, given two nodes \texttt{homer} and \texttt{bart},  the message corresponding to the edge \texttt{fatherOf(homer,bart)} will be different from the one corresponding to \texttt{sameFamily(homer,bart)}. However, these models were intended as classifiers of known relational structures (nodes and edges) and not to reason about the knowledge graph itself (e.g. determining whether an edge exists between two nodes). To perform relational reasoning, GNN-based models rely on techniques from the KGE community on top of the representations extracted by the GNN. This often takes the shape of an auto-encoding scheme: a GNN encodes an input graph in a latent representation and a KGE-based factorization technique is used to reconstruct the whole graph \cite{schlichtkrull2018modeling}.

	An important characteristic of GNNs  is that they rely exclusively on neural computation to perform inference (i.e. to compute messages) and there is no clear direction on how to inject external knowledge about inference, e.g. as logical rules. This contrasts with NeSy, where this is one of the main goals. 
	
	There are also some interesting connections between GNNs and StarAI models. In \cite{qu2019probabilistic, zhang2020efficient}, GNNs based on knowledge-graphs are  not used as a modeling choice but rather to approximate inference in Markov Logic Networks, which is somewhat similar to regularization based methods (see Section \ref{sec:proof_vs_model}). Similarly, in \cite{abboud2020learning} GNNs are used to encode logical formulae expressed as graphs  to approximate a weighted model counting problem.
	
	Finally, it is interesting to analyse GNNs in the spirit of some of the dimensions of  NeSy. GNNs act as directed models with a proof-based inference scheme: they perform a series of inference steps to compute the final answer. In the original version of GNNs \cite{scarselli2008graph}, the node states are updated until a fixed point is reached, which resembles forward-chaining in logic programming. The representation of nodes belongs to a  subsymbolic numerical space. Finally, GNNs can be considered as implicit structure learners: inference rules are learned through the learning of the neural message passing functions.

	Graph Neural Networks have recently received a lot of attention from many different communities, thanks to the representation power of neural networks and the capability of learning in complex relational settings. It is no surprise that people have started to study the expressivity of this class of models. One of the most interesting analyses from a neurosymbolic viewpoint is measuring the expressivity of GNNs in terms of variable counting logics. Recently, \cite{morris2019weisfeiler, barcelo2019logical, grohe2021logic} showed that GNNs are as expressive as 2-variable counting logic
	$C_2$. This fragment of first order logic admits formulas with at most two variables extended with counting quantifiers. The expressivity of this fragment is limited  compared to many neurosymbolic models, especially those based on logic. However, GNNs learn the logical structure of the problem implicitly as part of the message passing learning scheme and they rely neither on expert-provided knowledge nor on heavy combinatorial search strategies to structure learning (see Section \ref{sec:struct}).  An open and challenging question that unites the GNN and NeSy communities is how to bring the expressivity to higher-order fragments \cite{morris2019weisfeiler}, like in NeSy and StarAI, while keeping both the learning and the inference tractable, like in GNNs.

  \section{Fuzzy logic, fuzzyfication and soft-Satisfability}
	
	 Fuzzy logic, as many-valued extension over Boolean logic, has a very long tradition \cite{zadeh1975fuzzy,bvelohlavek2017fuzzy}. However, the use of fuzzy logic in StarAI and NeSy is not dictated by the need of dealing with vagueness, but by the advantageous computational properties of t-norms. Indeed, a common use case is to have an initial theory defined in Boolean logic which is \textit{fuzzyfied}. Inference is then carried out with the fuzzyfied theory and the answers are eventually discretized back to Boolean values (usually using a threshold at 0.5).

	
	The reason for this approach is that one would like to exploit the differentiability of t-norms to address logical inference of FOL theories in a more scalable way than standard combinatorial optimization algorithms (e.g. SAT solvers). This is particularly important in undirected and regularization-based methods (such as PSL \cite{bach2017psl} and  LTN \cite{ltn2021aij}). In fact, it has been shown \cite{giannini2018convex} that there are fragments of fuzzy logic that can even provide convex inference problems. Example \ref{ex:bool_vs_fuzzy}, however, shows that naively approaching logical inference through a fuzzy relaxation and gradient-based optimization can introduce unexpected behaviours.

	\begin{boxedexample}[label = ex:bool_vs_fuzzy]{Fuzzyfication and soft-Satisfability}

		Let us consider a disjunction, like $A \vee B \vee C$. In Boolean logic, if we state that the disjunction is \textit{satisfied} (i.e. \textit{True}), then we expect at least one among the three variables to be \textit{True}.

		Suppose we want to find a truth assignment for all the variables that satisfies the disjunction above. The approach of the majority of NeSy fuzzy approaches is the following. First, the rule is relaxed into a fuzzy real function. For example, using the \L ukasiewicz t-norm, $F_\oplus(A,B,C)= min(1, A+B+C)$. Secondly, a gradient-based algorithm (e.g. backpropagation with Adam \cite{kingma2014adam}) is used to maximize the value of the formula with respect to the \textit{fuzzy} truth degree of the three variables. Finally, the obtained \textit{fuzzy} solution $A^\star, B^\star, C^\star$ is translated back into a Boolean assignment using a $0.5$ threshold.

		Let us consider a possible optimal fuzzy solution, like $(A^\star, B^\star, C^\star) = (0.34, 0.34, 0.34)$ and its discretized version $(A^\star, B^\star, C^\star) = (False, False,$ $False)$, using a threshold at $0.5$. The discretized solution does not satisfy the initial Boolean formula, even though it is a global optimum in the fuzzyfied problem. 
	\end{boxedexample}
	
	Similarly, \cite{van2020analyzing} shows that, while it is very common to reason about universally quantified formulae in the form of $\forall x: A(x) \to B(x)$, like `all humans are mortal', using gradients and fuzzy logic to make inference can be extremely counterintuitive, especially with specific t-norms such as the product t-norm. It is unclear whether there exists a generally accepted subset of properties of Boolean logic that one wants to preserve and whether one can define a t-norm that guarantees such properties.
	
	\subsection{Distribution semantics and fuzzy logic semantics}
	Another common reason for using fuzzy logic is to exploit a differentiable semantics. Then, gradient-based methods can be used to train the parameters of a weighted logical theory such as in LRNNs \cite{sourek2018lrnn}. This contrasts with  gradient-based training of the parameters of probabilistic logics based on the distribution semantics.  Possible worlds in probabilistic logic are defined as possible assignments of truth values to all the ground atoms of a logical theory. The assignments of truth values specify the \textit{semantics} of the logic. On the contrary, fuzzy logic assigns continuous truth degrees to formulas or proofs, which are \textit{syntactic} structures. As a consequence, while the probability of an atom will always be equal to the sum of the probabilities of the worlds in which it is \textit{True} (cf. Equation \ref{eq:wmc}), the fuzzy degree of an atom may vary depending on how that atom has been proven or defined, as shown in Example \ref{ex:prob_vs_fuzzy}.

	\begin{boxedexample}[label = ex:prob_vs_fuzzy]{Distribution semantics vs fuzzy logic semantics}

		Consider the following annotated program (adapted from \cite{cussens2001parameter}).
		
		\begin{lstlisting}
		0.3::b.
		a1 <- b.
		a2 <- b,b.
		\end{lstlisting}
		Here, $0.3$ is the label of $\mathtt{b}$ without any particular semantics yet. 
		Given the idempotency of the Boolean conjunction, we would expect  the scores of both $\mathtt{a1}$ and $\mathtt{a2}$ to be identical since both $\mathtt{b}$ and $\mathtt{b} \wedge \mathtt{b}$ are \textit{True} when $\mathtt{b}$ is \textit{True}. In a probabilistic approach, the score is interpreted as a probability, i.e. $p(\mathtt{b}) = 0.3$. The probability of any atom is the sum of the probabilities of all the worlds where that atom is \textit{True}. It is easy to see that, for both $\mathtt{a1}$ and $\mathtt{a2}$, these are the worlds where $\mathtt{b}$ is \textit{True}, thus $p(\mathtt{a1}) = p(\mathtt{a2}) = p(\mathtt{b}) =  0.3$. On the other hand, in the fuzzy setting, the score of $\mathtt{b}$ is interpreted as its truth degree, i.e. $t(\mathtt{b}) = 0.3$. Let us consider the product t-norm, where $t(x \wedge y) = t(x)t(y)$, then $t(\mathtt{a1}) = t(\mathtt{b}) = 0.3$ while $t(\mathtt{a2}) = t(\mathtt{b})t(\mathtt{b}) = 0.09$. While this issue could be solved by choosing a different t-norm (e.g. the minimum t-norm), similar issues arise in different definitions.
	\end{boxedexample}

	The differences between the two semantics are not due to the probabilistic or the fuzzy semantics, but more to the distinction between semantics based on possible worlds and semantics based on proofs or derivations. In fact, a similar behaviour is observed in Stochastic Logic Programs \cite{cussens2001parameter} under the name of \textit{memoization}.

\end{document}